\documentclass{jpsj3}
\usepackage{txfonts}
\usepackage{graphicx,bm}

\usepackage{color}
\usepackage{mathrsfs}
\usepackage{multirow}
\usepackage{booktabs}

\newcommand{\V}[1]{\bm{#1} } 

\newcommand{\Ave}[1]{\left\langle {#1} \right\rangle} 

\newcommand{\lb}{\left(}
\newcommand{\rb}{\right)}
\newcommand{\lbb}{\left\{}
\newcommand{\rbb}{\right\}}

\newcommand{\Req}[1]{eq.\ (\ref{eq:#1})}
\newcommand{\BReq}[1]{Eq.\ (\ref{eq:#1})}
\newcommand{\NReq}[1]{(\ref{eq:#1})}
\newcommand{\Reqs}[2]{eqs.\ (\ref{eq:#1},\ref{eq:#2})}

\newcommand{\NReqs}[2]{(\ref{eq:#1},\ref{eq:#2})}

\newcommand{\Rfig}[1]{Fig.\ \ref{fig:#1}}
\newcommand{\Rfigs}[2]{Figs.\ \ref{fig:#1} and \ref{fig:#2}}
\newcommand{\Rfigss}[2]{Figs.\ \ref{fig:#1}-\ref{fig:#2}}
\newcommand{\NRfig}[1]{\ref{fig:#1}}
\newcommand{\Lfig}[1]{\label{fig:#1}}
\newcommand{\Leq}[1]{\label{eq:#1}}
\newcommand{\Rsec}[1]{sec.\ \ref{sec:#1}}
\newcommand{\Lsec}[1]{\label{sec:#1}}
\newcommand{\be}{\begin{eqnarray}}
\newcommand{\ee}{\end{eqnarray}}
\newcommand{\ba}{\begin{array}}
\newcommand{\ea}{\end{array}}
\newcommand{\no}{\nonumber}

\newcommand{\subbe}{\begin{subequations}}
\newcommand{\subee}{\end{subequations}}

\title{Boltzmann-Machine Learning of Prior Distributions of Binarized Natural Images}

\author{
$^{1,2}$Tomoyuki Obuchi, $^2$Hirokazu Koma, and $^3$Muneki Yasuda
}
\inst{
\address{
$^1$ Department of Computational Intelligence and Systems Science, Tokyo Institute of Technology, Yokohama, 226-8502, Japan.} 
\\
$^2$ Cybermedia Center, Osaka University, Osaka 560-0043, Japan.
\\
$^3$ Graduate School of Science and Engineering, Yamagata University, Yonezawa, 992-8510, Japan.
}

\abst{
Prior distributions of binarized natural images are learned by using a  Boltzmann machine. According  the  results of this study, there emerges a structure with two sublattices in the interactions, and the nearest-neighbor and next-nearest-neighbor interactions correspondingly take two discriminative values, which reflects the individual characteristics of the three sets of pictures that we process. Meanwhile, in a longer spatial scale, a longer-range, although still rapidly decaying, ferromagnetic interaction commonly appears in all cases. The characteristic length scale of the interactions is universally up to approximately four lattice spacings $\xi \approx 4$. These results are derived by using the mean-field method, which effectively
reduces the computational time required in a Boltzmann machine. An improved mean-field method called the Bethe approximation also gives the same results, as well as the Monte Carlo method does for small size images. These reinforce the validity of our analysis and findings. Relations to criticality, frustration, and simple-cell receptive fields are also discussed.
}

\kword{
Boltzmann machine, Markov random fields, neural networks, image processing, mean-field methods
}

\begin{document}
\maketitle

\section{Introduction}
The Bayesian framework of image processing was initiated
in~\cite{Geman:84}, and  is currently an active research field
in several disciplines~\cite{Ruderman:94-1,Ruderman:94-2,Olshausen:96,Simoncelli:01,Tanaka:02,Hyvrinen:09}. This research field has advanced in close relation to neural networks~\cite{Simoncelli:01}, the interest in which has recently begun to grow rapidly by virtue of  new algorithms and
concepts, such as multiple layers of hidden units allowing high orders of statistics to be taken into account naturally~\cite{Hinton:06,Hinton:07,Bengio:09}, as well as the sparseness of signals allowing input signals to be expressed efficiently and universally~\cite{Olshausen:96}. These advances have provided increasingly improved generative models of images that need to be processed.

The Bayesian image processing framework  inevitably requires to introduce prior distributions of the images. Most earlier
studies, in a sense, focused on using one prior distribution that
is applicable to any image of interest. However, we may use
different prior distributions according to the sets of images we
are processing, which may result in a better image processing
 performance.

To examine this possibility, in this study, we evaluate the prior
distributions of several different sets of binarized natural
images by using a Boltzmann machine~\cite{Hinton:85}. The
Boltzmann machine is a classical model of neural networks and is
much simpler than recently developed models, such as deep belief
networks, which consist of multiple layers of hidden
units~\cite{Hinton:06,Hinton:07,Bengio:09}, or methods that
utilize sparseness~\cite{Olshausen:96}. However, to the best of
our knowledge, even such a simple model has never been examined in
this context. We considered that the simple Boltzmann machine
would be a good starting point for the present purpose, because
the correlations among visible units of the Boltzmann machine
become simpler than those of the advanced models with hidden
units, and thus, it should be easier to find individual
characteristics in different categories of natural images.

In particular, the fully connected Boltzmann machine is treated
in the framework of maximum likelihood estimation. Maximization of
the likelihood is known to be difficult in general, and hence, we
use two variants of the mean-field approximations in this study: the naive mean-field approximation and the improved one, called Bethe approximation. The consistency between these two
variants supports the validity of our approximation. In the
analysis, we treat three sets of images as representatives:
aerial geographic pictures, face pictures, and forest pictures.
These pictures have discriminable properties, such as domain and
fractal structures, and we can expect that some discriminative
characteristics would appear in the learning results.

Our learning results of the Boltzmann machine in fact show that
certain discriminative characteristics exist in different
individual sets of images, which clearly appear in the
interactions of the Boltzmann machine. One strong characteristic
is the value of the nearest neighboring (NN) interactions. These
can strongly change according to the individual sets of images,
and can be both positive and negative. The
next-nearest-neighboring (NNN) interactions also depend on the
individual sets, but the tendency is relatively weak. Meanwhile, we
also found some  properties that the different sets of images have
in common. One is the length scale of the interactions. We
observed that, for a spatial range longer than the next-nearest
neighbors, ferromagnetic interactions commonly appear and they
decay in a rapid (approximately exponential) manner; the
characteristic length scale of interactions is commonly
approximately $\xi\approx 4$. This universality of rapidly
 decaying interactions may explain the good performance in image
segmentation reported in \cite{Krahenbuhl:11}, where  an
exponentially decaying interaction in the Markov random field was
introduced. In addition, the Boltzmann machine after learning
universally shows an absence of criticality. This may be a
surprising observation, because natural images are known to
exhibit a certain criticality (power law of the Fourier amplitude
in the spatial frequency)~\cite{Ruderman:94-1,Ruderman:94-2,Simoncelli:01,Hyvrinen:09,Stephens:13}.
This absence of criticality in the Boltzmann machine after
learning implies that the criticality in natural images originates
in higher order statistics than the second one, since the
Boltzmann machine takes into account  statistics only up to the
second order. In addition, it  is commonly observed that
frustration is absent or quite weak, even when it exists. One
possible consequence of the frustration being very weak is a
smooth phase space with a few minima. This potentially provides an
intuitive interpretation of the energy and minima, as suggested
in~\cite{Stephens:13}.

The remaining parts of this paper are organized as follows. In the
following section, we introduce the Boltzmann machine and state
the setup of our problem. The Boltzmann machine is known to be
computationally infeasible and thus we use two approximations,
which are also explained in the section. In \Rsec{Results}, we
display the estimated parameters of the Boltzmann machine. The
results described above are illustrated in detail using these
numerical data. The robustness of the result is also examined in this section by Monte Carlo simulations for small sizes and by testing a number of dithering methods. In \Rsec{Discussion}, we  discuss the criticality
and a relation to simple-cell receptive fields, and finally,
propose a model of the prior distribution of natural images based
on our observations. The final section is devoted to the
conclusion.

\section{Model, Task and Methods}

\subsection{Boltzmann Machine}
In this section, we explain the definition of the Boltzmann machine to state our objectives. The Boltzmann machine is defined by the following energy function or the Hamiltonian 
\be
\mathcal{H}\lb \V{S} |\V{w},\V{h}
\rb=-\sum_{\Ave{i,j}}w_{ij}S_{i}S_{j}-\sum_{i}h_i S_{i},
\Leq{Hamiltonian} 
\ee 
where $S_{i}=\pm 1$ denotes the bit or spin, $w_{ij}$ the interaction, and $h_i$ the field. The probability distribution of spins is given by 
\be p\lb \V{S}|\V{w},\V{h}
\rb=\frac{1}{Z(\V{w},\V{h})}e^{-\mathcal{H}(\V{ S}|\V{w},\V{h})}.
\Leq{Boltzmann} 
\ee 
Note that our Boltzmann machine is fully connected and there is no hidden unit. 

The average of an observable $\hat{O}$ over this distribution is hereafter denoted by 
\be 
\sum_{\V{S}}\hat{O}p \lb \V{S} | \V{w}, \V{h} \rb \equiv \Ave{\hat{O}}_{\V{w},\V{h}}.
\ee

\subsection{Learning of Effective Interactions in Images}
Let us consider the Boltzmann machine in image processing. Suppose
we have $B$ pictures $\{ \V{S}^{(\mu)}  \}_{\mu=1,\cdots,B}$ and
assume that they are generated from a certain distribution. This
distribution is approximated by the empirical distribution 
\be
p_{D}\lb \V{S}|\{ \V{S}^{(\mu)}  \}
\rb=\frac{1}{B}\sum_{\mu=1}^{B}\delta_{\V{S},\V{S}^{(\mu)} }, 
\ee
where $\delta_{\V{S},\V{T}}$ denotes the indicator function giving
unity if $\V{S}=\V{T}$, and zero otherwise. We write the average
by this distribution as $\sum_{\V{S}}p_{D}\hat{O}=\Ave{\hat{O}}_{D}$. A typical learning
scheme is formulated to reproduce the average of several
observables over the empirical distribution by using that over the
Boltzmann distribution 
\be
\Ave{\hat{O}}_{\V{w},\V{h}}=\Ave{\hat{O}}_{D}.
\Leq{moment-matching} 
\ee 
We seek $\V{w},\V{h}$ to satisfy this
moment-matching condition for appropriate observables. The choice
of observables clearly influences the results; here we choose the first
and second moments of spins, magnetization and pairwise
correlations, respectively. Let us fix the notations of the relevant quantities 
\be &&
m_{i}=\Ave{S_{i}}_{\V{w},\V{h}},\,\, \mu_{i}=\Ave{S_{i}}_{D},
\Leq{magnetization}
\\
&& C_{ij}=\Ave{S_{i}S_{j}}_{\V{w},\V{h}}-m_im_j,\,\,
\Gamma_{ij}=\Ave{S_{i}S_{j}}_{D}-\mu_{i}\mu_{j}.
\Leq{connected-correlation} 
\ee 
This choice of observables is
natural, since they are in a conjugate relation with $\V{h}$  and
$\V{w}$. In particular, the solution of these equations,
$p(\V{S}|\V{w}^{*},\V{h}^{*})$, can be written as the maximizer of
the log likelihood 
\be
p(\V{S}|\V{w}^{*},\V{h}^{*})=\arg\max_{p(\V{S}|\V{w},\V{h}) } \lbb
\sum_{\V{S}} p_{D}\lb \V{S}|\{ \V{S}^{(\mu)}  \} \rb \ln  p \lb
\V{S}|\V{w},\V{h} \rb \rbb. 
\Leq{max-loglikelihood}
\ee 
From this equation, the
moment-matching conditions with respect to magnetizations and
pair-wise correlations are naturally derived. The task to be
solved in this study is to infer $\V{w}$ and $\V{h}$ from several
sets of natural images and to find characteristic features.

\subsection{Mean-Field Methods}
To find the optimal values of $\V{w}$ and $\V{h}$,
steepest-descent-type algorithms with respect to $\V{w}$ and
$\V{h}$ are typically used. However, at each step of changing
$\V{w}$ and $\V{h}$,  $\Ave{\hat{O}}_{\V{w},\V{h}}$  need to be
evaluated, which is infeasible for large systems, since the
evaluation of the partition function $Z(\V{w},\V{h})$ in general
requires an exponentially growing time as the system size
increases. To overcome this difficulty, we employ two variants
of the mean-field methods. Namely, the first is the naive
mean-field method (NMF)~\cite{Peterson:87,Hinton:89,Kappen:97} and the second is the improved mean-field method, called Bethe approximation (BA)~\cite{Yasuda:09,Ricci-Tersenghi:12}.

The original model of the Boltzmann machine \NReq{Boltzmann}
involves interactions among variables, and thus, it is difficult
to calculate the partition function. To overcome this difficulty,
the basic strategy of the mean-field methods is to decompose this
multi-body probability distribution into a batch of effective
probability distributions consisting of a small number of
variables. The NMF breaks \Req{Boltzmann} into a batch of
single-spin distributions, while the BA approximates
\Req{Boltzmann} by an appropriate combination of single-spin and
two-spin distributions. Here, we omit the detailed descriptions
and present only the results. Readers interested in the details may refer to~\cite{Peterson:87,Hinton:89,Kappen:97,Yasuda:09,Ricci-Tersenghi:12,ADVA,INFO}.

The NMF results are 
\be 
&&w_{ij}=-\lb
\Gamma^{-1}\rb_{ij}+\frac{\delta_{ij}}{1-\mu_i^2}, 
\Leq{NMF1}
\\
&& 
h_{i}=\tanh^{-1}\mu_{i}-\sum_{j}w_{ij}\mu_j. 
\Leq{NMF2} 
\ee
Note that, in the original model \NReq{Boltzmann}, the self-interaction terms $w_{ii}$ have no meaning, but in the inverse problem  these terms should match the dimension of the
given data $\V{\mu}$ and $\V{\Gamma}$. In addition, the BA
provides the formulas 
\be 
&& 
w_{ij}=\tanh^{-1} 
\Biggl\{
\mu_{i}\mu_{j} -\frac{1}{2\lb \Gamma^{-1} \rb_{ij} } D_{ij}
 \no \\
&&
+\frac{ 1}{ \lb \Gamma^{-1} \rb_{ij} }
 \sqrt{ \frac{1}{4}-\mu_i \mu_j\lb \Gamma^{-1}\rb_{ij}D_{ij}
+\lb 2\mu_{i}^2\mu_{j}^2-\mu_{i}^2-\mu_{j}^2\rb \lb \Gamma^{-1} \rb_{ij}^{2}
}
\Biggr\}
,
\Leq{BA1}
\\
&&
h_{i}=\tanh^{-1}(\mu_i)-\sum_{j}\tanh^{-1}(t_{ij}f(\mu_j,\mu_i,t_{ij})),
\Leq{BA2}
\ee
where $t_{ij}=\tanh w_{ij}$ and
\be
&&
D_{ij}=\sqrt{ 1+4(1-\mu_{i}^2)(1-\mu_{j}^2) \lb \Gamma^{-1}\rb_{ij}^{2}    },
\\
&& f(\mu_1,\mu_2,t)=\frac{ 1-t^2-  \sqrt{(1-t^2)^2
-4t(\mu_1-\mu_2t)(\mu_2-\mu_1t) }    }{2t(\mu_2-\mu_1 t)}. \ee The
BA can be expected to yield more accurate results, since it
includes higher body correlations than does the NMF, and  the
computational time of \Reqs{BA1}{BA2} remains comparable with that
of NMF \NReqs{NMF1}{NMF2}. The above formula of the BA is taken
from~\cite{Ricci-Tersenghi:12}.

\section{Results}\Lsec{Results}
In this section, we show the estimated $\V{w}$ and $\V{h}$ for
three sets of natural pictures. One comprises aerial pictures, the
second human face pictures, and the third forest pictures. We
selected these three sets rather arbitrarily, meaning that each
set appears to have some discriminative properties, such as the
sizes of clusters and shapes of edges, of the other two sets.

Each set of pictures, aerial, face, and forest, are downloaded
from the database of the Geospatial Information Authority of
Japan~\cite{GSI}, the Color FERET Database~\cite{FERET}, and the SUN Database~\cite{SUN}, respectively.
The original pictures are multi-colored and are binarized by ImageMagick~\cite{ImageMagick}, and we assume a black dot is represented by $+1$ and a white one by $-1$. More precisely, a spatial color quantization~\cite{Quantization} with gray colorspace and the Riemersma dither~\cite{Riemersma} are employed in the binarizing process. Typical examples of each
set, aerial, face, and forest pictures, are given in
\Rfigss{aerials}{forests}, respectively. The empirical
distribution of pictures having a fixed size is generated by
cutting many binarized pictures into several square-shaped patches
with equal sides having a length of $L$, and mixing them equally.
\begin{figure}[htbp]
\begin{center}
  \includegraphics[width=0.25\columnwidth]{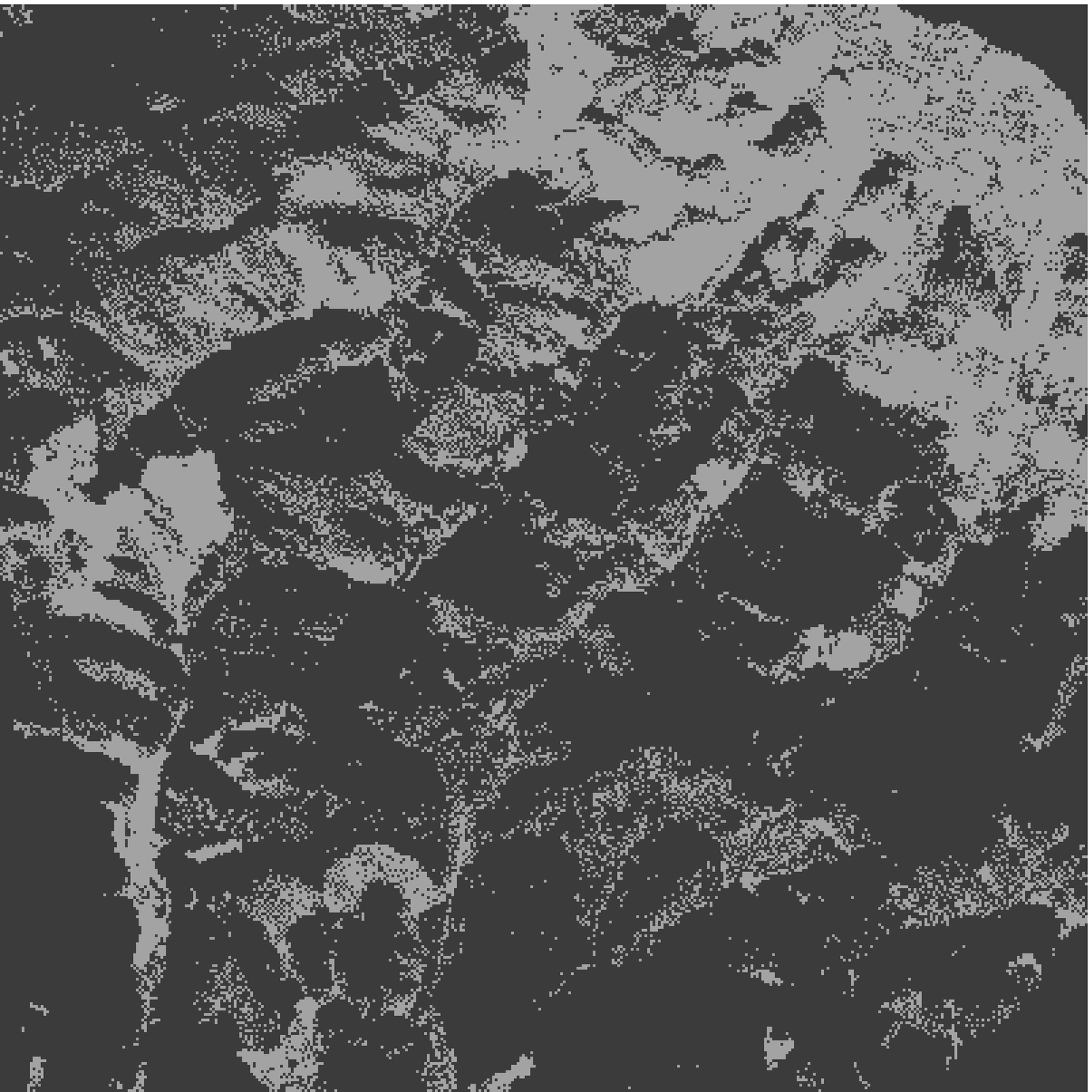}
  \includegraphics[width=0.25\columnwidth]{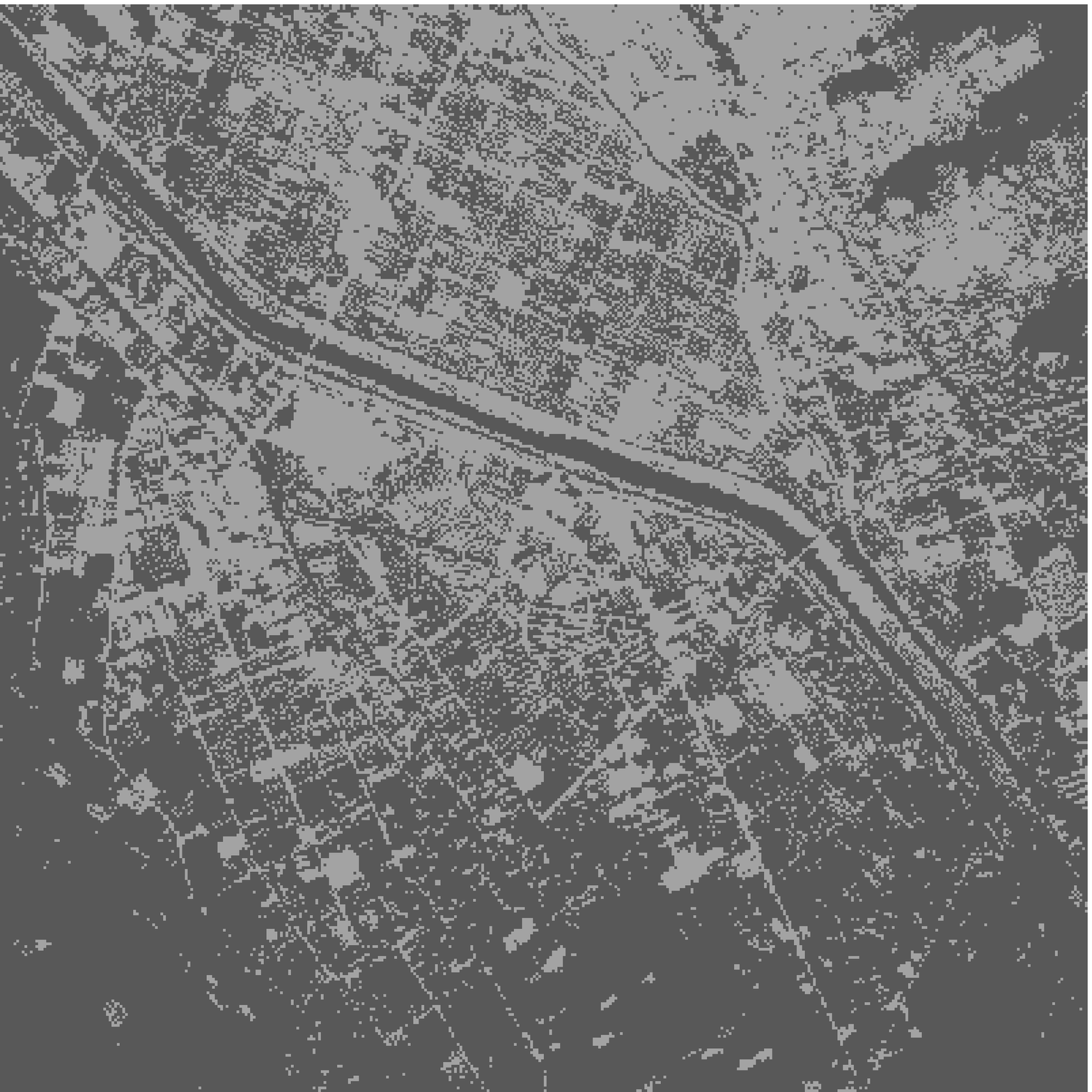}
  \includegraphics[width=0.25\columnwidth]{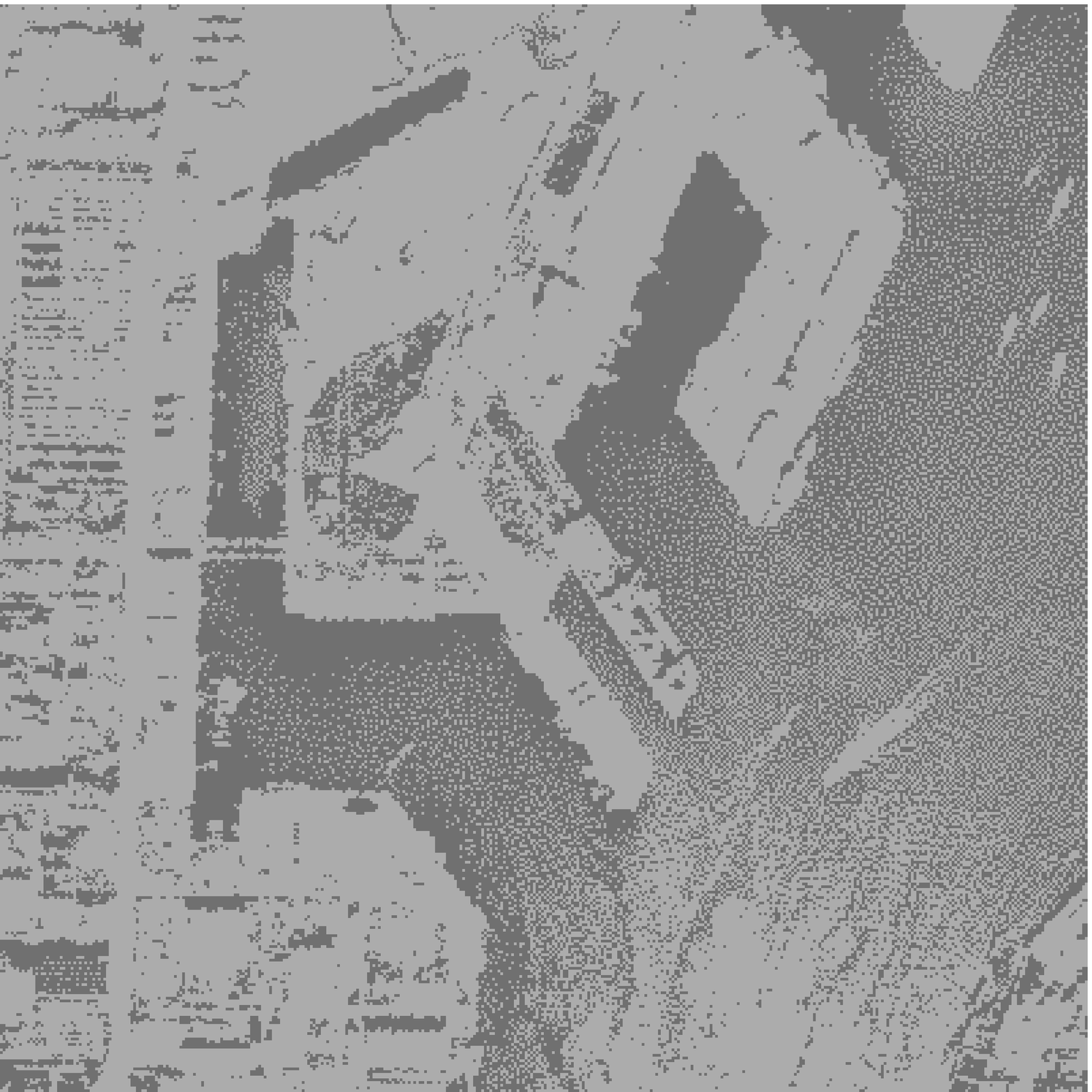}
\caption{Typical binarized aerial pictures used in learning. From
\cite{GSI}.} \Lfig{aerials}
\end{center}
\end{figure}
\begin{figure}[htbp]
\begin{center}
  \includegraphics[width=0.25\columnwidth]{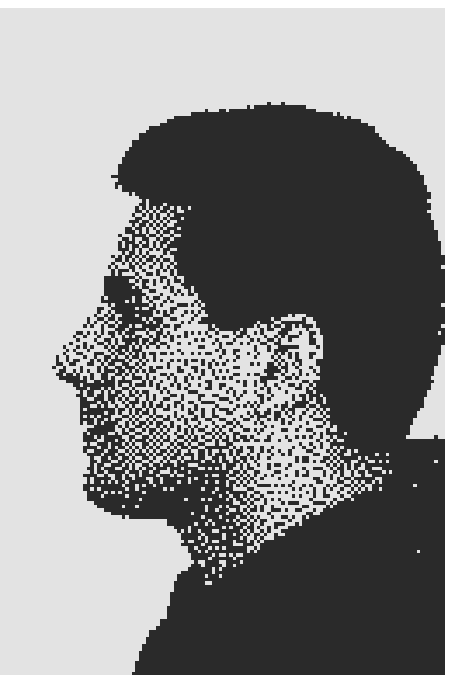}
  \includegraphics[width=0.25\columnwidth]{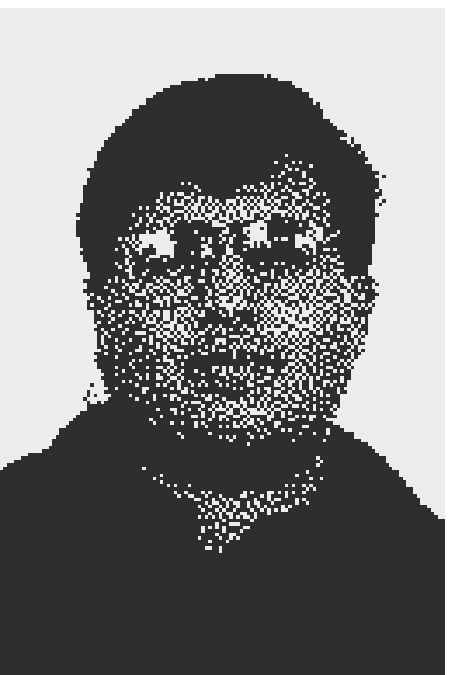}
  \includegraphics[width=0.25\columnwidth]{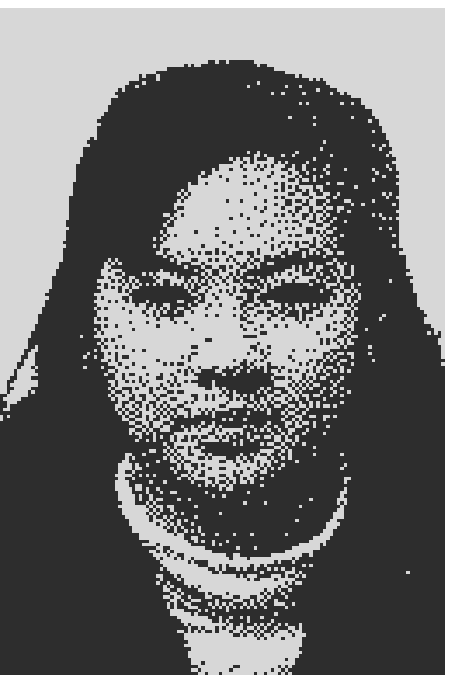}
\caption{Typical binarized face pictures used in learning. From
\cite{FERET}.} \Lfig{faces}
\end{center}
\end{figure}
\begin{figure}[htbp]
\begin{center}
  \includegraphics[width=0.25\columnwidth]{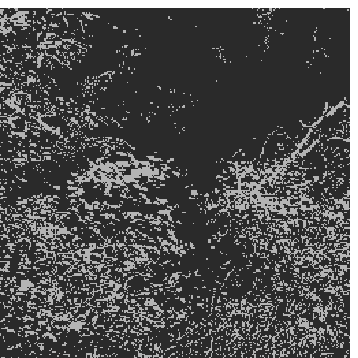}
  \includegraphics[width=0.25\columnwidth]{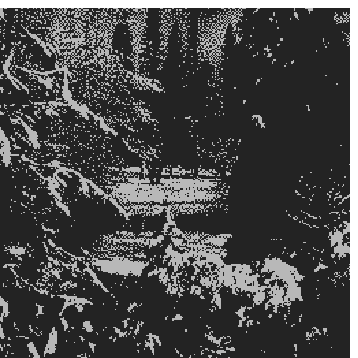}
  \includegraphics[width=0.25\columnwidth]{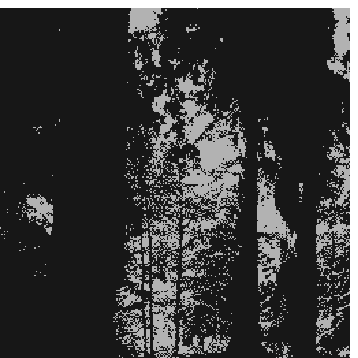}
\caption{Typical binarized forest pictures used in learning.
From \cite{SUN}.} \Lfig{forests}
\end{center}
\end{figure}

For simplicity of notation, hereafter, let us represent each site
$i$ by the orthogonal coordinates $\V{r}_i=(x_i,y_i)$ with
integers $x_{i},y_{i}=1,2,\cdots$. We allocate these coordinates $(x_i,y_i)$ to pixels, as shown in \Rfig{geometry}.
\begin{figure}[htbp]
\begin{center}
  \includegraphics[width=0.25\columnwidth]{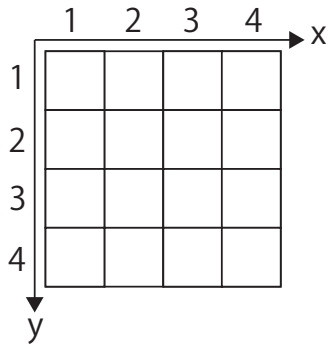}
\caption{Correspondence between coordinates and pixels. Each square represents a pixel.
The origin locates at the upper left edge of the patch.}
\Lfig{geometry}
\end{center}
\end{figure}
The interaction between pixels $i$ and $j$ is rewritten as
$w_{ij}=w(\V{r}_i,\V{r}_j)$. We assume the distance between two
pixels $i$ and $j$ is defined in the Euclid manner,
$r_{ij}=\sqrt{(x_{i}-x_{j})^2+(y_{i}-y_{j})^2}$.

\subsection{Inferred Interactions}\Lsec{Interactions}

\subsubsection{Aerial Pictures}\Lsec{Interactions-aerial}
In this section, we observe the inferred interactions of the
aerial pictures. As clarified later by a comparison of this set
with the other two sets of pictures, this case is the simplest,
where the interactions take only positive values.

First, let us observe some basic behavior of $w(\V{r}_i,\V{r}_j)$
inferred by the NMF in \Reqs{NMF1}{NMF2}, focusing in particular
on the dependency on the distance $r=|\V{r}_j-\V{r}_i|$ and on the
origin $\V{r}_i$. In this context, we rewrite
$w(\V{r}_i,\V{r}_j)=w(r|\V{r}_i)$. In
\Rfig{L=16-aerials-NMF-basic}, we plot $w(r|\V{r}_i)$ of common
$y$-coordinate $y_i=y_j$ against $r=|x_j-x_i|$ with
$x_j=x_i+1,x_i+2,\cdots x_i+8$. Namely, we plot the interactions
in the row as they change across the column from left to right.
Several different curves are shown when the origin
$\V{r}_i=(x_i,y_i)$ changes. The size and number of the patches
are $L=16$ and $B=100,000$, respectively.
\begin{figure}[htbp]
\begin{center}
  \includegraphics[width=0.40\columnwidth]{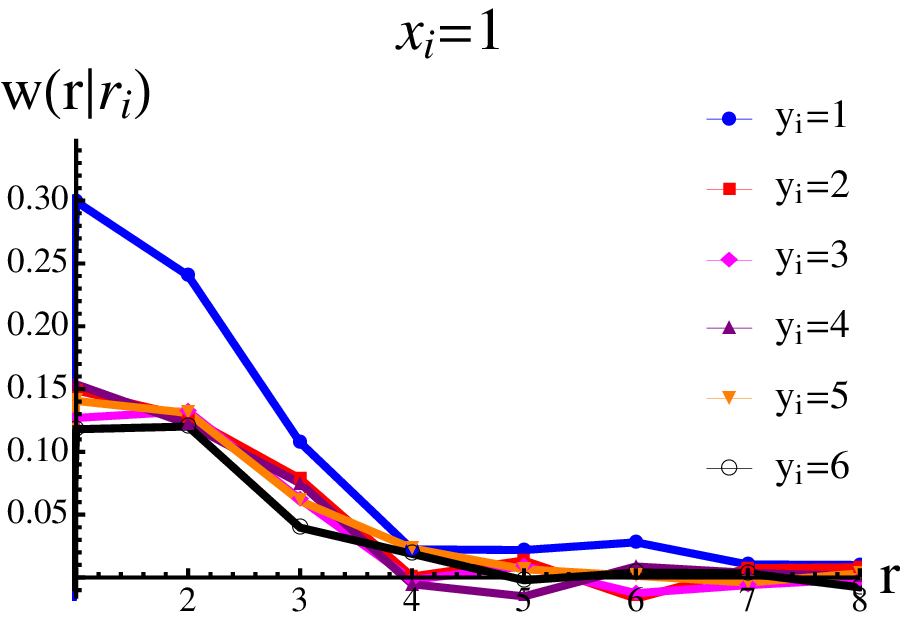}
  \includegraphics[width=0.40\columnwidth]{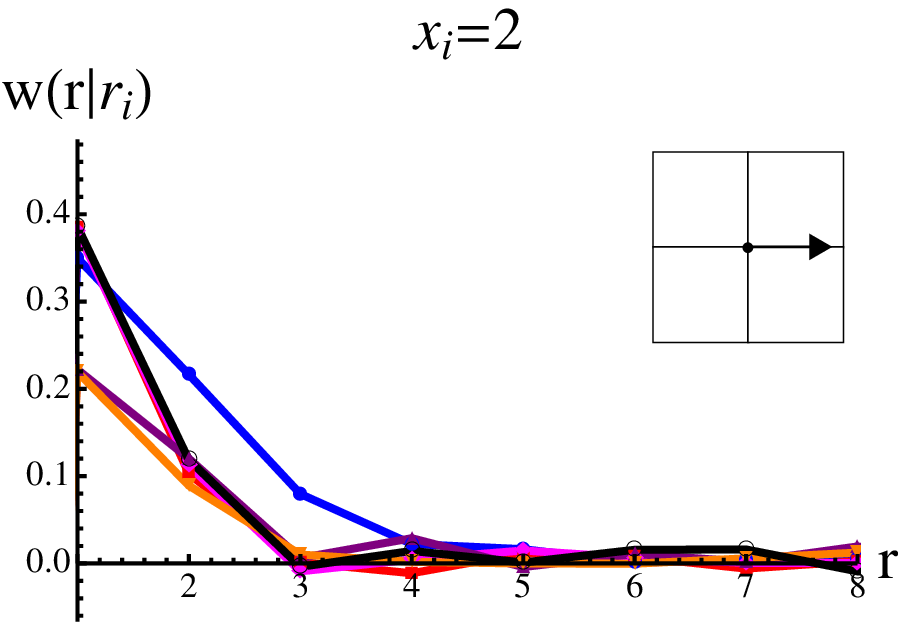}
  \includegraphics[width=0.40\columnwidth]{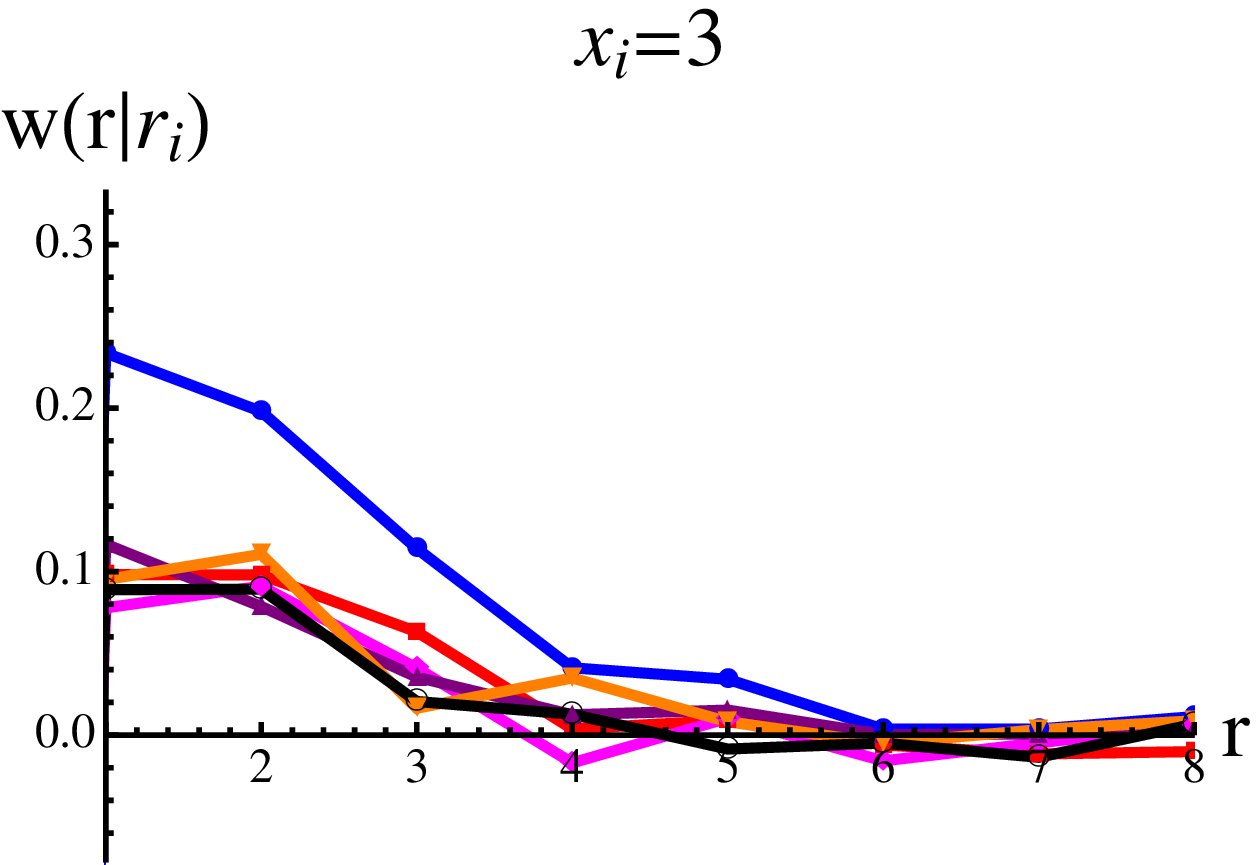}
  \includegraphics[width=0.40\columnwidth]{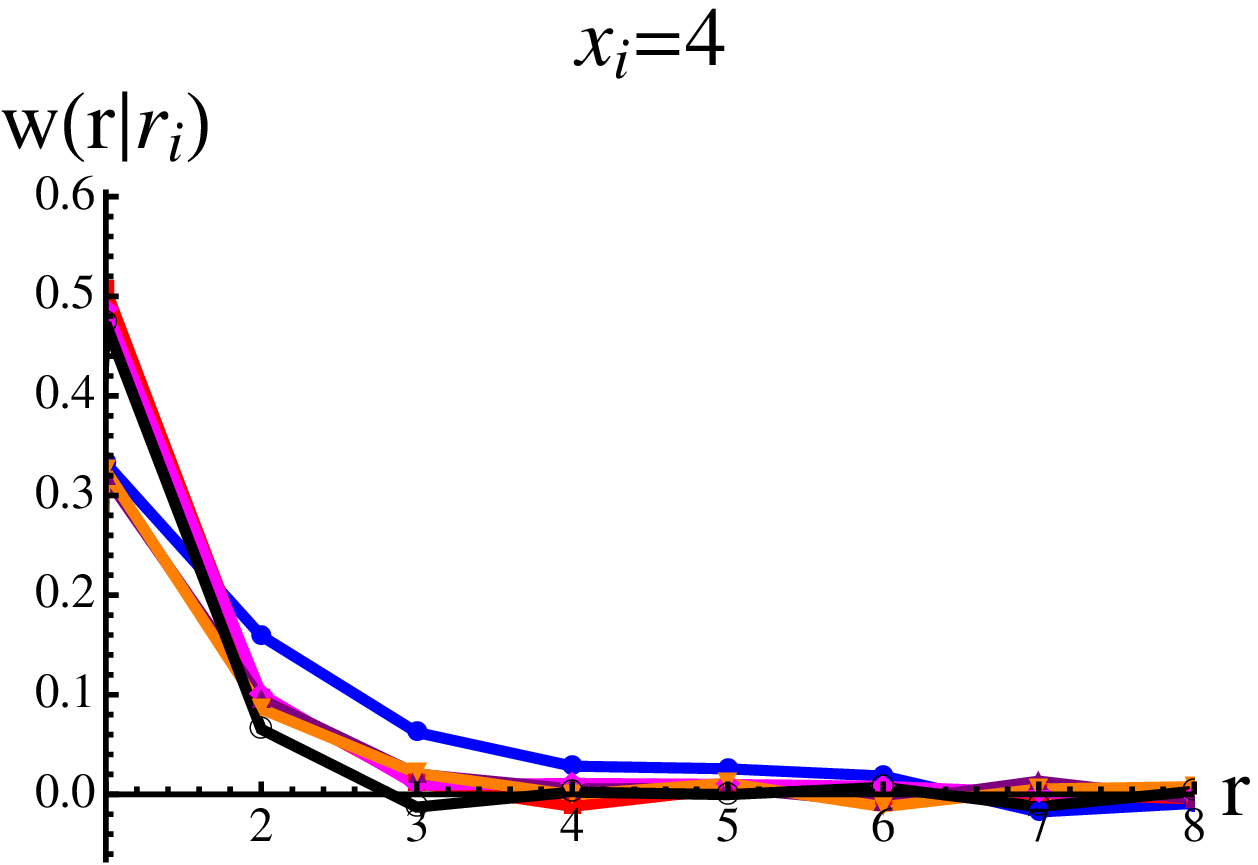}
 \caption{(Color online) Plots of interactions $w(r|\V{r}_i)$ against distance $r=|x_j-x_i|$
 in a common row $y_i=y_j$
 as  the origin of the plot $\V{r}_i=(x_i,y_i)$ changes
 (the top-right inset represents the moving direction of $\V{r}_{j}$).
 The interactions are inferred by the NMF using  $B=100,000$ patches of size $L=16$
 created from the aerial pictures. }
\Lfig{L=16-aerials-NMF-basic}
\end{center}
\end{figure}
We confirm that $w(r|\V{r}_i)$ along the opposite direction,
namely, from right to left, shows quite similar behavior, and that
along the vertical direction in the column it does as well. Thus,
\Rfig{L=16-aerials-NMF-basic} is a good representative. Crucial
observations from \Rfig{L=16-aerials-NMF-basic} are as follows.
\begin{itemize}
\item{Most of the interactions are positive for $r\leq 4\equiv \xi $ and almost vanish for $r>\xi$. } 
\item{A clear boundary effect exists for $y_i=1$ (blue curves). Namely, the absolute values of
$w(r|y_i=1)$ are  larger than the other values of $y_i$. } 
\item{A periodic behavior exists in the interactions, in particular for the NN ones at $r=1$. Namely, the absolute values of the NN interactions for $x_i=2$ and $4$ are similar, as well as those for
$x_i=1$ and $3$, but the former  for $x_i=2$ and $4$ are larger
than the latter for $x_i=1$ and $3$. } 
\item{A weaker periodic behavior appears to exist among different $y_i$ for even $x_i=2$
and $4$. For example, for $x_i=2$, curves with $y_i=2,3$ and $6$
behave similarly, as do other curves of $y_i=4$ and $5$. }
\end{itemize}
The first three characteristics listed above are clearer than the
fourth, and we checked that they are common among other rows and
columns and similar counterparts exist for the other sets of
pictures, faces and forests. Hence, the analyses below are based
on these three findings. The final fourth characteristic, a weaker
periodic behavior, also seems to hold for the other two sets of
pictures, but we avoid performing the analysis since it is not
easy to treat systematically because of the weakness of the
tendency and the complexity of the periodicity.

Next, we observe the orientation dependency of
$w(\V{r}_i,\V{r}_j)$. We plot $w(\V{r}_i,\V{r}_j)$ when we vary
$\V{r}_j$ as $\V{r}_j=(x_i+s,y_i+s)$ with $s=1,2,\cdots,6$,
namely, along the downward slope of the 45-degree angle from fixed
$\V{r}_i$. The distance then becomes $r_{ij}=\sqrt{2}s$.
\begin{figure}[htbp]
\begin{center}
  \includegraphics[width=0.40\columnwidth]{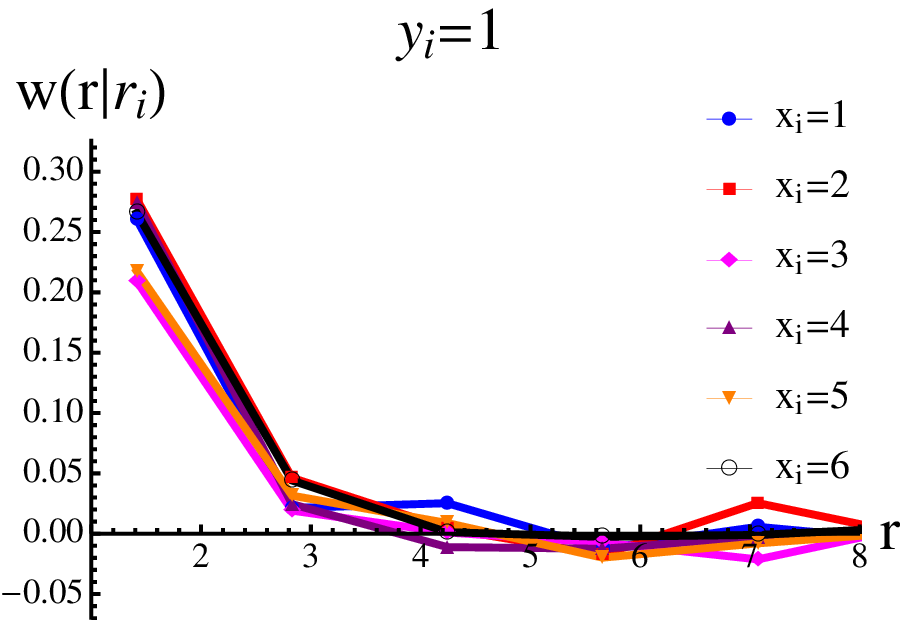}
  \includegraphics[width=0.40\columnwidth]{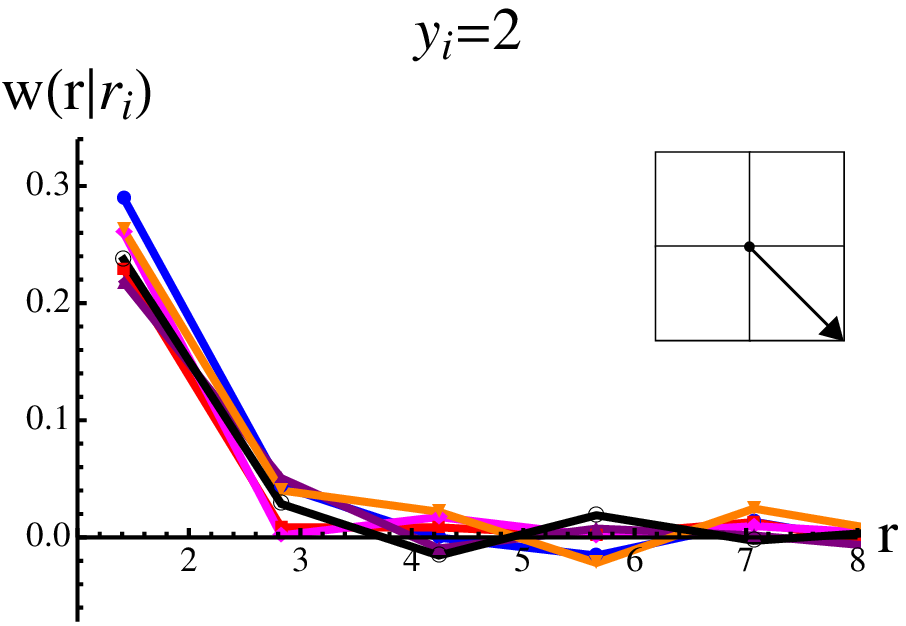}
  \includegraphics[width=0.40\columnwidth]{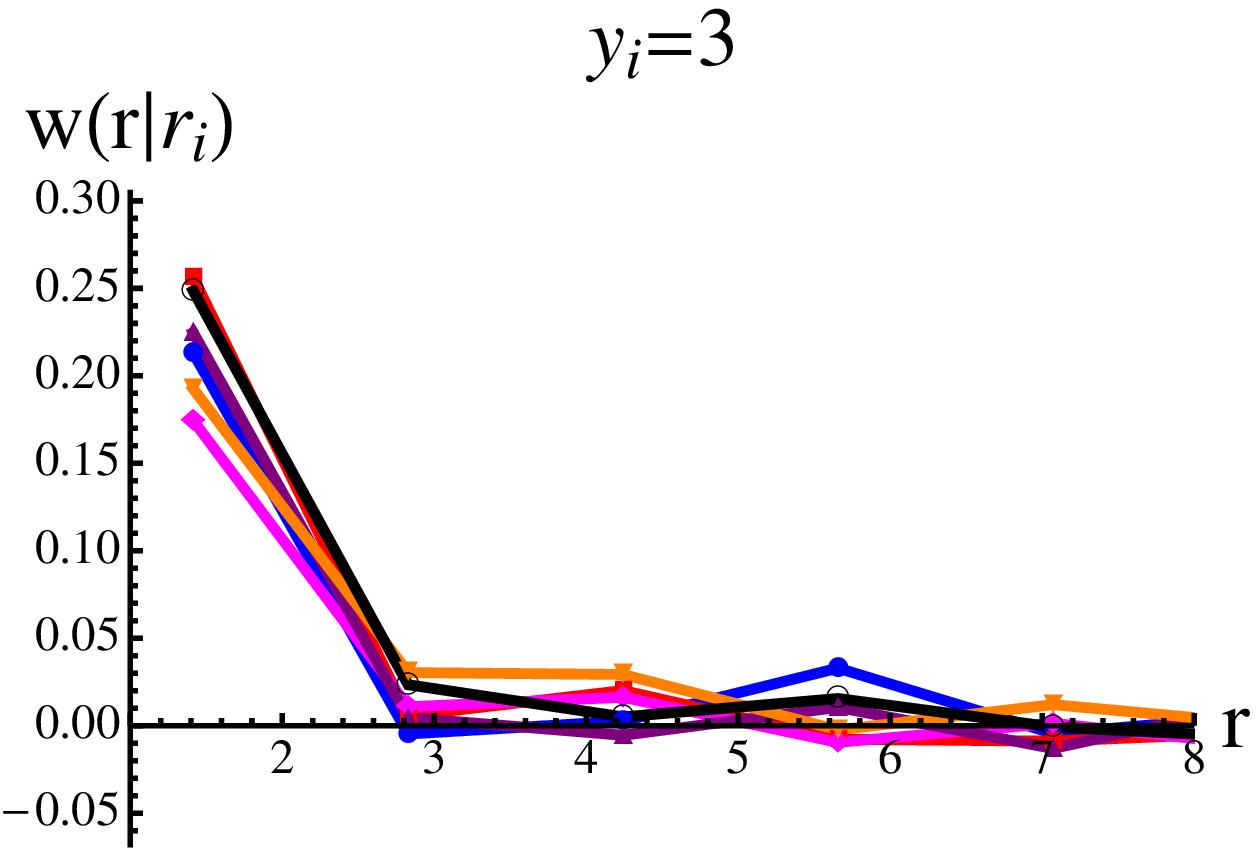}
  \includegraphics[width=0.40\columnwidth]{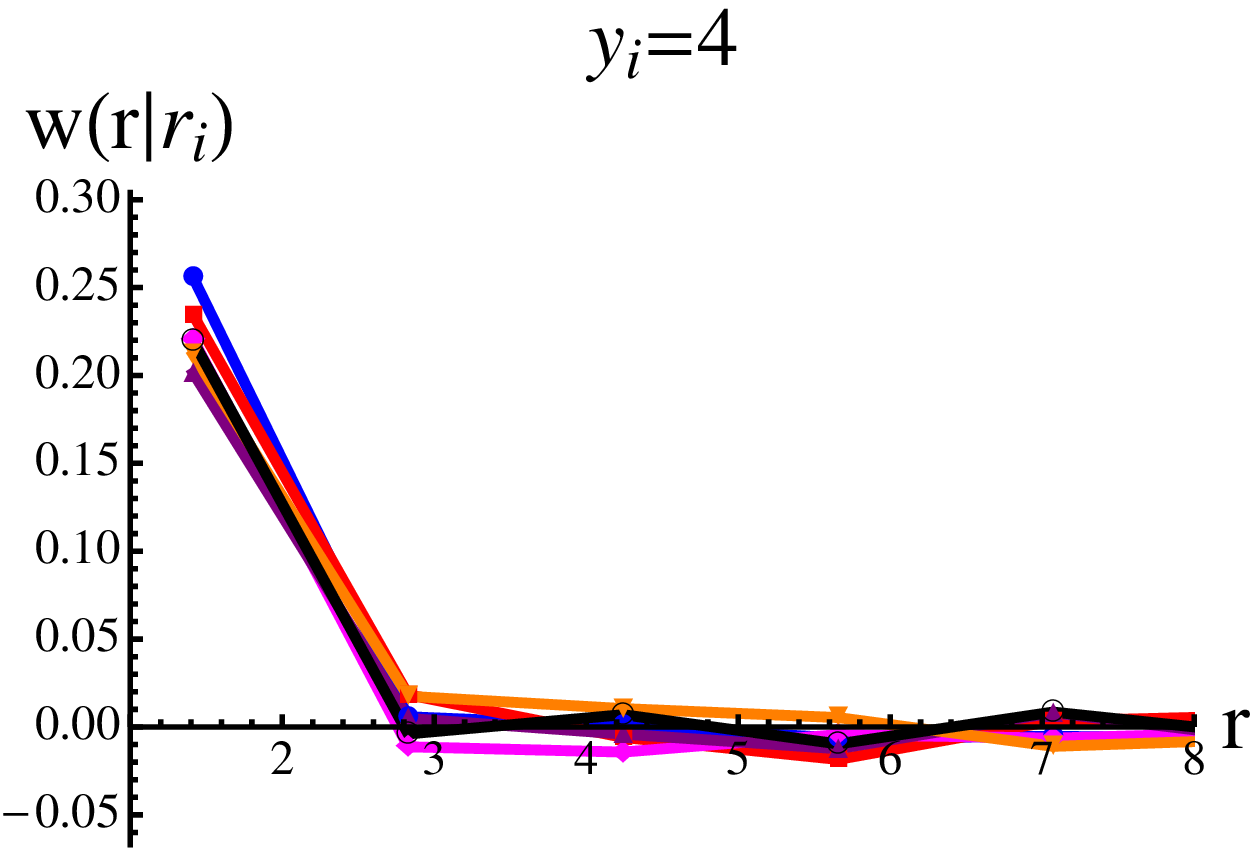}
 \caption{(Color online) Plots of interactions $w(r|\V{r}_i)$
 against distance $r_{ij}=\sqrt{2}s$ with $s=1,2,\cdots 6$ along the downward slope of the 45-degree
 angle from fixed $\V{r}_i$. The same images and parameters as in \Rfig{L=16-aerials-NMF-basic} are used.}
\Lfig{L=16-aerials-NMF-ds}
\end{center}
\end{figure}
\Rfig{L=16-aerials-NMF-ds} again shows that the interactions are
ferromagnetic and are not long ranged; they almost vanish for $r
\geq 3$. This length scale is shorter than that  shown in
\Rfig{L=16-aerials-NMF-basic}, implying that the interactions
along the horizontal or vertical lines are stronger than those
along other inclined directions. Here, it is not easy to find any
clear dependency on the choice of origins $\{ \V{r}_i \}$, and
thus, the periodicity becomes weaker in this direction. The
absence of direction dependency is again checked by examining the
interactions along the opposite direction and those along the
upward slope of the 45-degree angle.

The above findings of periodicity imply the existence of a sublattice structure as represented in the left panel of \Rfig{sublattice}. For  convenience later in this  paper, we also depict the sublattice structure of the NNN interactions, which is not seen in aerial pictures, but is seen in face pictures, as shown in the center panel in \Rfig{sublattice}.
\begin{figure}[htbp]
\begin{center}
  \includegraphics[width=0.26\columnwidth,height=0.26\columnwidth]{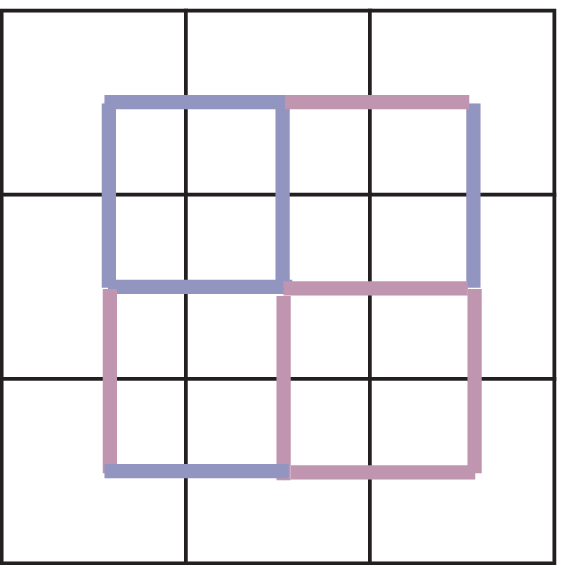}
\hspace{3mm}
  \includegraphics[width=0.26\columnwidth,height=0.26\columnwidth]{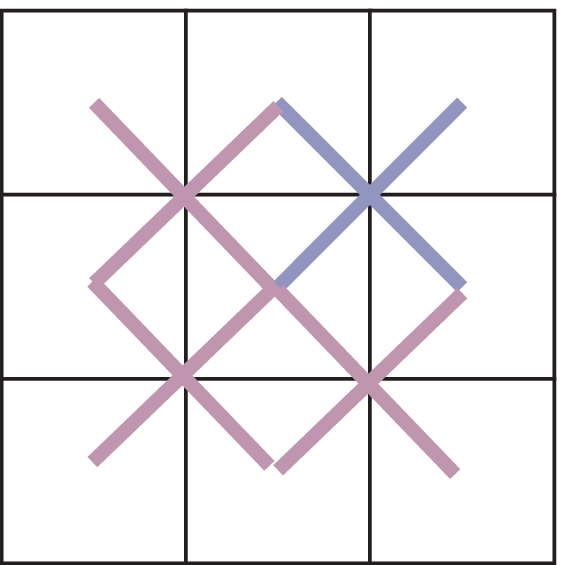}
\hspace{3mm}
  \includegraphics[width=0.36\columnwidth,height=0.26\columnwidth]{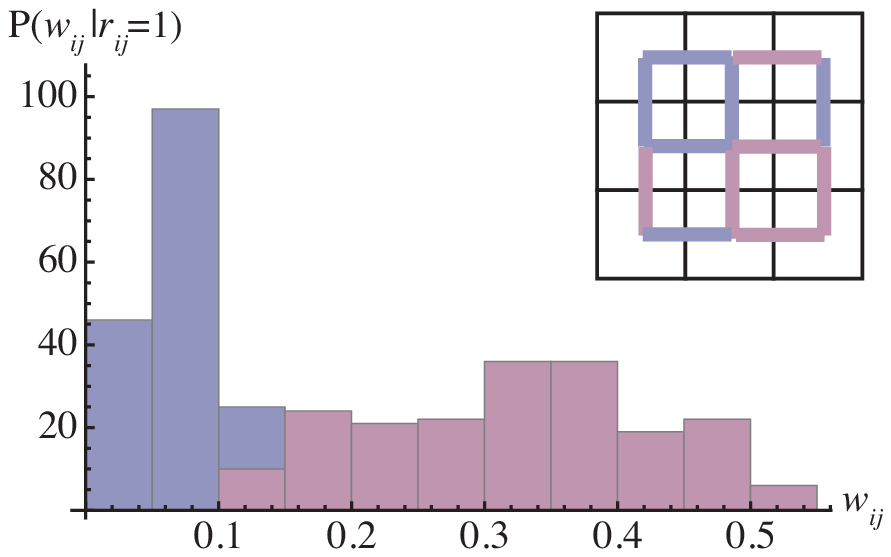}
 \caption{(Color online) 
 (Left) Sublattice structure implied by the periodicity in the NN interactions, where the squares correspond to pixels and links between neighboring squares represent the interactions. Two types of link, which are represented by blue and magenta bars, emerge.
 (Center) Sublattice structure in the NNN interactions (clearly seen in the case of face pictures).
 (Right) Histogram of NN interactions  discriminating the two types of links, which are represented by two different colors, blue and magenta, in correspondence with the left panel. The inset is the reduced scale version of the left panel. The distribution of the values of $w_{ij}$ are clearly different in these two types of NN links. }
\Lfig{sublattice}
\end{center}
\end{figure}
This sublattice structure periodically paves the whole patch without overlap, and thus each link is uniquely determined to have which color. Note that the color just indicates which sublattice the link belongs to, and has no relation to the sign of the corresponding interaction. The sublattice structure of the NN interactions is captured well by the histogram of the NN interactions $P(w_{ij}|r_{ij}=1)$, which is shown in the right panel of \Rfig{sublattice}. The histograms of these two different types of interaction (blue and magenta) are clearly differently distributed. This evidence strongly supports the presence of the sublattice structure.

The system-size dependence of the above results, in particular the
characteristic length scale of the interaction range, needs to be
examined. For this purpose, we define the following averaged
interactions \be \overline{w} \lb r \bigr| (\hat{x}_{i},\hat{y}_i)
\rb= &&\hspace{-5mm} \frac{1}{2(L-2)} \Biggl\{
\sum_{x_j=x_i+1}^{L}\sum_{y_i=2}^{L-1} w\lb r_{ij}=|x_j-\hat{x}_i|
\bigr| \V{r}_i=(\hat{x}_i,y_i) \rb \delta \lb r-r_{ij} \rb
\no \\
&& + \sum_{y_j=y_i+1}^{L}\sum_{x_i=2}^{L-1} w\lb
r_{ij}=|y_j-\hat{y}_i| \bigr| \V{r}_i=(x_i,\hat{y}_i) \rb \delta
\lb r-r_{ij} \rb \Biggr\}. \Leq{w-average} \ee As seen thus far,
the behavior of $w(r|\V{r}_i)$ in $r \geq 2$ is stable and does
not significantly fluctuate over different directions and origins,
which justifies taking this average. The first sum in
\Req{w-average} is the contributions of $w(r|\V{r}_i)$ along the
horizontal direction moving right from $\hat{x}_i$ and the second
is those along the vertical direction moving down from
$\hat{y}_i$. Contributions from the boundaries (blue curves in
\Rfig{L=16-aerials-NMF-basic}) are excluded in \Req{w-average}. We
can see sublattices in \Rfig{sublattice} by appropriately choosing
the coordinates $(\hat{x}_i,\hat{y}_i)$. We define and plot
$\overline{w_{A}}(r)\equiv \overline{w} \lb r \bigr| (1,1) \rb $
and $\overline{w_{B}}(r)\equiv \overline{w} \lb r \bigr| (2,2) \rb
$ of different patch sizes $L=8,$  $16,$ and $32$ in
\Rfig{sizecomp-aerials}.
\begin{figure}[htbp]
\begin{center}
  \includegraphics[width=0.40\columnwidth]{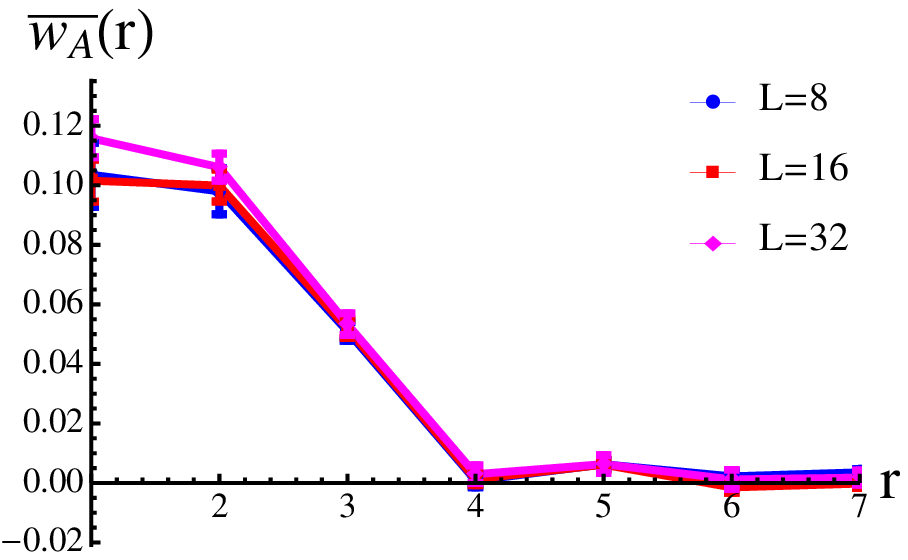}
  \includegraphics[width=0.40\columnwidth]{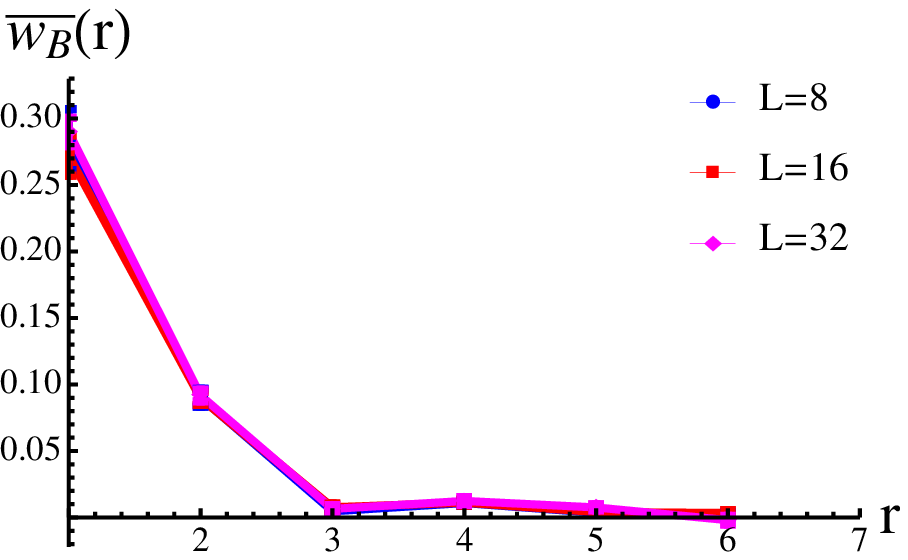}
 \caption{(Color online) Plots of $\overline{w_{A}}(r)\equiv \overline{w} \lb r \bigr| (1,1) \rb $ (left)
 and $\overline{w_{B}}(r)\equiv \overline{w} \lb r \bigr| (2,2) \rb $ (right)
  for different patch sizes $L=8,~16,$ and $32$. The number of  used patches  is $B=400,000$
  and $23400$ for $L=8$ and $32$, respectively. The size effect is clearly absent.
}
\Lfig{sizecomp-aerials}
\end{center}
\end{figure}
The number of used patches  in  the plots shown in
\Rfig{sizecomp-aerials} is $B=400,000$ and $23400$ for $L=8$ and
$32$, respectively. The error bar of each data point in
\Rfig{sizecomp-aerials} is defined by $\sigma(r)/\sqrt{2(L-2)}$,
where $\sigma(r)$ is the standard deviation of the terms in
\Req{w-average} from $\overline{w} \lb r \bigr|
(\hat{x}_{i},\hat{y}_i) \rb$. The results clearly demonstrate the
absence of the size effect, which is consistent with  the
characteristic length scale of $w(r|\V{r}_i)$ being $\xi \approx
4$. Patches larger than $L=8$, which is twice $\xi\approx 4$,
suffice to reproduce our findings.

Next, we compare the results obtained by the NMF and BA shown in
\Rfig{NMFBAcomp-aerials}.
\begin{figure}[htbp]
\begin{center}
  \includegraphics[width=0.40\columnwidth]{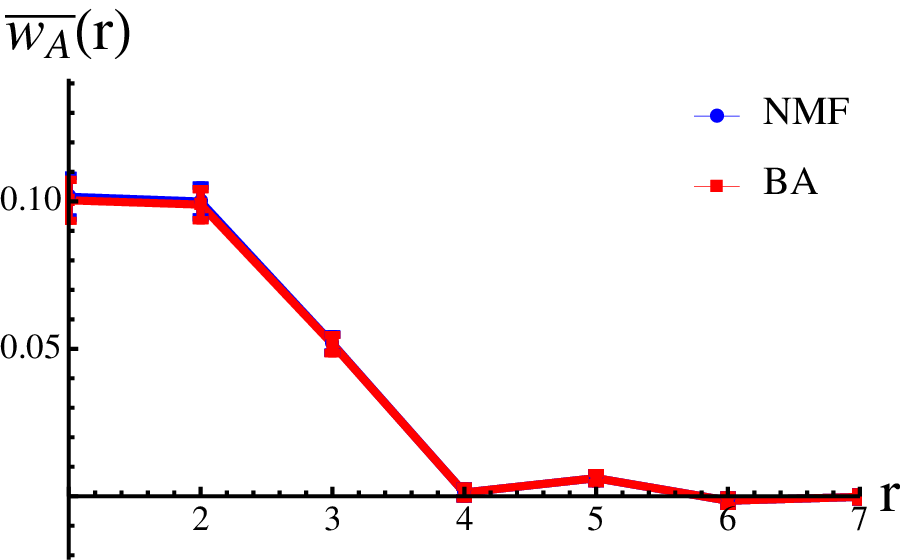}
  \includegraphics[width=0.40\columnwidth]{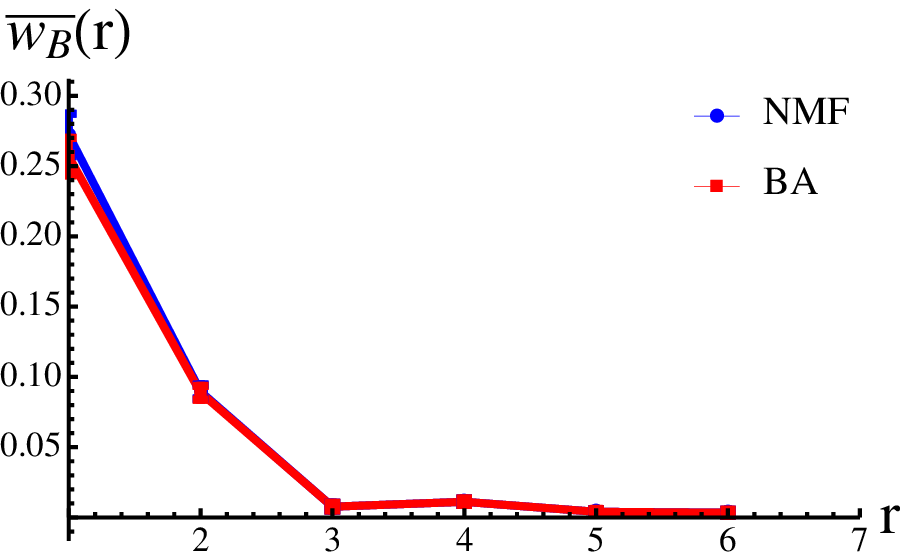}
 \caption{(Color online) Plots of $\overline{w_{A}}(r)\equiv \overline{w} \lb r \bigr| (1,1) \rb $ (left)
 and $\overline{w_{B}}(r)\equiv \overline{w} \lb r \bigr| (2,2) \rb $ (right)
 inferred by the NMF and BA for patch size $L=16$. The absolute values of $w$ are slightly larger in the NMF
 than in the BA, but the difference is negligibly tiny.
}
\Lfig{NMFBAcomp-aerials}
\end{center}
\end{figure}
We can see the difference between the NMF and the BA is negligibly
small. Thus, for our purpose of finding the characteristics of the
interactions, the NMF appears to suffice, at least for the aerial
pictures we treat.

\subsubsection{Face Pictures}\Lsec{Interactions-face}
Next, we show the learned interactions of face pictures. In
general, we can identify three regions in each picture in this
category: the background, the hair and cloth, and the face itself,
as seen in \Rfig{faces}. The face region is expressed by patterns
where both black and white pixels emerge frequently and
alternately, which is considered to be produced by a dither
process to discriminate this region from the others. Presumably as
a result of this dither process, some antiferromagnetic
interactions, which are absent in the aerial pictures, are
observed, as seen below.

We first observe the NMF results in the case of the aerial
pictures. In \Rfig{L=16-faces-NMF-basic}, we display the
interactions $w(r|\V{r}_i)$ against the distance $r$ in a column,
which correspond to \Rfig{L=16-aerials-NMF-basic}.
\begin{figure}[htbp]
\begin{center}
  \includegraphics[width=0.40\columnwidth]{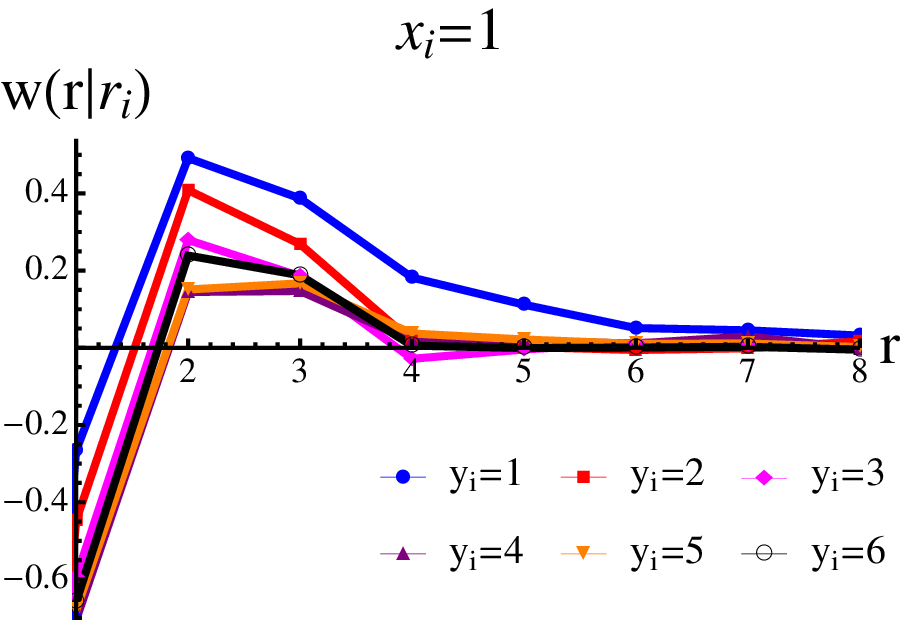}
  \includegraphics[width=0.40\columnwidth]{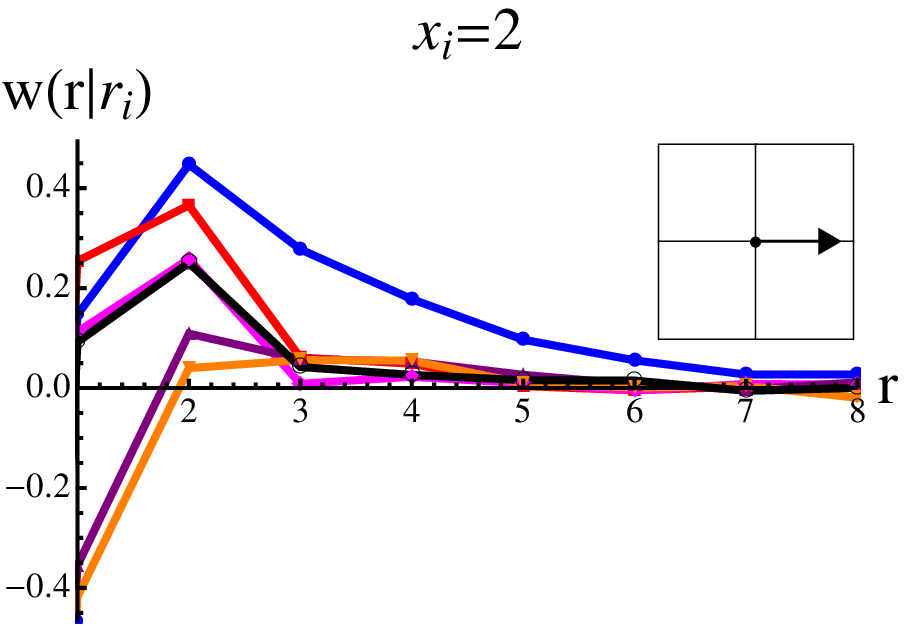}
  \includegraphics[width=0.40\columnwidth]{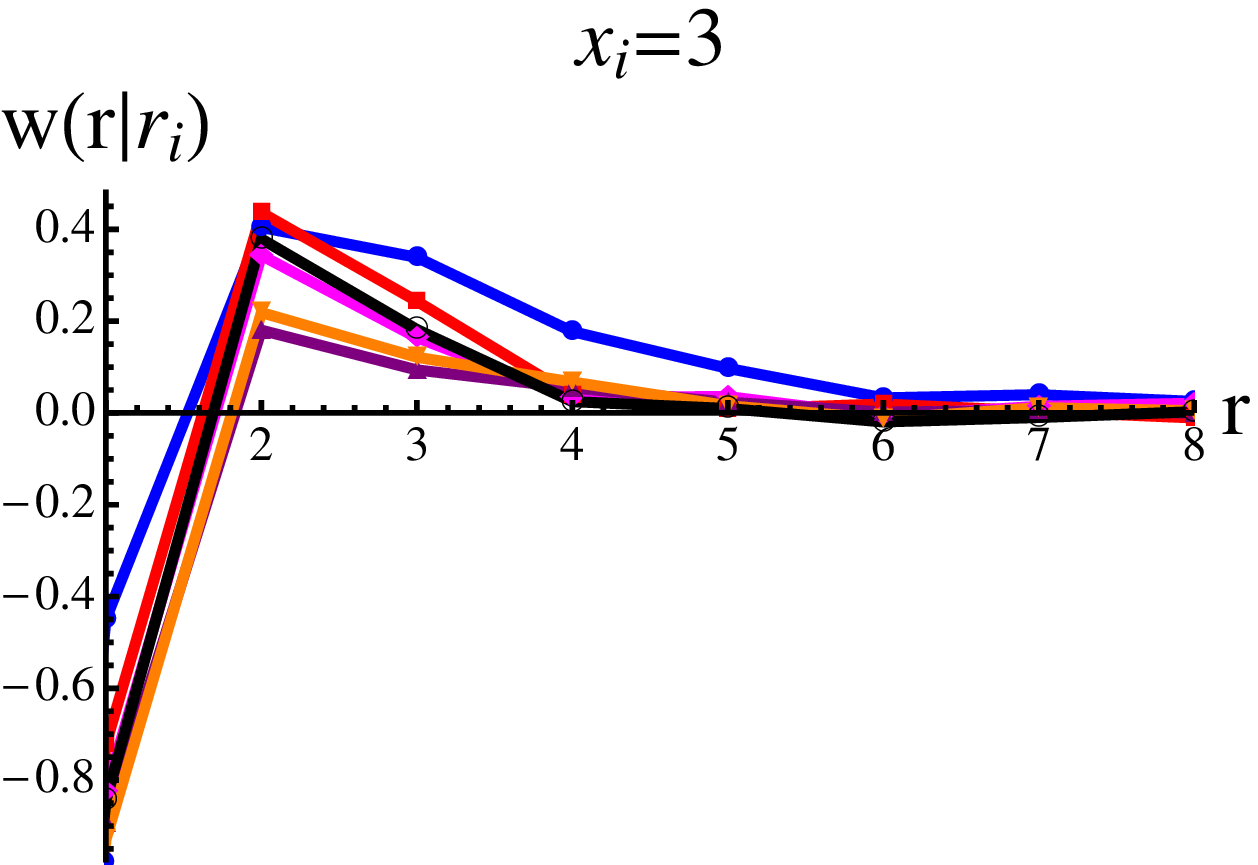}
  \includegraphics[width=0.40\columnwidth]{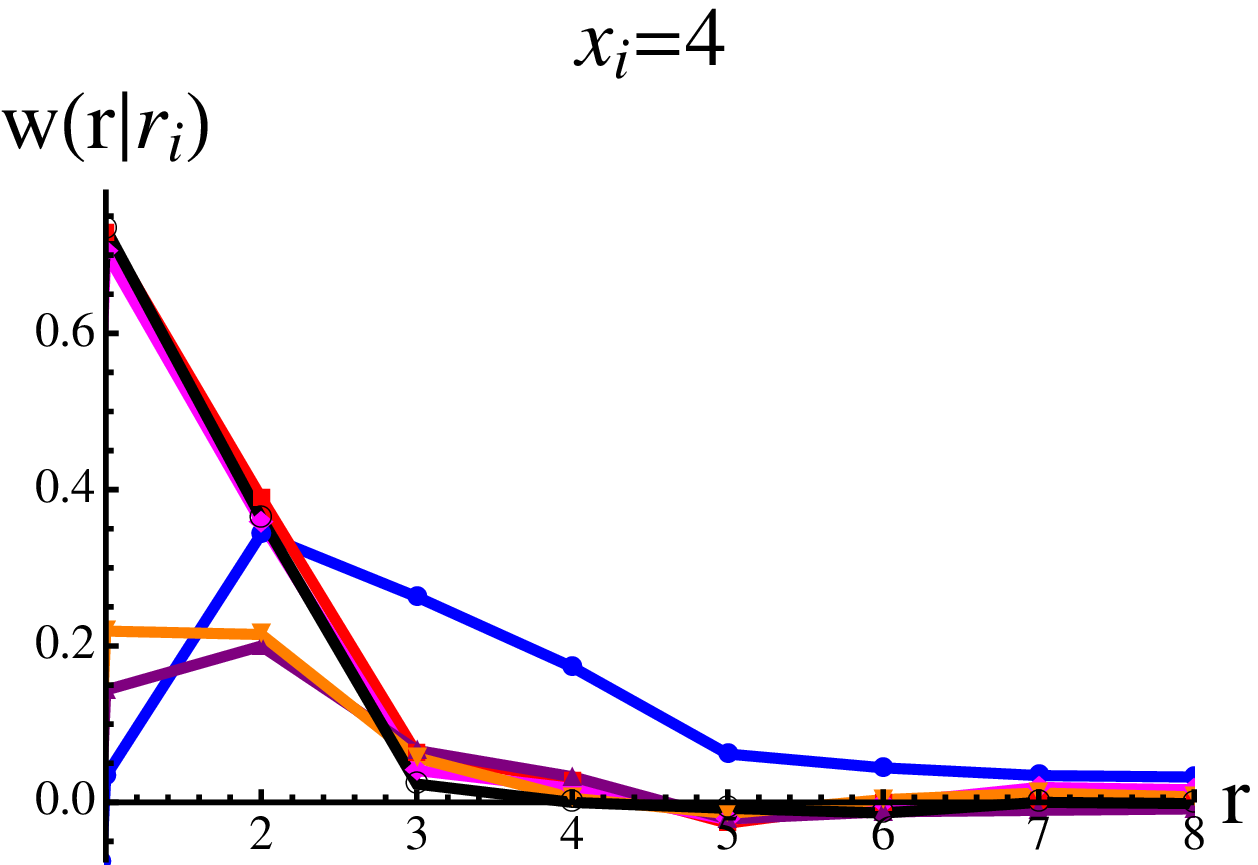}
 \caption{(Color online) Plots of interactions $w(r|\V{r}_i)$ against distance $r=|x_j-x_i|$
 in a common row $y_i=y_j$ as   the origin of the plot
 $\V{r}_i=(x_i,y_i)$ changes
 (the top-right inset represents the moving direction of
$\V{r}_{j}$).
 The interactions are inferred by the NMF using  $B=576,096$ patches of size $L=16$ created from face pictures. }
\Lfig{L=16-faces-NMF-basic}
\end{center}
\end{figure}
Again, we see that the inferred interactions decrease around
$r\approx 4$. The boundary effect at $y_i=1$ is also present and
the NN interactions appear to have  a periodicity similar to that
seen in \Rfig{sublattice}. However, a new observation in
\Rfig{L=16-faces-NMF-basic} is
\begin{itemize}
\item{Some NN interactions take negative values.}
\end{itemize}
As noted above, these antiferromagnetic interactions are
considered to emerge for expressing patterns where black and white
pixels alternatively appear, which are presumably produced by a
dither process to discriminate the face  from other regions.

The orientation dependency of the inferred interactions is
examined by observing $w(r|\V{r}_i)$ along the downward slope as
in \Rfig{L=16-aerials-NMF-ds}. The results are shown in
\Rfig{L=16-faces-NMF-ds}.
\begin{figure}[htbp]
\begin{center}
  \includegraphics[width=0.40\columnwidth]{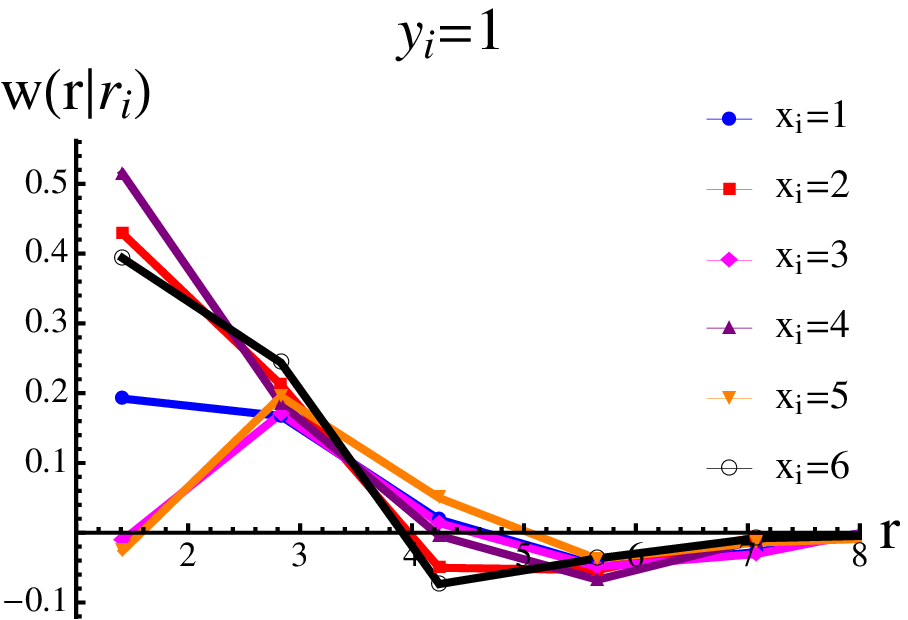}
  \includegraphics[width=0.40\columnwidth]{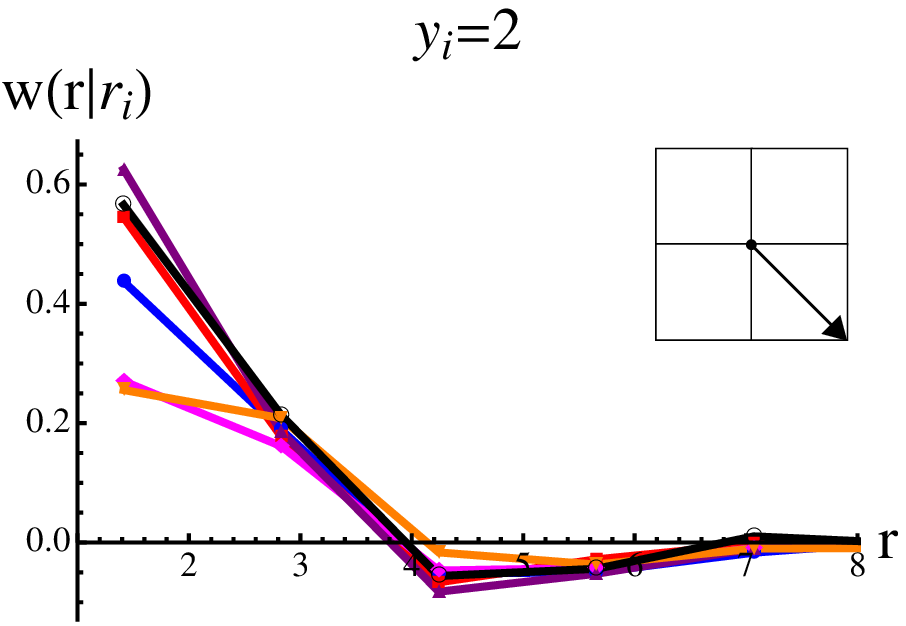}
  \includegraphics[width=0.40\columnwidth]{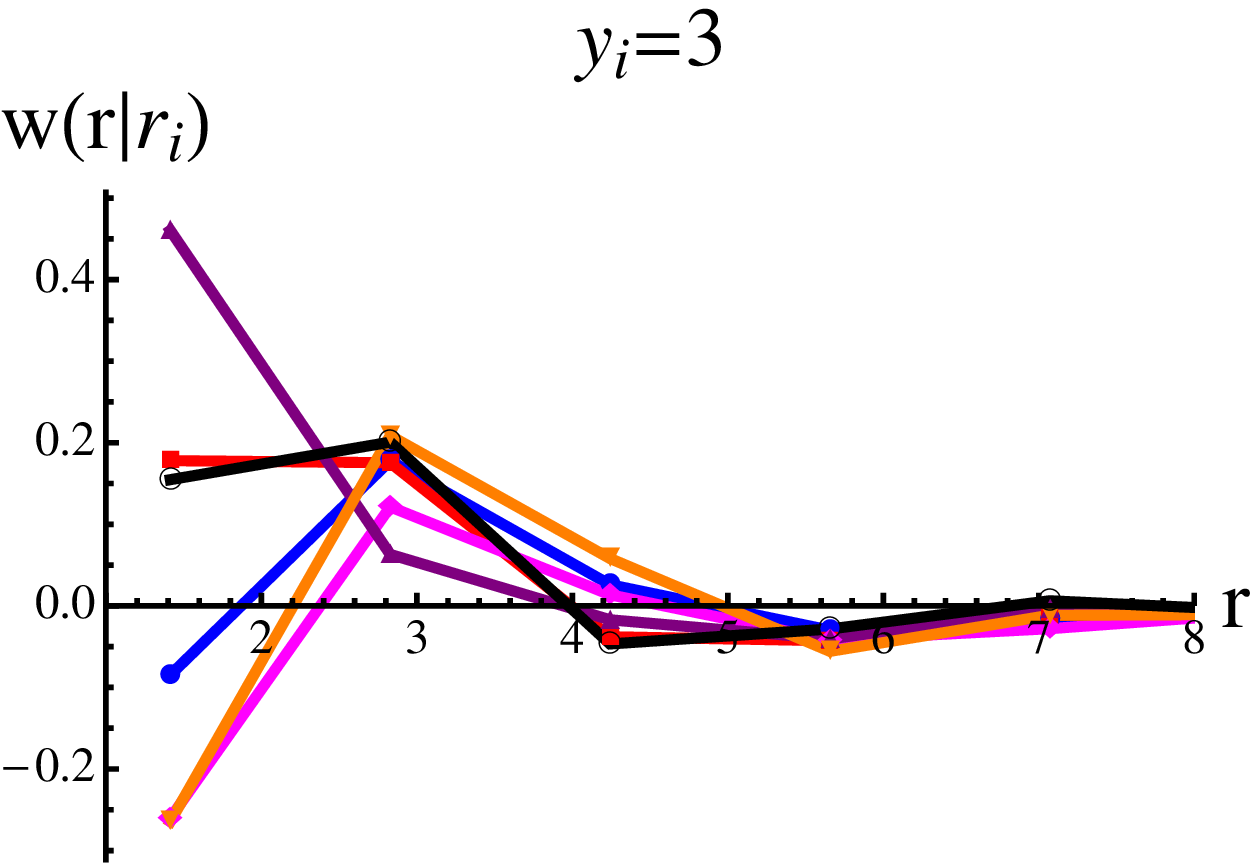}
  \includegraphics[width=0.40\columnwidth]{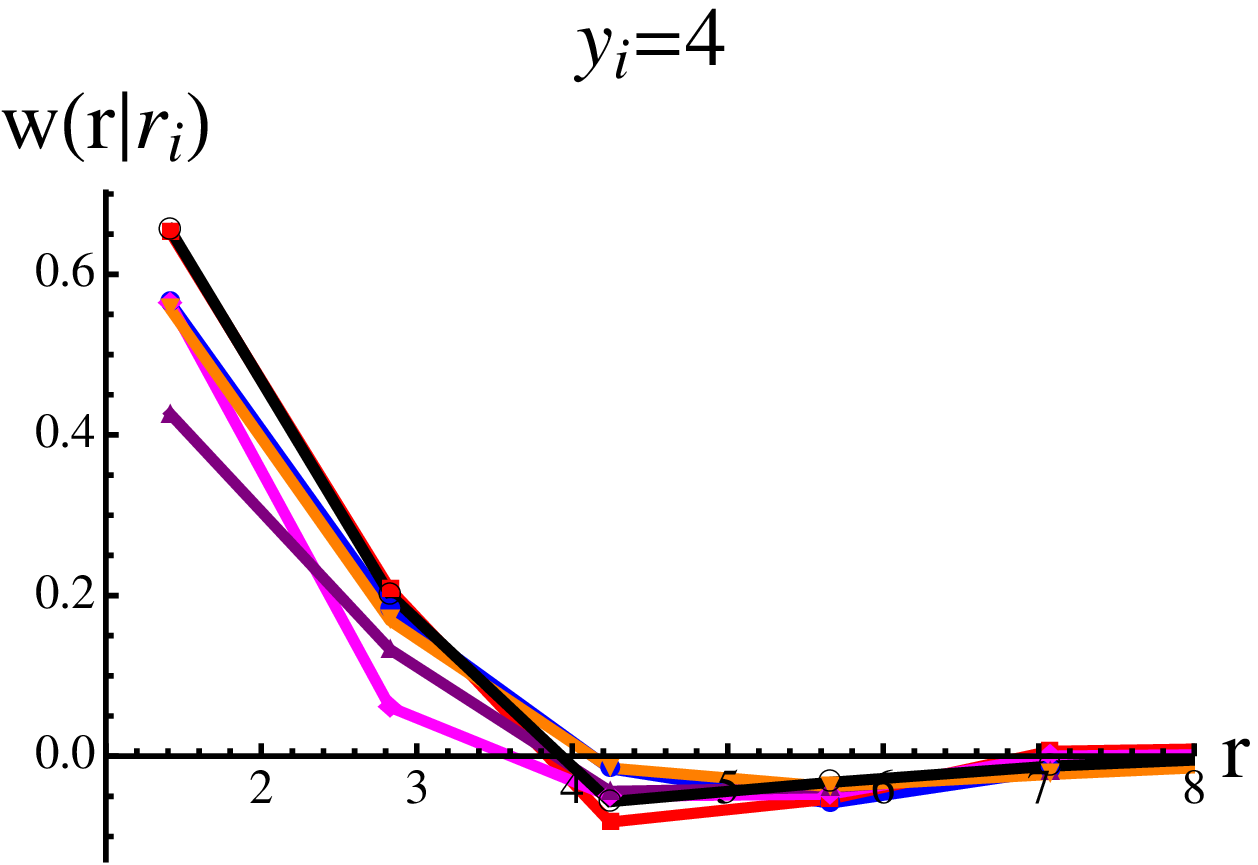}
 \caption{(Color online) Plots of interactions $w(r|\V{r}_i)$ against distance $r_{ij}=\sqrt{2}s$ with $s=1,2,\cdots 6$
 along the downward slope of the 45-degree angle from fixed $\V{r}_i$.
 The same  images and  parameters as in \Rfig{L=16-faces-NMF-basic} are used.}
\Lfig{L=16-faces-NMF-ds}
\end{center}
\end{figure}
New observations are:
\begin{itemize}
\item{Some NNN interactions also take negative values.} 
\item{A periodicity in the NNN interactions is present. Namely, the behavior of the NNN interactions at $y_i=1$ and $3$  is similar, while it differs from those at $y_i=2$ and $4$ with similar values.}
\end{itemize}
To visualize the periodicities of the NN and NNN interactions, we
show graph representations of these interactions in
\Rfig{L=16-faces-NMF-NNandNNN}, employing the fact that the signs
of the interactions are different among the sublattices, namely,
the positive and negative interactions are colored by blue and red
links, respectively.
\begin{figure}[htbp]
\begin{center}
  \includegraphics[width=0.4\columnwidth]{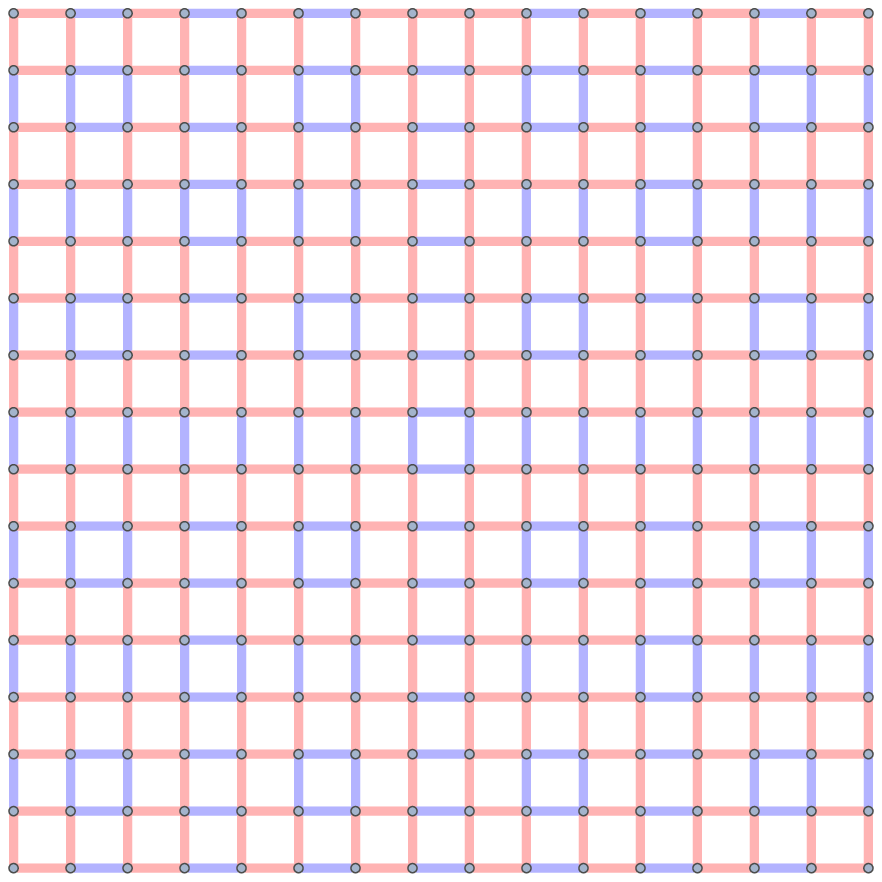}
  \includegraphics[width=0.4\columnwidth]{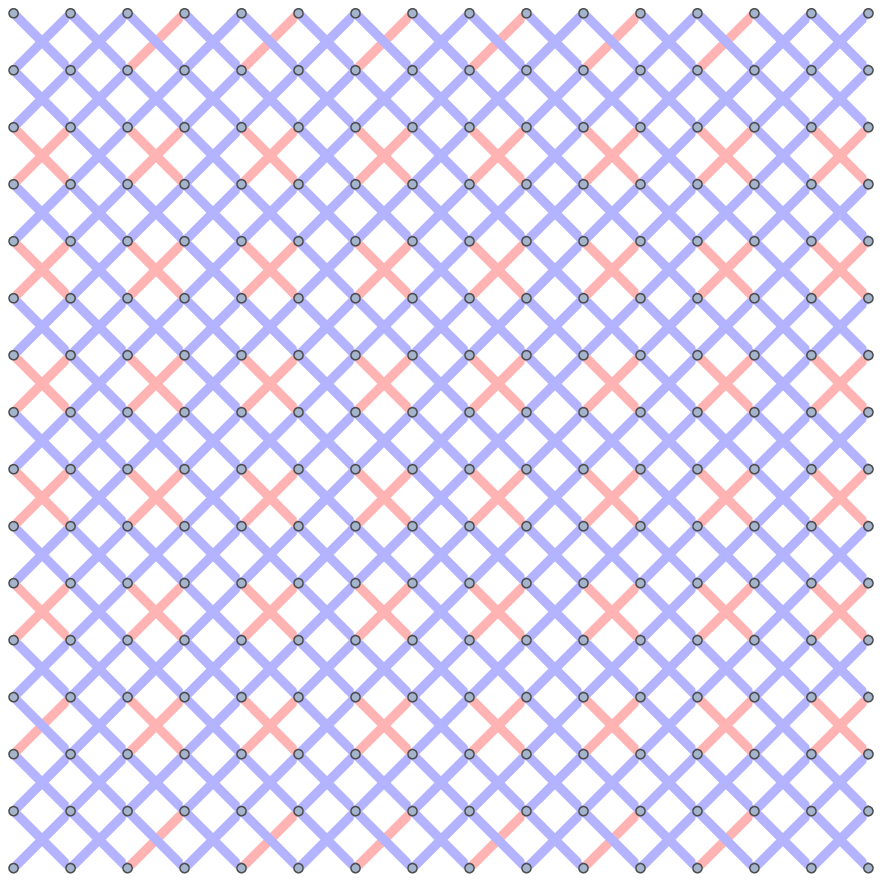}
 \caption{(Color online) Graph representations of the NN (left) and NNN (right) interactions
 inferred from the face pictures of patch size $L=16$ plotted in the coordinate space $\V{r}=(x,y)$.
 Each site corresponds to each pixel location and the links correspond to the interactions between the sites.
 Blue and red links denote positive and negative interactions, respectively.
 Clear periodicity and a checker board-like structure are observed.}
\Lfig{L=16-faces-NMF-NNandNNN}
\end{center}
\end{figure}
\Rfig{L=16-faces-NMF-NNandNNN} clearly exhibits the periodicity of the NN and NNN interactions
 and we can see  that a checker board-like structure in the interacting network emerges,
 which is consistent with the sublattice structure shown in the left and center panels of \Rfig{sublattice}.
 Deviations from the checker board structure are also observed,
 which introduce frustration into the system. However, the number of frustrated plaquettes is not large
 and we expect the effect of the frustration to be small, meaning the nature of the ground state is simple.
 If we take into account the NNN interactions, frustration can be enhanced by the antiferromagnetic NNN interactions
 in plaquettes consisting of four antiferromagnetic NN interactions,
 but the effect on the ground state is again expected to be weak, since these antiferromagnetic NNN interactions
 are small in absolute value as compared to other NN and NNN interactions,
 which is confirmed in the histogram of the NN and NNN interactions,
 $P(w_{ij}|r_{ij}=1)$ and $P(w_{ij}|r_{ij}=\sqrt{2})$, given in \Rfig{L=16-faces-NMF-Hist}.
\begin{figure}[htbp]
\begin{center}
  \includegraphics[height=0.28\columnwidth]{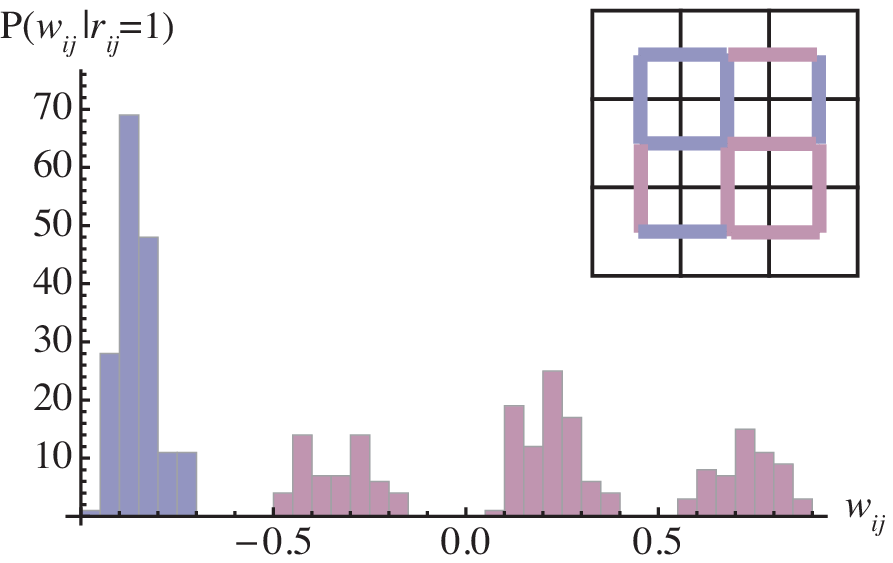}
  \includegraphics[height=0.28\columnwidth]{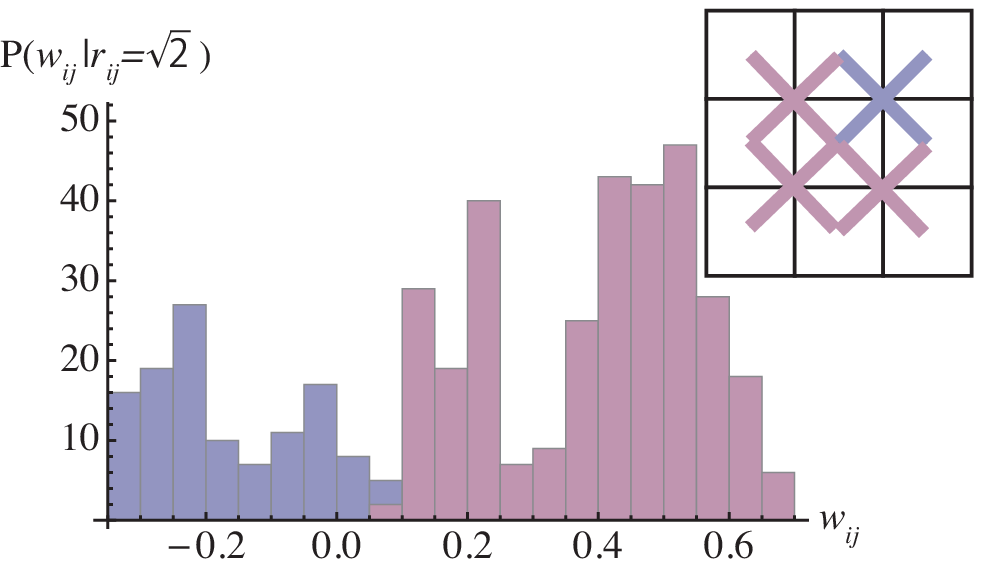}
 \caption{(Color online) Histograms of the NN (left) and NNN (right) interactions derived by the NMF
 from face pictures of patch size $L=16$. Each histogram consists of two categories
 corresponding to the sublattice structure, which is shown in the inset.
 The colors of the histograms correspond to the two different types of links in the inset.}
\Lfig{L=16-faces-NMF-Hist}
\end{center}
\end{figure}
The histograms clearly reflect the sublattice structure shown in
\Rfig{sublattice}. The insets show the reduced versions of the
left and center panels of \Rfig{sublattice}. Multiple peaks
observed in the NN interactions, the magenta part in the left panel of \Rfig{L=16-faces-NMF-Hist}, imply another additional
periodicity in the NN interactions, but we do not pursue this
point to avoid complexity, as we declared in
\Rsec{Interactions-aerial}.

Although the frustration is weak and possibly does not affect the
ground state, some metastable states can emerge because of the
frustration and can influence the nature of the system.  In
\cite{Stephens:13}, the role of these metastable states was
discussed in connection with biological visual systems, based on
observations that the patterns of the metastable states can be
interpreted as filters selecting certain directions of edges
discriminating two uniform regions. This may in fact be
interesting, but we do not pursue this point since the enumeration
of the metastable states is not easy in our case, because the
system size is significantly larger than that in
\cite{Stephens:13}; however,  additional remarks are presented
in \Rsec{Discussion}.

Further, we examine the finite size effects and the difference between the NMF and BA. For this, we plot
$\overline{w_{A}}(r) $ and $\overline{w_{B}}(r)$, the definitions
of which are the same as in the case of aerial pictures, for
different patch sizes $L=8,16,$ and $32$ in \Rfig{sizecomp-faces}.
\begin{figure}[htbp]
\begin{center}
  \includegraphics[width=0.4\columnwidth]{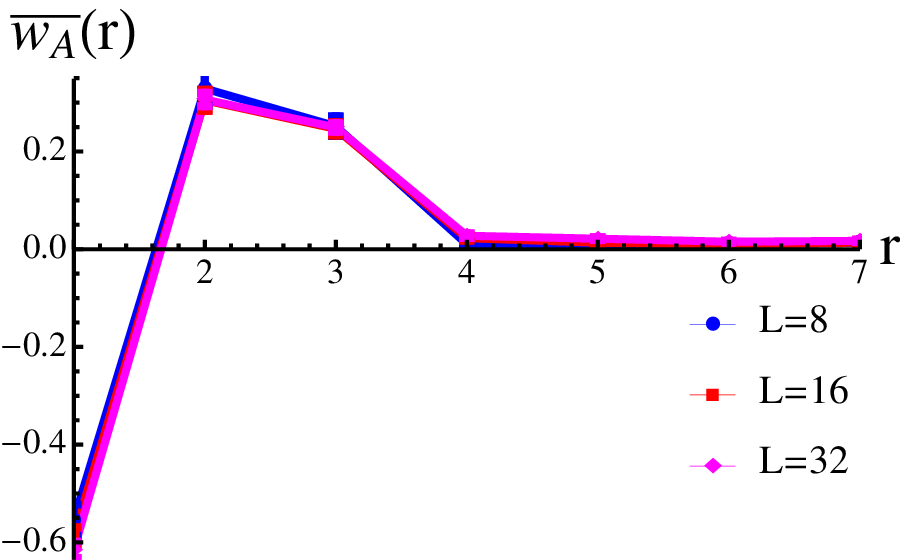}
  \includegraphics[width=0.4\columnwidth]{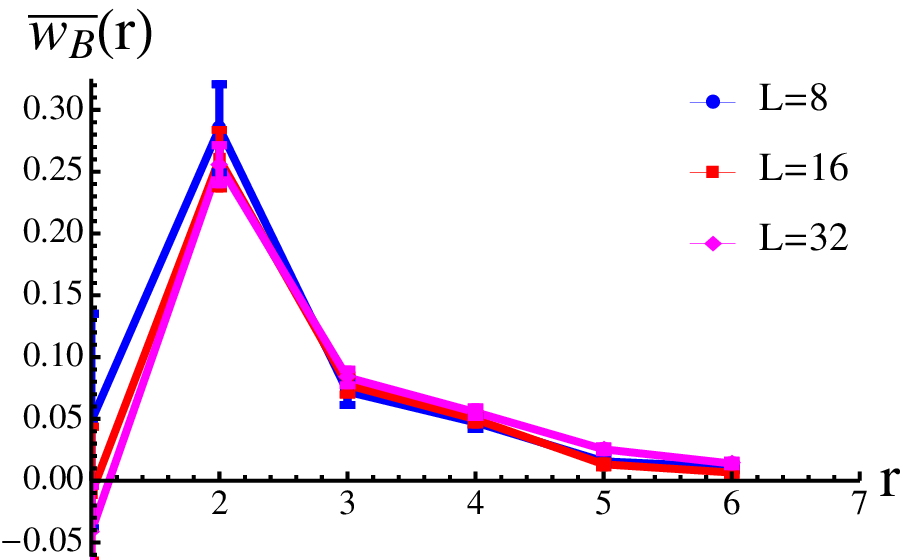}
 \caption{(Color online) Plots of $\overline{w_{A}}(r)$ (left) and $\overline{w_{B}}(r) $ (right)
 for different patch sizes $L=8,$ $16,$ and $32$. The number of  used patches  is $B=1,536,384$ and $174,096$
 for $L=8$ and $32$, respectively. The size effect is quite tiny.}
\Lfig{sizecomp-faces}
\end{center}
\end{figure}
This figure shows that the size effect is again absent and the
characteristic length scale is about $\xi\approx 4$. Similar plots
to compare the NMF and BA are given in \Rfig{NMFBAcomp-faces}.
\begin{figure}[htbp]
\begin{center}
  \includegraphics[width=0.40\columnwidth]{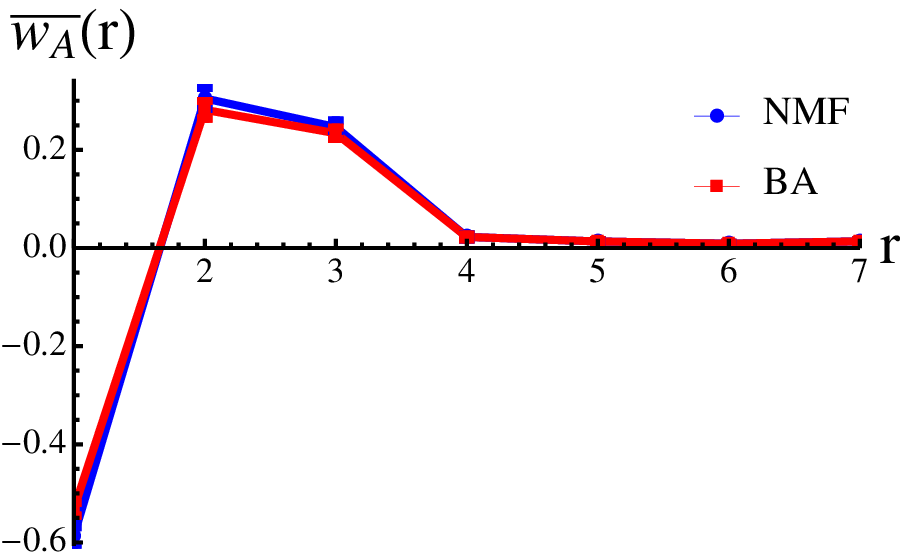}
  \includegraphics[width=0.40\columnwidth]{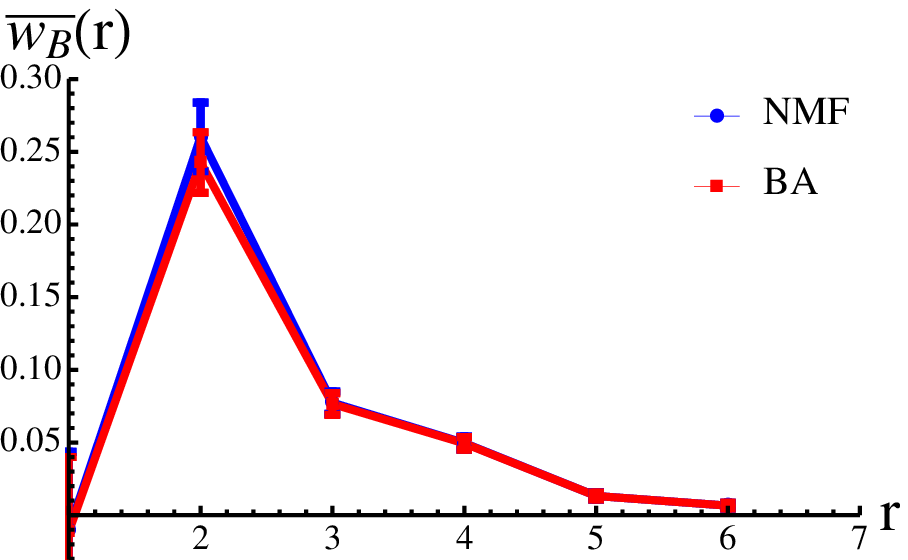}
 \caption{(Color online) Comparison in $\overline{w_{A}}(r)$ (left) and $\overline{w_{B}}(r)$ (right)
 between the NMF and BA for patch size $L=16$. The difference is negligible.
}
\Lfig{NMFBAcomp-faces}
\end{center}
\end{figure}
We again see that almost no difference exists between the NMF and
BA. Thus, the findings by the NMF for patch size $L=16$ are
expected to hold.

\subsubsection{Forest Pictures}\Lsec{Interactions-forest}
Let us move to the case of the forest pictures. The interactions in a row inferred by the NMF are plotted in
\Rfig{L=16-forestneedles-NMF-basic}, as \Rfig{L=16-aerials-NMF-basic} and \Rfig{L=16-faces-NMF-basic}.
\begin{figure}[htbp]
\begin{center}
  \includegraphics[width=0.40\columnwidth]{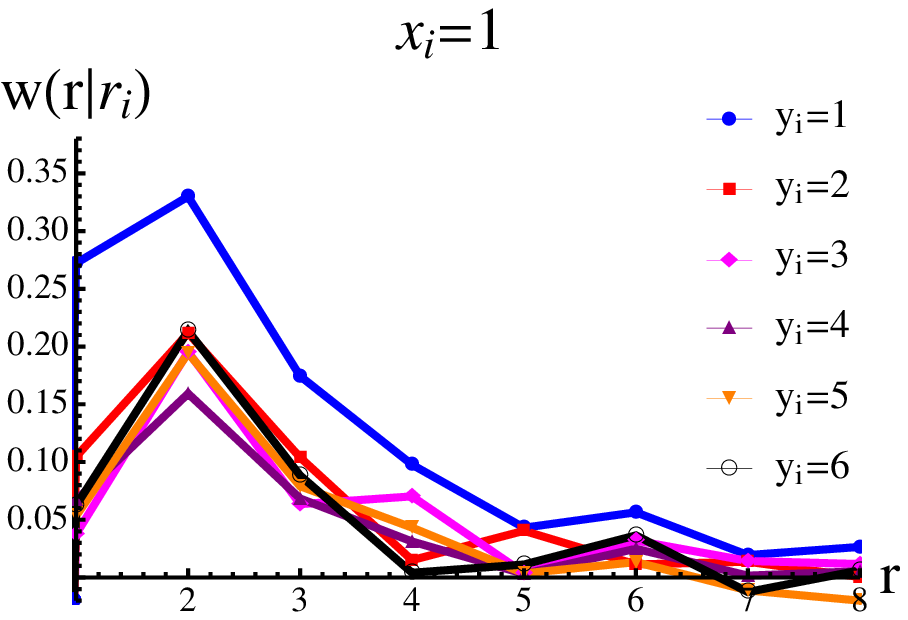}
  \includegraphics[width=0.40\columnwidth]{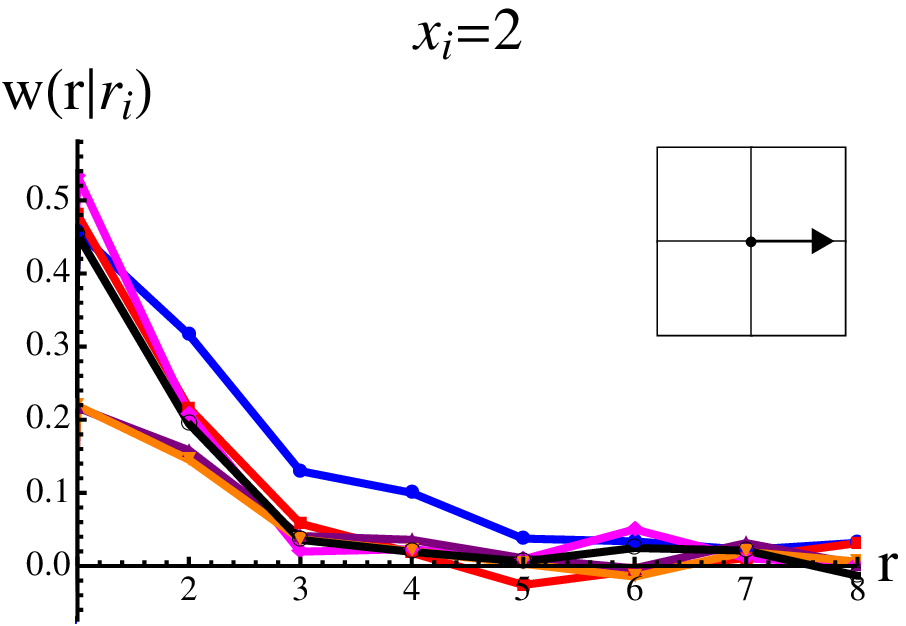}
  \includegraphics[width=0.40\columnwidth]{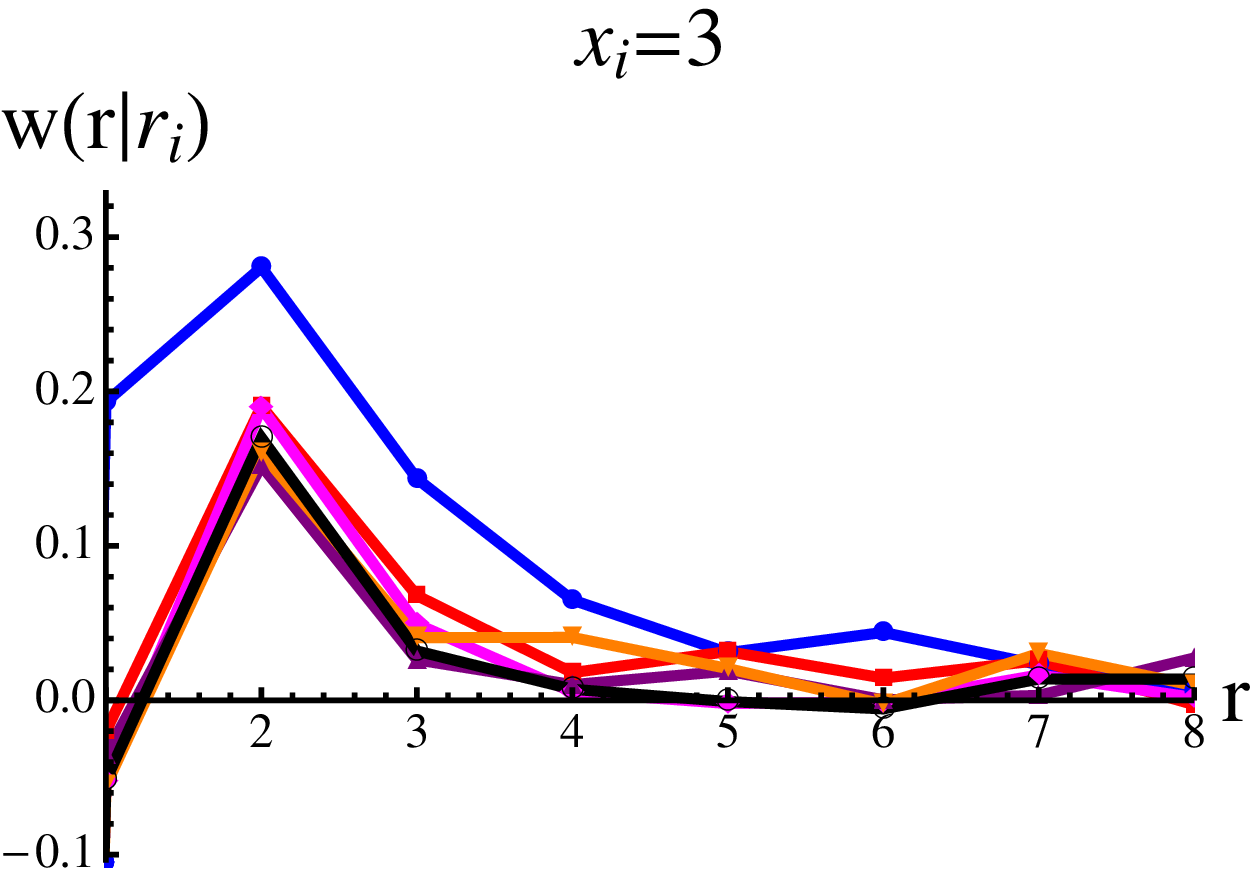}
  \includegraphics[width=0.40\columnwidth]{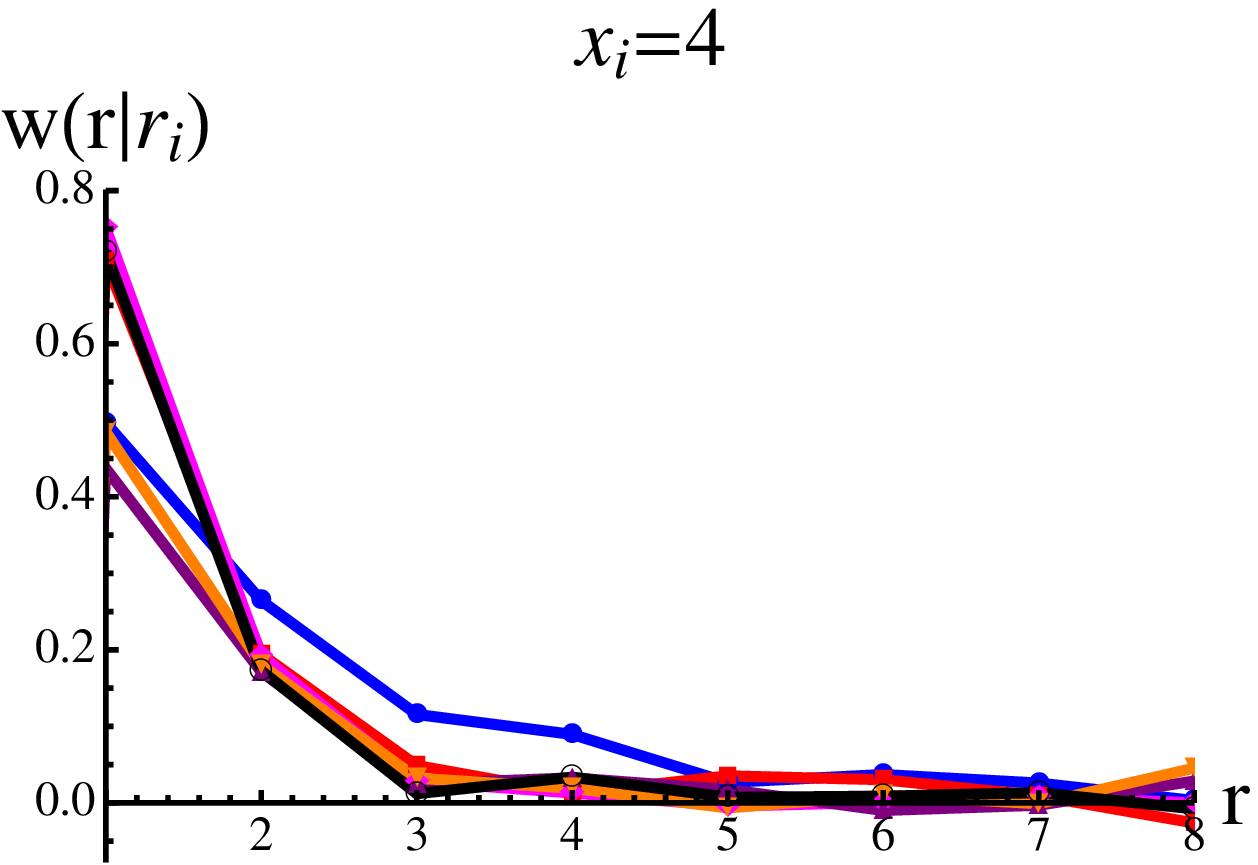}
 \caption{(Color online) Plots of interactions $w(r|\V{r}_i)$ against distance $r=|x_j-x_i|$
 in a common row $y_i=y_j$ as  the origin of the plot
 $\V{r}_i=(x_i,y_i)$ changes
 (the top-right inset represents the moving direction of $\V{r}_{j}$).
 The interactions are inferred by the NMF using  $B=576,096$ patches of size $L=16$ created from forest pictures. }
\Lfig{L=16-forestneedles-NMF-basic}
\end{center}
\end{figure}
The interaction range, boundary effect, and periodicities are
common, as in the previous two cases. As in
\Rfigs{L=16-aerials-NMF-ds}{L=16-faces-NMF-ds}, the interactions
along the downward slope of the forest pictures are displayed in
\Rfig{L=16-forestneedles-NMF-ds}.
\begin{figure}[htbp]
\begin{center}
  \includegraphics[width=0.40\columnwidth]{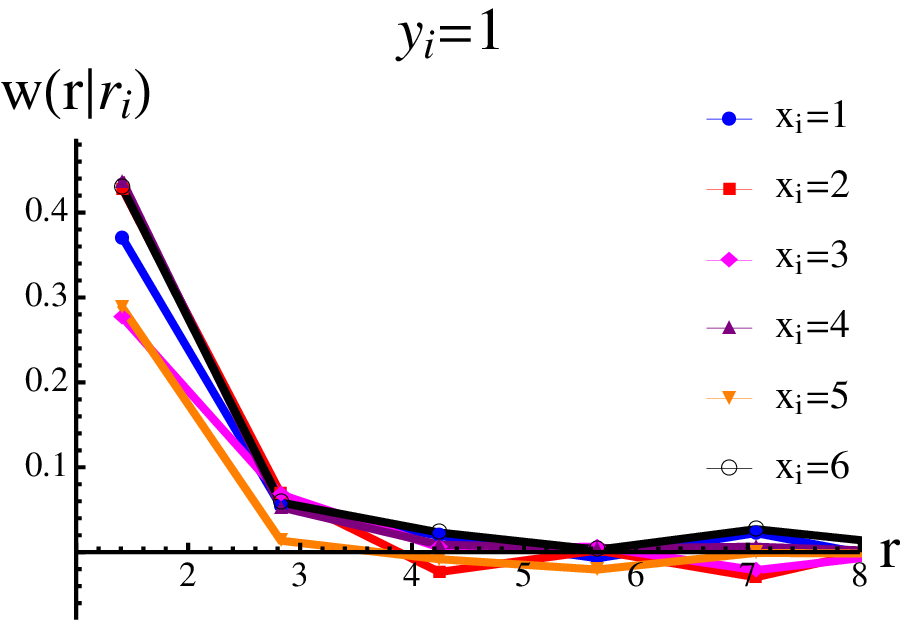}
  \includegraphics[width=0.40\columnwidth]{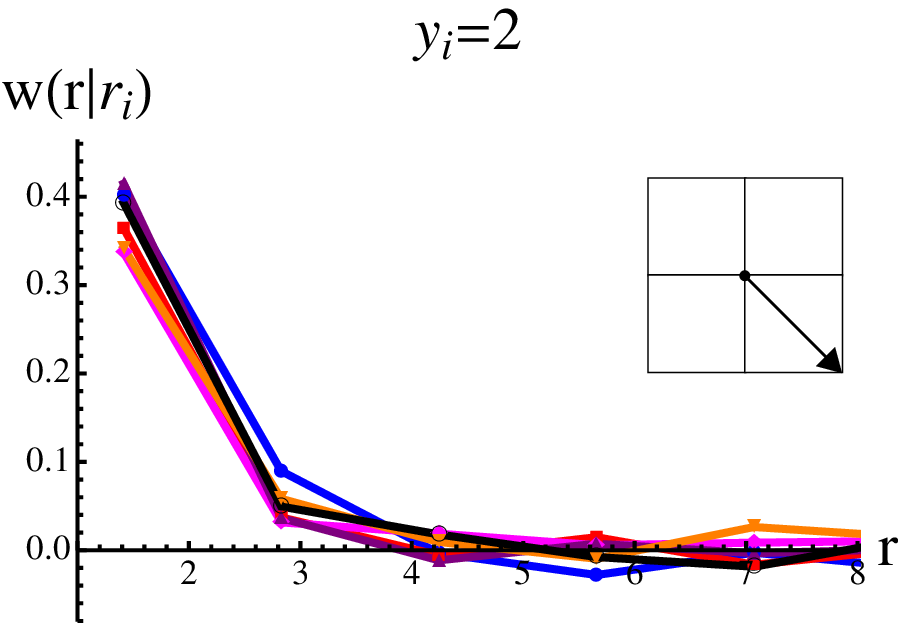}
  \includegraphics[width=0.40\columnwidth]{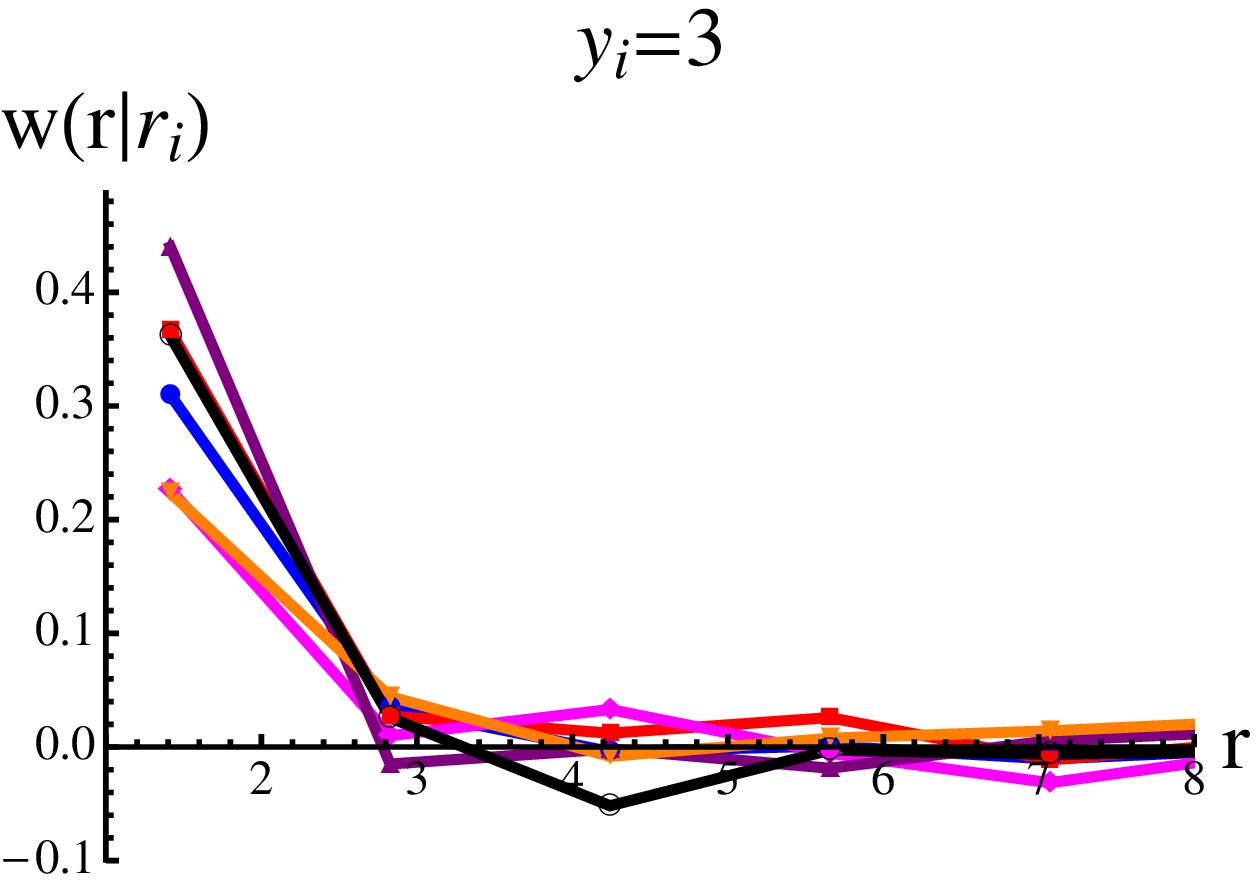}
  \includegraphics[width=0.40\columnwidth]{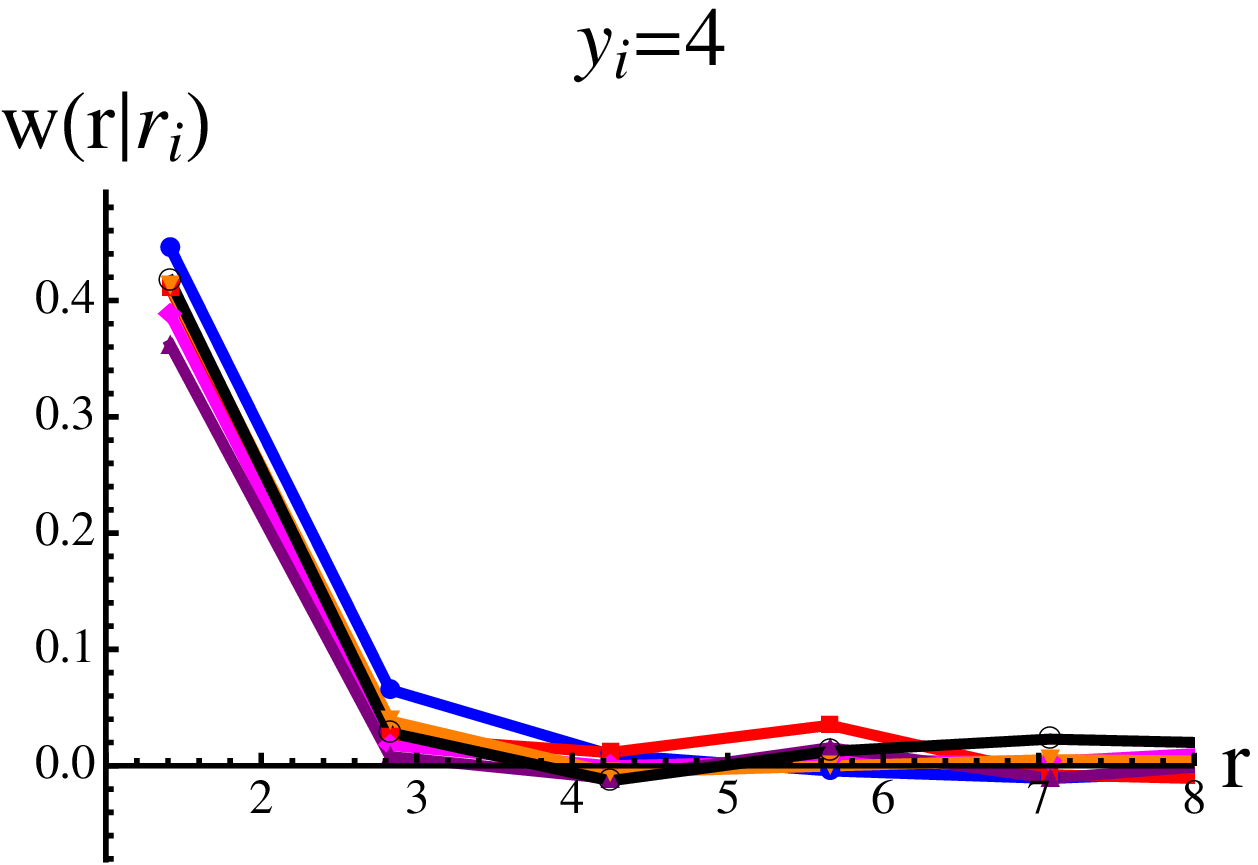}
 \caption{(Color online) Plots of interactions $w(r|\V{r}_i)$ against distance $r_{ij}=\sqrt{2}s$ with $s=1,2,\cdots 6$
 along the downward slope of the 45-degree angle from fixed $\V{r}_i$
 (the top-right inset represents the moving direction of $\V{r}_{j}$). The same images and  parameters
 as in \Rfig{L=16-forestneedles-NMF-basic} are used.}
\Lfig{L=16-forestneedles-NMF-ds}
\end{center}
\end{figure}
The behavior is similar to that in the aerial picture case and the
periodicity is not clearly seen in the NNN interactions. The graph
representations of the NN and NNN interactions are shown in
\Rfig{L=16-forestneedles-NMF-NNandNNN}, corresponding to
\Rfig{L=16-faces-NMF-NNandNNN}.
\begin{figure}[htbp]
\begin{center}
  \includegraphics[width=0.4\columnwidth]{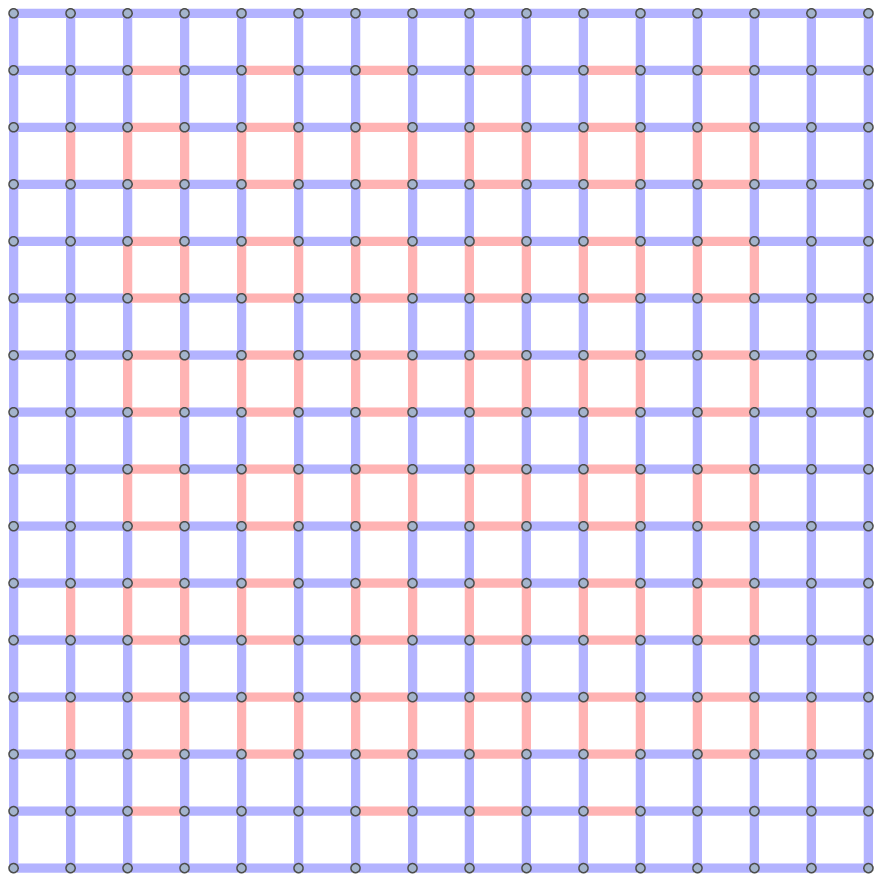}
  \includegraphics[width=0.4\columnwidth]{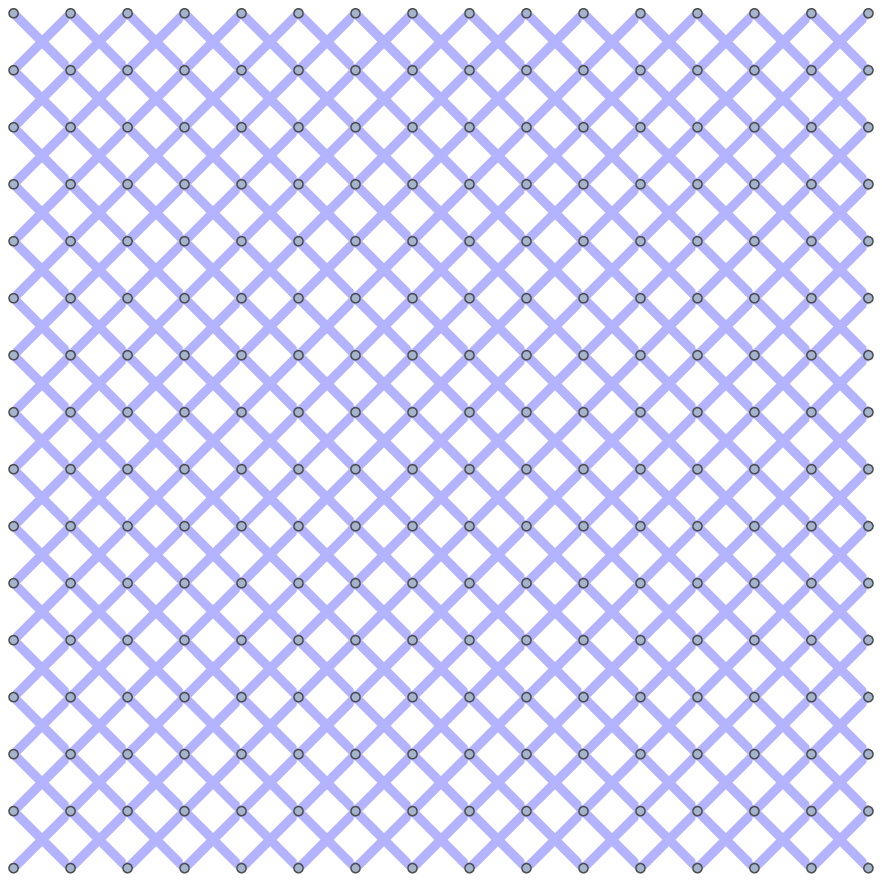}
 \caption{(Color online) Graph representations of the NN (left) and NNN (right) interactions inferred from the forest pictures
 of size $L=16$ plotted in the coordinate space $\V{r}=(x,y)$.
 Checker  board-like structures are observed in the NN interaction network, except for in the boundaries. }
\Lfig{L=16-forestneedles-NMF-NNandNNN}
\end{center}
\end{figure}
The checker board-like structure is again observed in the NN, but
not in the NNN interaction network, indicating the absence of
frustration. Quantitative information about the NN and NNN
interactions is obtained from the histograms in
\Rfig{L=16-forestneedles-NMF-Hist}.
\begin{figure}[htbp]
\begin{center}
  \includegraphics[width=0.4\columnwidth]{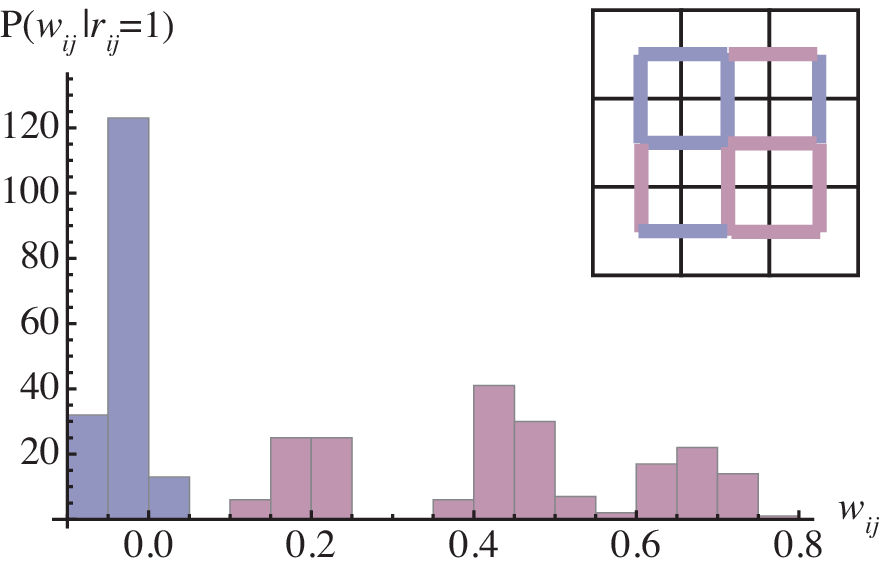}
  \includegraphics[width=0.4\columnwidth]{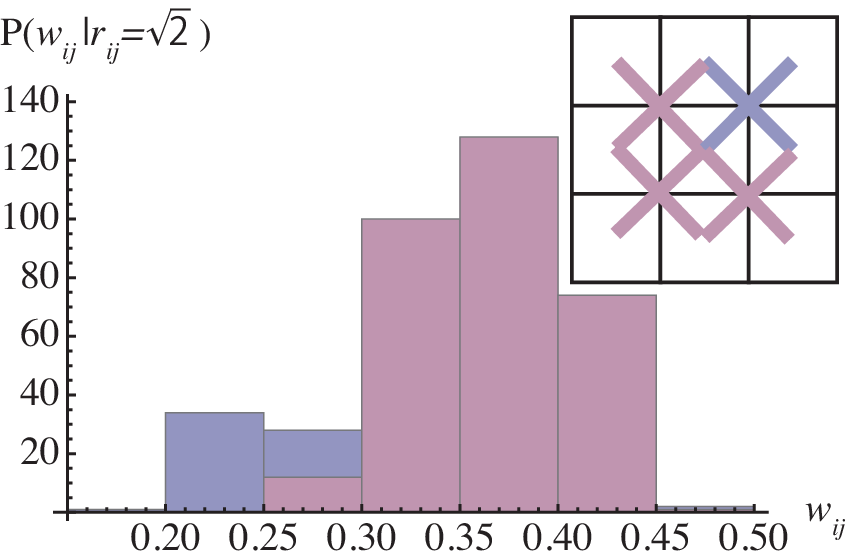}
 \caption{(Color online) Histograms of the NN (left) and NNN (right)
 interactions derived by the NMF from forest pictures of patch size $L=16$.
The interactions are characterized  based on \Rfig{sublattice},
but no clear difference is observed in the NNN interactions.  }
\Lfig{L=16-forestneedles-NMF-Hist}
\end{center}
\end{figure}
The sublattice structure given in \Rfig{sublattice} is again
observed in the NN interactions, including the additional
periodicity, in common with the face pictures, signaled by the
multiple peaks in the histogram.

The finite-size effect is examined in \Rfig{sizecomp-forestneedles}.
\begin{figure}[htbp]
\begin{center}
  \includegraphics[angle=-90,width=0.40\columnwidth]{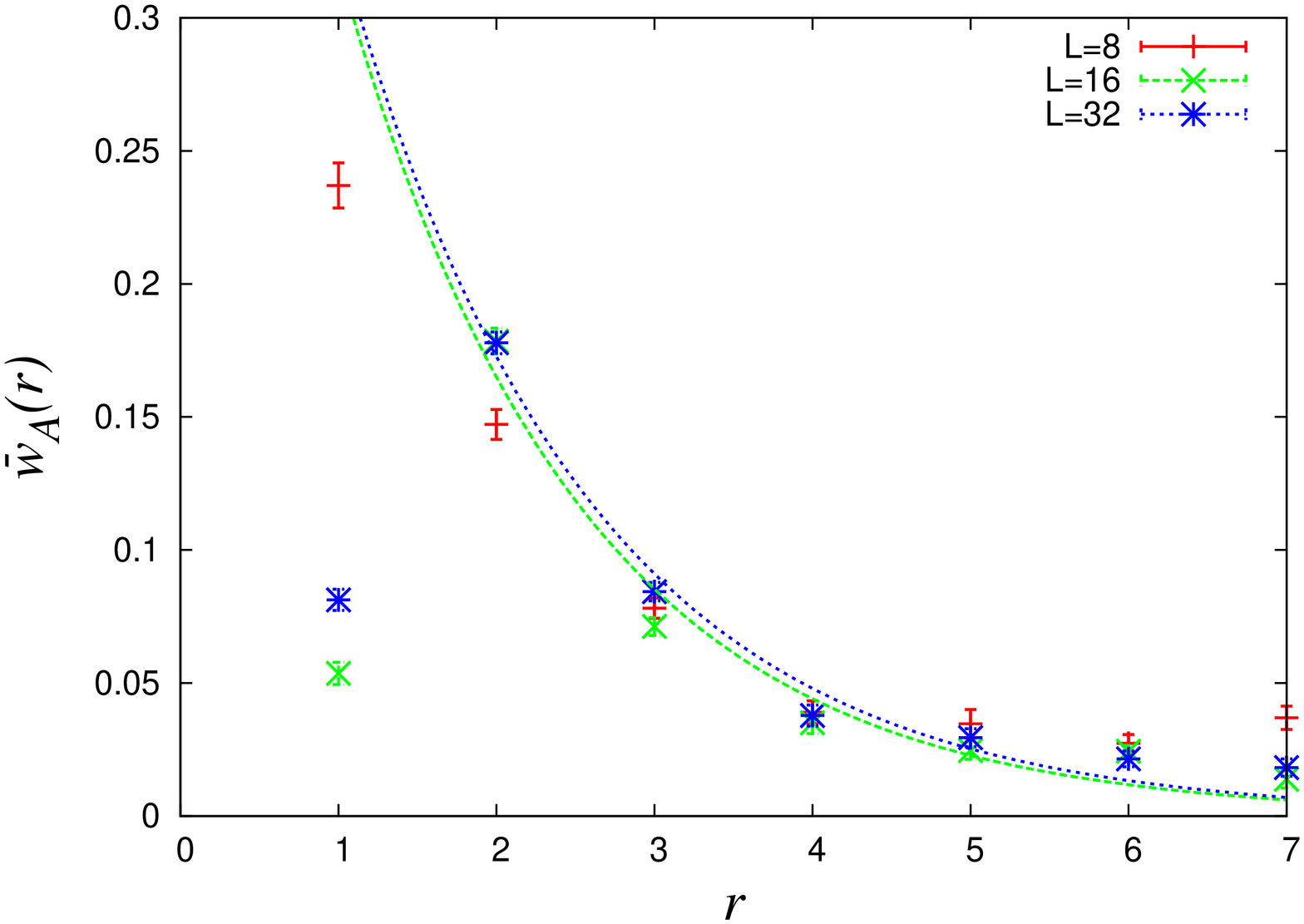}
  \includegraphics[angle=-90,width=0.40\columnwidth]{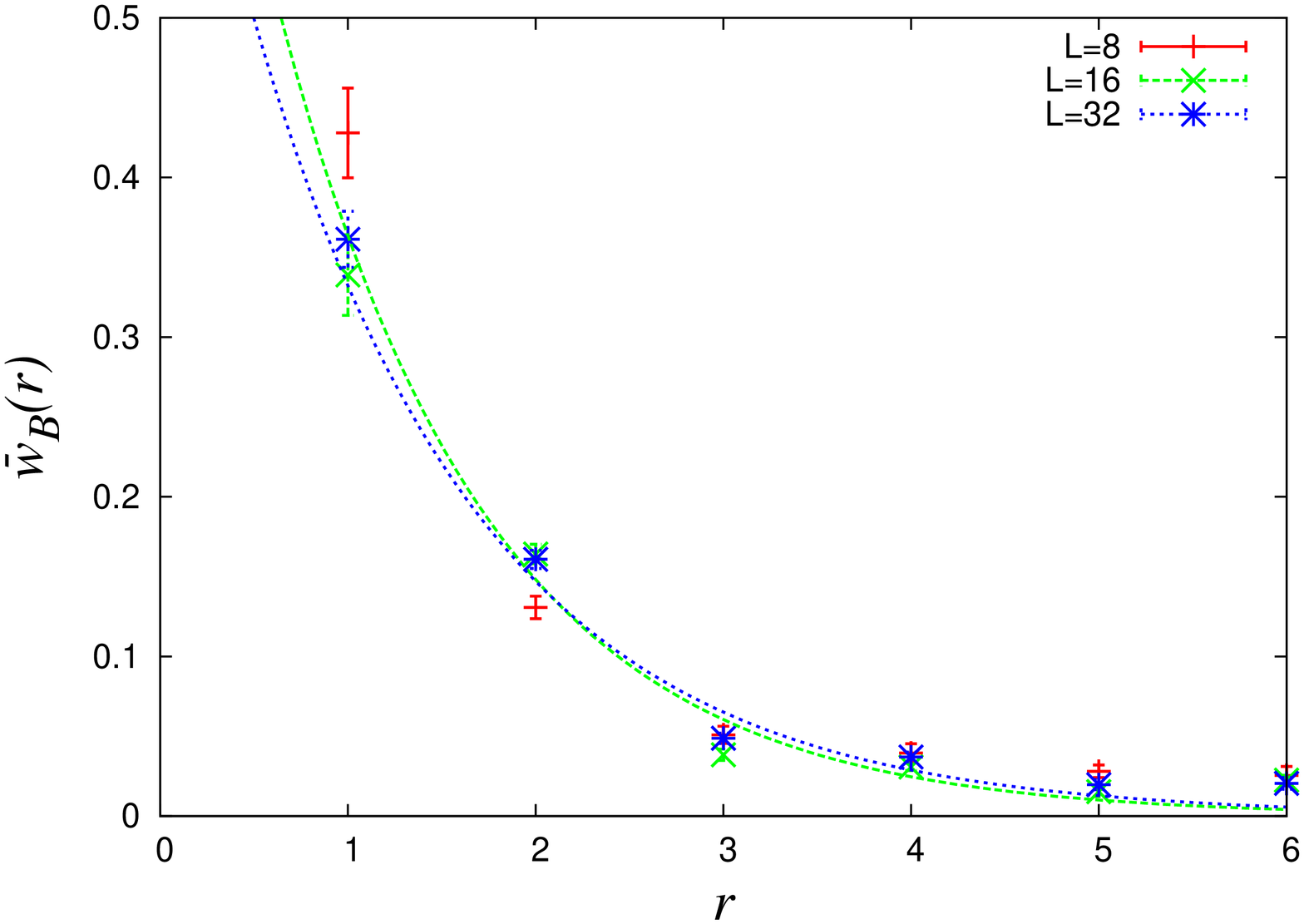}
 \caption{(Color online) Plots of $\overline{w_{A}}(r)$ (left) and $\overline{w_{B}}(r) $ (right)
 for different patch sizes $L=8,$ $16,$ and $32$. The number of used patches is $B=1,536,384$ and $174,096$ for $L=8$ and $32$, respectively. The interactions tend to be longer range than in the cases of aerial and face pictures, which appear to lead to a finite size effect at $L=8$. The curves of the
 exponential fit to the region $2 \leq r\leq 6$ are given for $L=16$ and $32$. }
\Lfig{sizecomp-forestneedles}
\end{center}
\end{figure}
In comparison with the previous two cases, we can see that the
curves of $\overline{w_{A}}(r)$ and $\overline{w_{B}}(r)$ do not
disappear completely around $r\approx 4$, and appear to be longer
ranged. A finite-size effect is observed at $L=8$, where the
values of interactions at $r=1$ and $7$ are larger than for the
other two patch sizes, $L=16$ and $32$. These facts imply the
range of interactions for the forest pictures is  longer than in
the cases of aerial and face pictures. To examine this point, we attempt the following exponential fit with two parameters $a$ and $b$: \be \overline{w}(r)=a e^{-(r-2)/b}. \Leq{exp} \ee The
fitting is performed against the region $2 \leq r \leq 6$ for
$L=16$ and $32$;  the resultant curves are given in
\Rfig{sizecomp-forestneedles}. The estimated parameters are
summarized in Table \ref{tab:fitting-forests}.
\begin{table}[htbp]
\begin{center}
\caption{Parameters fitted to \Req{exp} for forest pictures
corresponding to \Rfig{sizecomp-forestneedles}}
\begin{tabular}{| l | l | l | l | l | l |}
\hline
\hline
Sublattice & Size $L$ & $a$ & $b$ \\
\hline
\hline
A & 16 & $0.16 \pm 0.02$ & $1.52\pm 0.25$ \\
\hline
B & 16 & $0.15 \pm 0.03$ & $ 1.11 \pm 0.28$ \\
\hline
A & 32 & $0.17 \pm 0.01$ & $ 1.56 \pm  0.17$ \\
\hline
B & 32 & $0.15 \pm 0.03$ & $ 1.23 \pm 0.28$ \\
\hline
\end{tabular}
\label{tab:fitting-forests}
\end{center}
\end{table}
At least visually, the exponential fit is good, and the
estimated parameter $b$ takes values around $1.1$-$1.6$. This
implies that the interactions in this case are seemingly
longer-reached but still rapidly decaying, and it suffices to
consider the range $r \leq \xi = 2+b\approx 4$, which is in
accordance with the cases of aerial and face pictures.

A comparison of the NMF and BA is given in
\Rfig{NMFBAcomp-forestneedles}.
\begin{figure}[htbp]
\begin{center}
  \includegraphics[width=0.40\columnwidth]{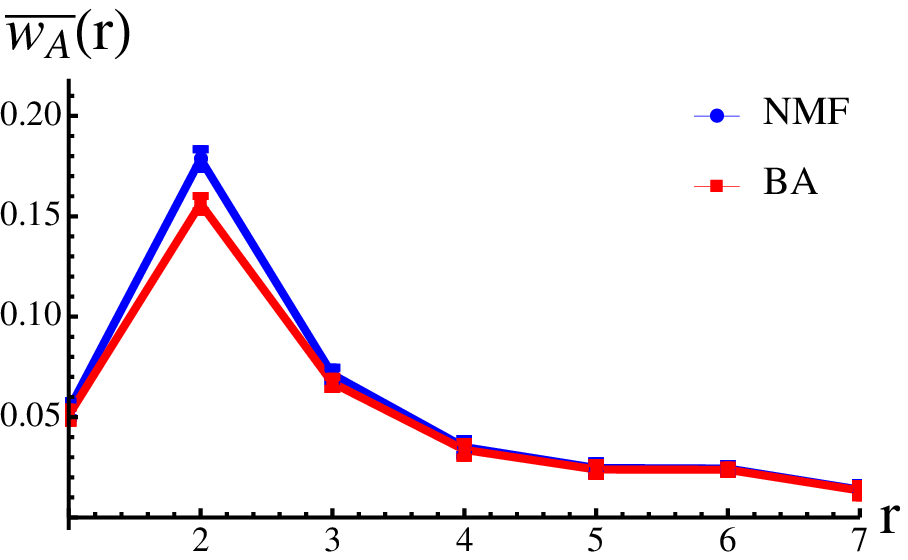}
  \includegraphics[width=0.40\columnwidth]{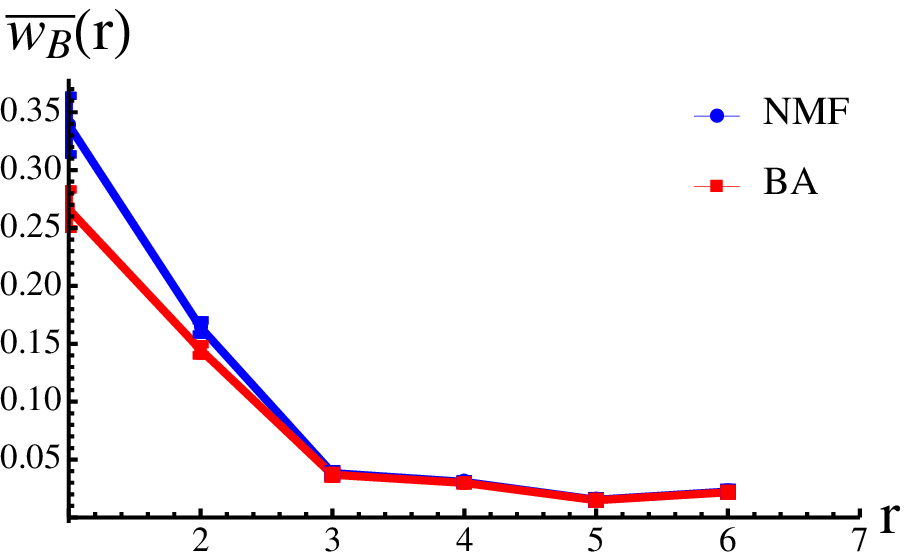}
 \caption{(Color online) Comparison in $\overline{w_{A}}(r)$ (left) and $\overline{w_{B}}(r)$ (right)
of the NMF and BA for patch size $L=16$. Relatively large
differences are present at $r=2$
 for $\overline{w_{A}}$ and at $r=1$ for $\overline{w_{B}}$.}
\Lfig{NMFBAcomp-forestneedles}
\end{center}
\end{figure}
We observe that the NMF gives larger values of $\overline{w}(r)$,
which in particular is clear in the NN interaction of
$\overline{w_{B}}$. This tendency was absent for the previous two
sets of pictures and may be related to the fractal nature of the
forest pictures, which can be connected to the strong criticality.

\subsection{Robustness of the Results}\Lsec{Stability}
For checking the robustness of the results thus far, in this section we present the result by the Monte Carlo (MC) simulation, as well as the result for other dithering methods. For simplicity, we only treat face pictures in this section, but we confirmed that the similar conclusion is obtained for other sets of images. 

\subsubsection{Comparison with Monte Carlo Simulation}\Lsec{Comparison}
The MC simulation is computationally demanding and only the size $L=8$ is treated here. The computation procedure is as follows. We search the maximum of log likelihood, \Req{max-loglikelihood}, by the Newton method, and the required average over the model in the moment matching condition is computed by the MC method. In the MC method, the well-known Metropolis algorithm is used, and the MC steps for sampling is fixed to be $N_{mc}=10000$. Once the absolute value of the gradient is smaller than a threshold value, here it is chosen to be $10^{-6}$, we regard the algorithm converges and the corresponding $\V{w}$ and $\V{h}$ are returned as the solution.

The \Rfig{NMFBAMCcomp-face} are the plots of $\overline{w_{A}}(r)$ and $\overline{w_{B}}(r)$ for the NMF, BA  and MC methods.  
\begin{figure}[htbp]
\begin{center}
  \includegraphics[width=0.40\columnwidth]{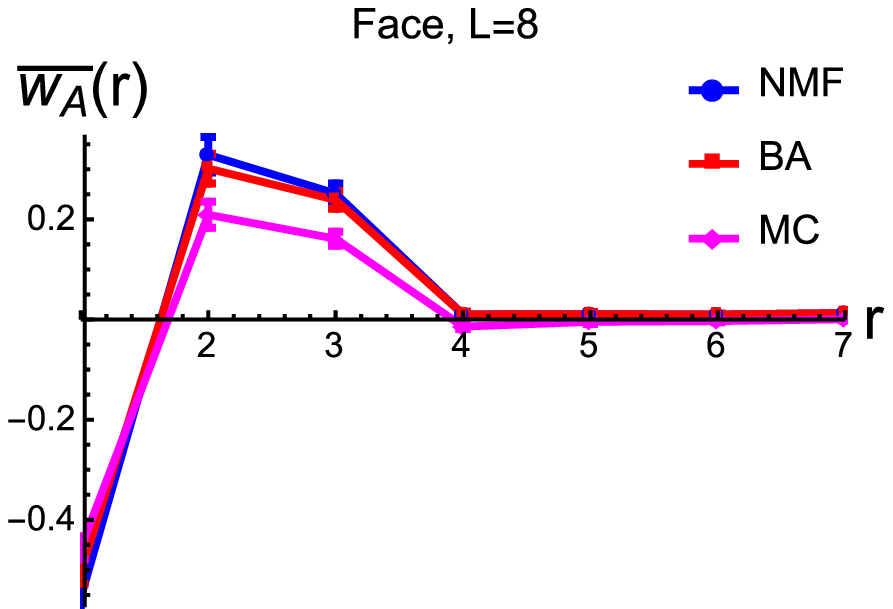}
  \includegraphics[width=0.40\columnwidth]{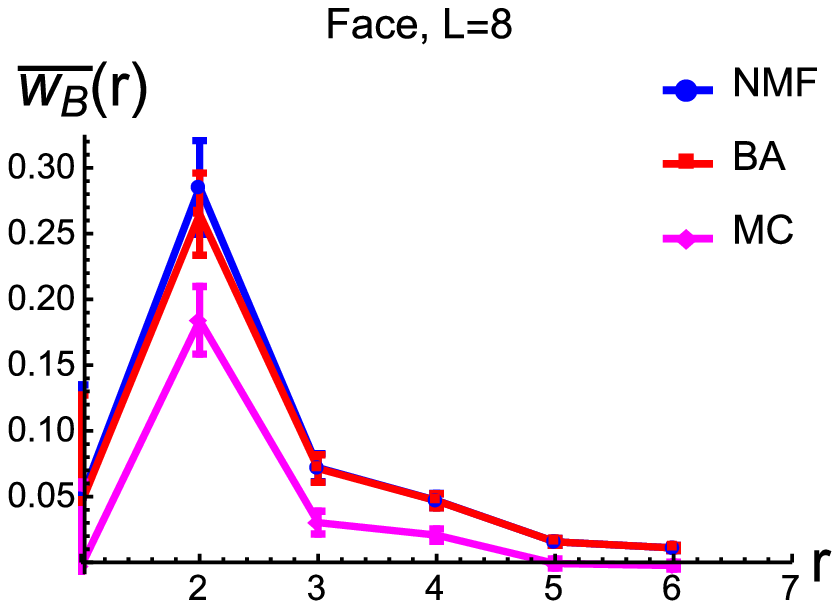}
 \caption{(Color online) Comparison in $\overline{w_{A}}(r)$ (left) and $\overline{w_{B}}(r)$ (right)
of the NMF, BA and MC methods, for patch size $L=8$ of face pictures. The functional shape of the MC result is similar to the other twos but the absolute value of $w$ tends to be smaller.}
\Lfig{NMFBAMCcomp-face}
\end{center}
\end{figure}
This figure clearly shows the functional form of $w$ of the MC result is similar to the ones of the other two methods, though its absolute value of $w$ tends to be smaller. Besides, \Rfig{L=8-faces-MC-Hist} displays the graph representation and the histogram of NN interactions. The result is very similar to the one by the mean-field methods given in \Rfigs{L=16-faces-NMF-NNandNNN}{L=16-faces-NMF-Hist}.
\begin{figure}[htbp]
\begin{center}
  \includegraphics[width=0.3\columnwidth]{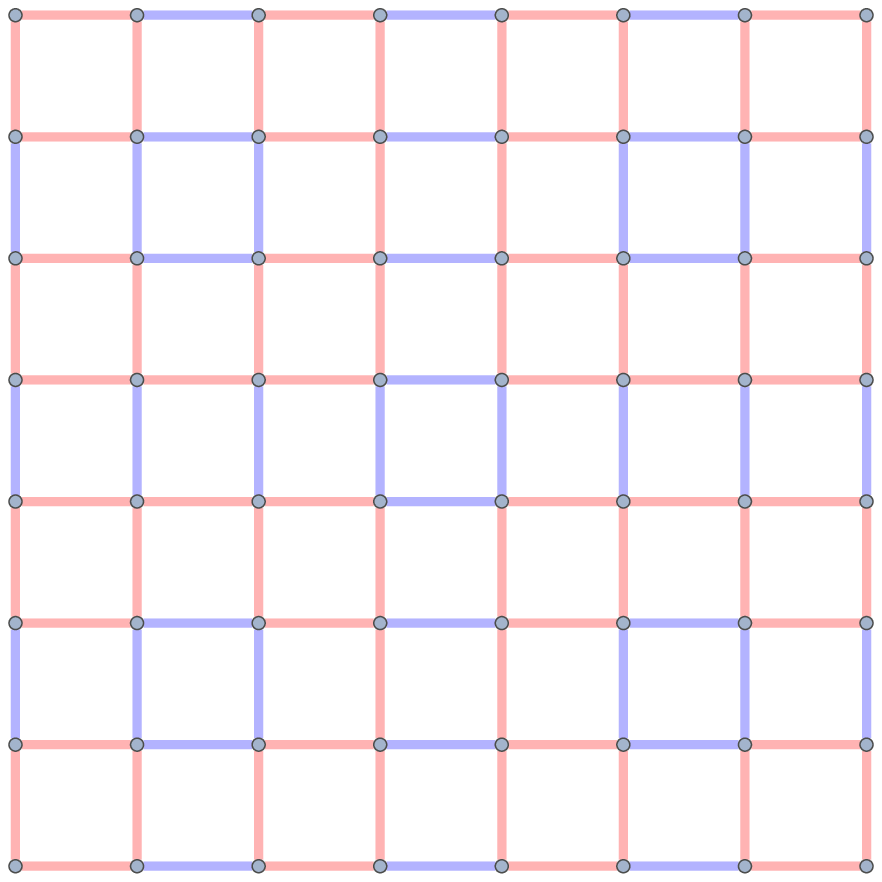}
  \hspace{2mm}
  \includegraphics[height=0.28\columnwidth]{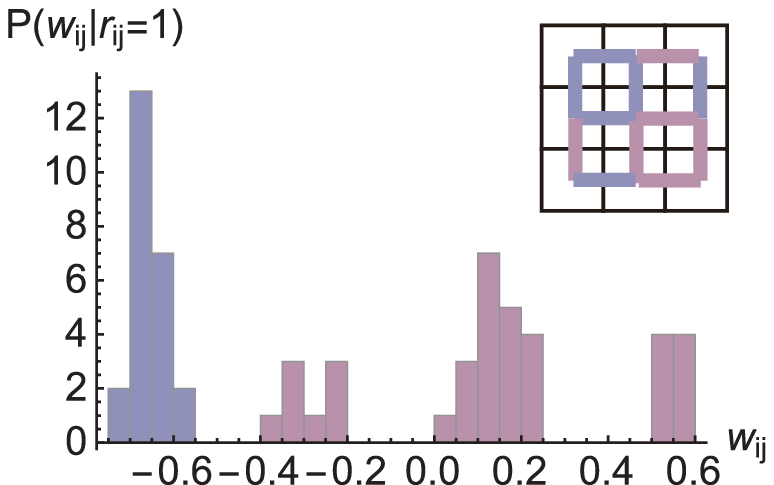}
 \caption{(Color online) Graph representation (left) and the histogram (right) of the NN  interactions derived by the MC simulation from face pictures of patch size $L=8$. The behavior is quite similar to the ones by the NMF in \Rfigs{L=16-faces-NMF-NNandNNN}{L=16-faces-NMF-Hist}. }
\Lfig{L=8-faces-MC-Hist}
\end{center}
\end{figure}
Hence, these finding well supports our main claim, the emergence of sublattice structure and the presence of characteristic length scale $\xi\approx 4$.

\subsubsection{Other Dithering Methods}\Lsec{Other}
The presented results thus far can be affected by the dithering method employed in preprocessing. Here we examine how the results change if other dithering methods are employed. The inferred method is fixed to be NMF.

There are so many dithering methods such as Floyd, Jarvis, Stucki, Burkes, and Sierra dithering. It is not easy to check all of them, and here we test only the cases with Floyd and no dither. 

\Rfig{NNandNNN-Floyd} shows graph representations of the NN and NNN interactions for the case of Floyd dither. The sublattice structure is clearly matching to the one in \Rfig{sublattice}, and hence our observation thus far can be applied to Floyd dither.
\begin{figure}[htbp]
\begin{center}
  \includegraphics[width=0.4\columnwidth]{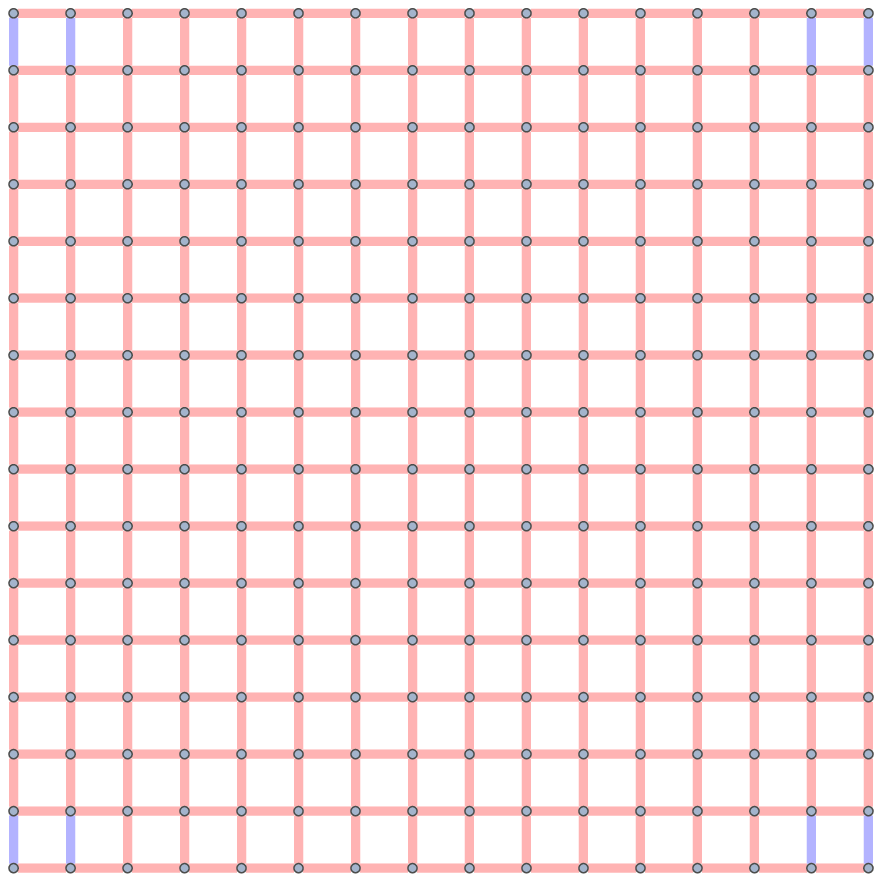}
  \includegraphics[width=0.4\columnwidth]{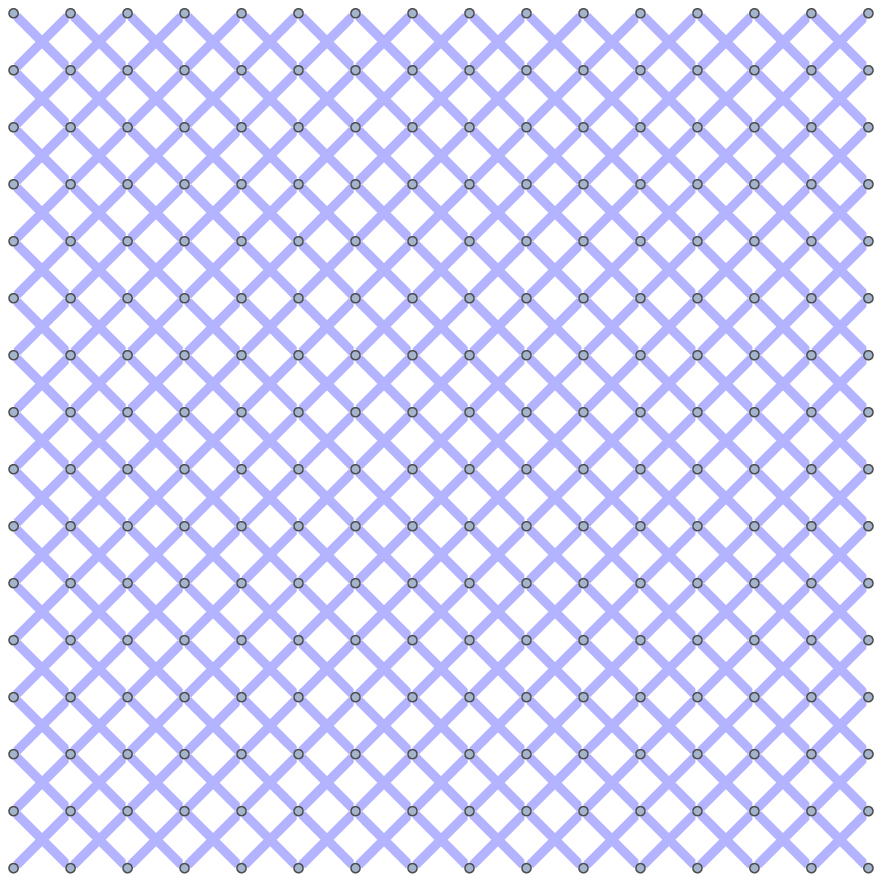}
 \caption{(Color online) Graph representations of the NN (left) and NNN (right) interactions in Floyd dithering for face pictures of patch size $L=16$. The sublattice structure is matching to the structure in \Rfig{sublattice}, but is simpler than the one by Riemersma dither shown in \Rfig{L=16-faces-NMF-NNandNNN}.  }
\Lfig{NNandNNN-Floyd}
\end{center}
\end{figure}
Meanwhile, the result with no dither is given in \Rfig{NNandNNN-None}. Unfortunately, the NNN interaction network does not match to the sublattice structure studied thus far.  
\begin{figure}[htbp]
\begin{center}
  \includegraphics[width=0.4\columnwidth]{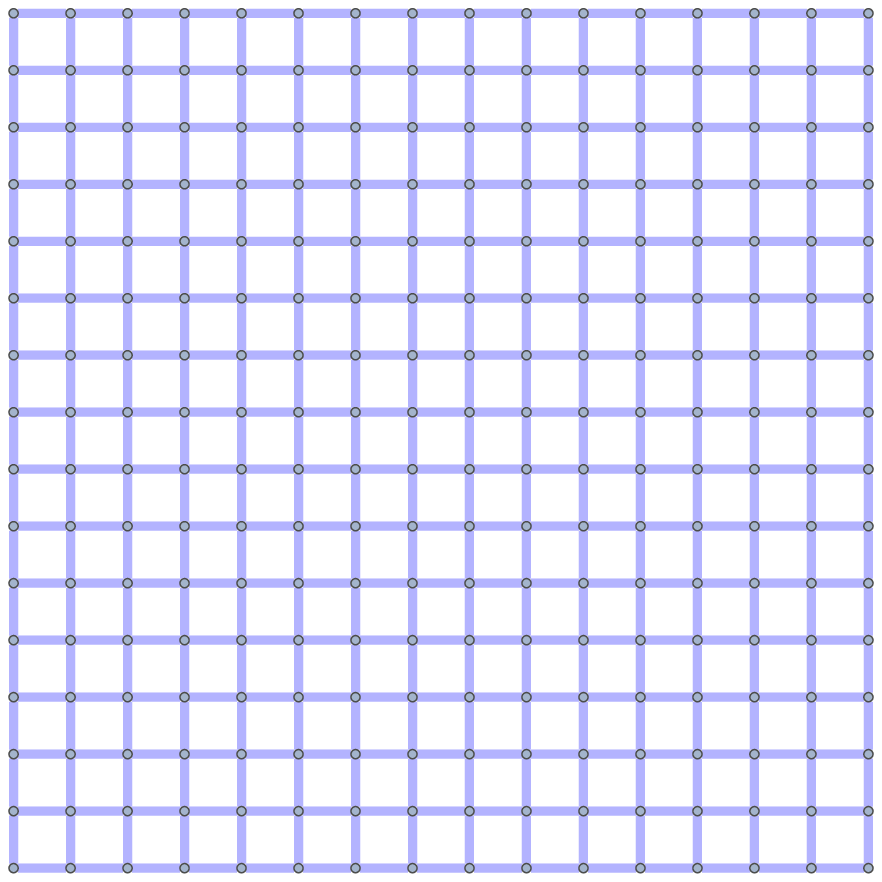}
  \includegraphics[width=0.4\columnwidth]{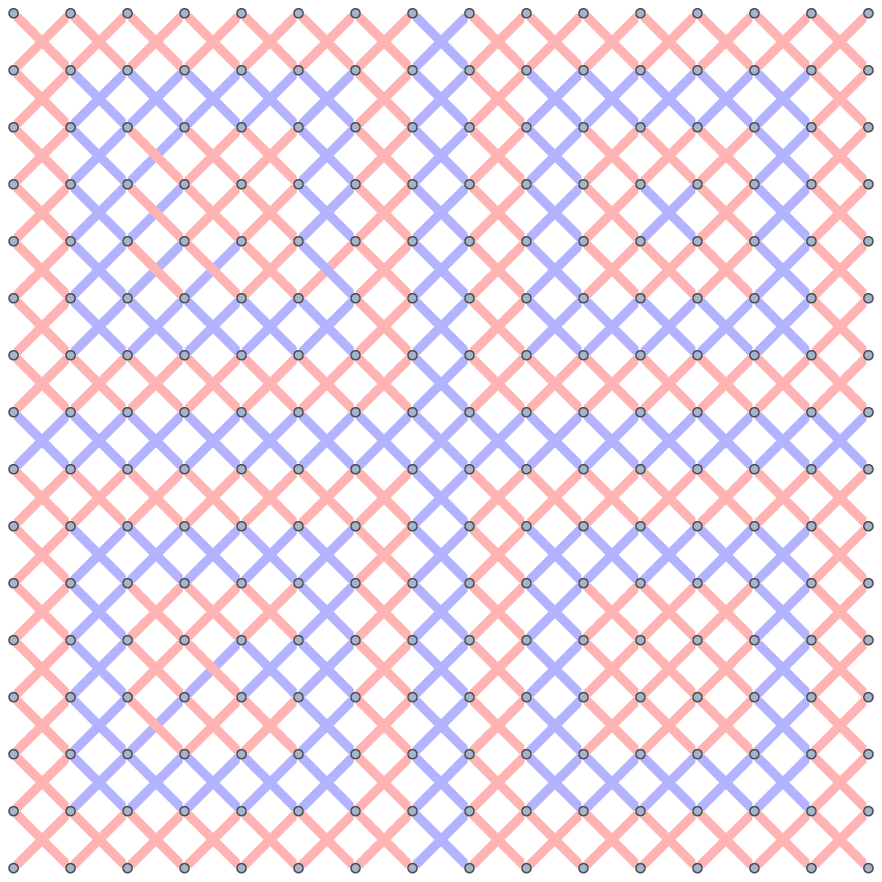}
 \caption{(Color online) Graph representations of the NN (left) and NNN (right) interactions without any dither for face pictures of patch size $L=16$. The sublattice structure for the NNN interactions is not matching to the structure in \Rfig{sublattice}.   }
\Lfig{NNandNNN-None}
\end{center}
\end{figure}
This difference in the no-dither case from other cases may be understood as follows. Any dither process introduces certain periodicity in the processed pictures to make contrasts among macroscopically discriminable regions. The induced periodicity is expected to have a relatively high frequency or short wavelength, to make the contrast vivid. Our proposed sublattice structure in \Rfig{sublattice} has the wavelength $2$ (in unit of the number of pixels), which is the nontrivial sublattice structure with the minimum wavelength. Hence based on this reasoning, it is natural that the dither images have this sublattice structure, while images without dither may not have such a contrast sublattice structure. 

Let us move to a quantitative comparison among different dithers. \Rfig{RiemFloydcomp-face} shows $\overline{w_{A}}(r)$ and $\overline{w_{B}}(r)$ for Riemersma and Floyd dithers.
\begin{figure}[htbp]
\begin{center}
  \includegraphics[width=0.40\columnwidth]{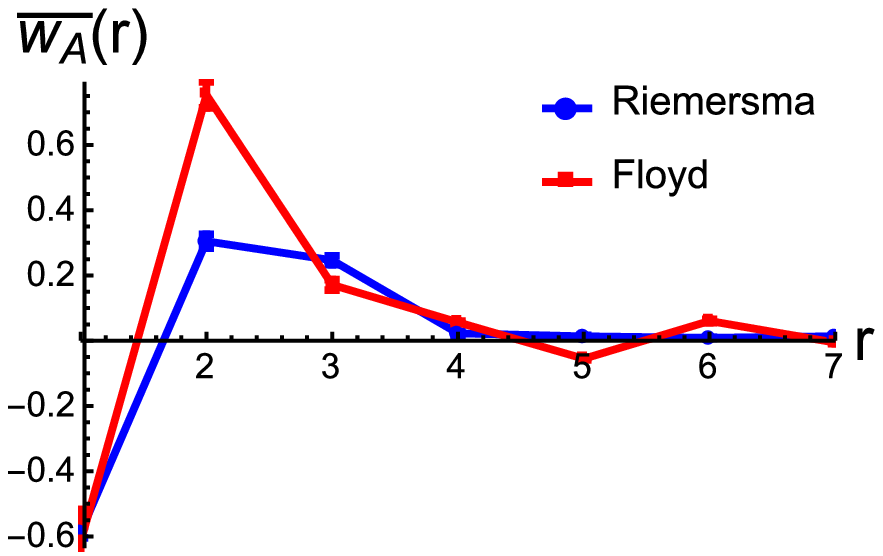}
  \includegraphics[width=0.40\columnwidth]{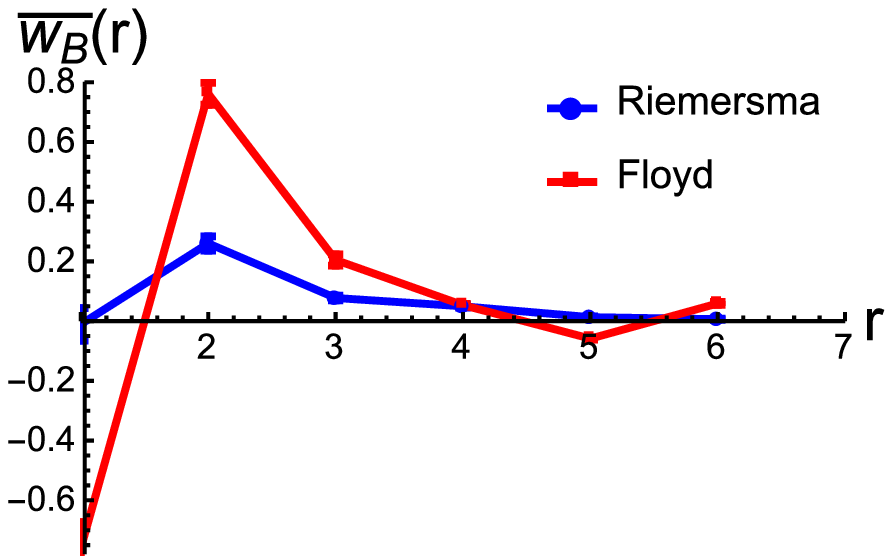}
 \caption{(Color online) Comparison in $\overline{w_{A}}(r)$ (left) and $\overline{w_{B}}(r)$ (right)
of the Riemersma and Floyd dithers, for patch size $L=16$ of face pictures. }
\Lfig{RiemFloydcomp-face}
\end{center}
\end{figure}
As understood from \Rfig{NNandNNN-Floyd}, the result of Floyd dither has no difference between $\overline{w_{A}}(r)$ and $\overline{w_{B}}(r)$, since it has a simpler structure. The absolute values of $w$ tend to be larger in Floyd dithering, although in both dithers the interactions rapidly decay as $r$ grows, and almost vanish around $r\approx 4$. The same plot for the no dither case is in \Rfig{Nonecomp-face}.  
\begin{figure}[htbp]
\begin{center}
  \includegraphics[width=0.40\columnwidth]{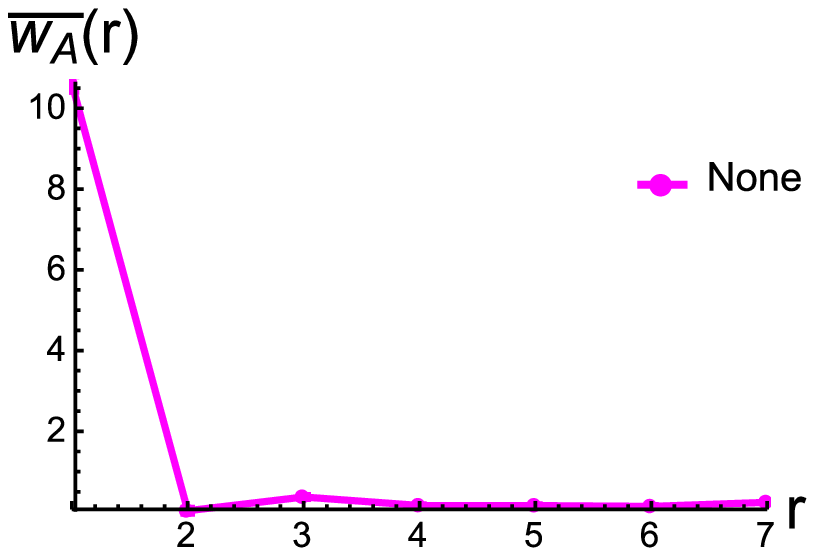}
  \includegraphics[width=0.40\columnwidth]{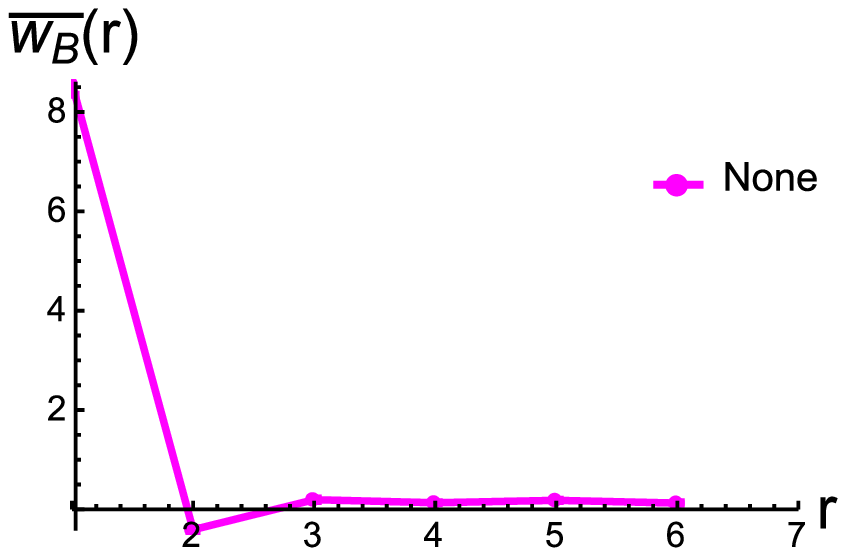}
 \caption{(Color online) Plots of $\overline{w_{A}}(r)$ (left) and $\overline{w_{B}}(r)$ (right) for patch size $L=16$ of face pictures in the no dither case. }
\Lfig{Nonecomp-face}
\end{center}
\end{figure}
In this case, the value of $w(r)$ at $r=1$ is significantly larger than the dither cases, while it more rapidly decays as $r$ grows (it almost vanishes at $r=2$). This might be because in the no dither case any meaningful structure common in any patch of images is absent and only the NN interaction $w(r=1)$ holds the information of the images. 

\subsection{Inferred Fields}\Lsec{Fields}
In this section, we present the histograms of local magnetizations, $P(m_i)$, and of local fields, $P(h_i)$, inferred by the NMF and the MC simulation, in each set of pictures. We show the results only for $L=16$, since we did not find any meaningful size effect.

\Rfig{L=8-magnetizations} shows the histograms of local magnetizations of aerial (left), face  (center), and forest pictures (right).
\begin{figure}[htbp]
\begin{center}
  \includegraphics[width=0.32\columnwidth]{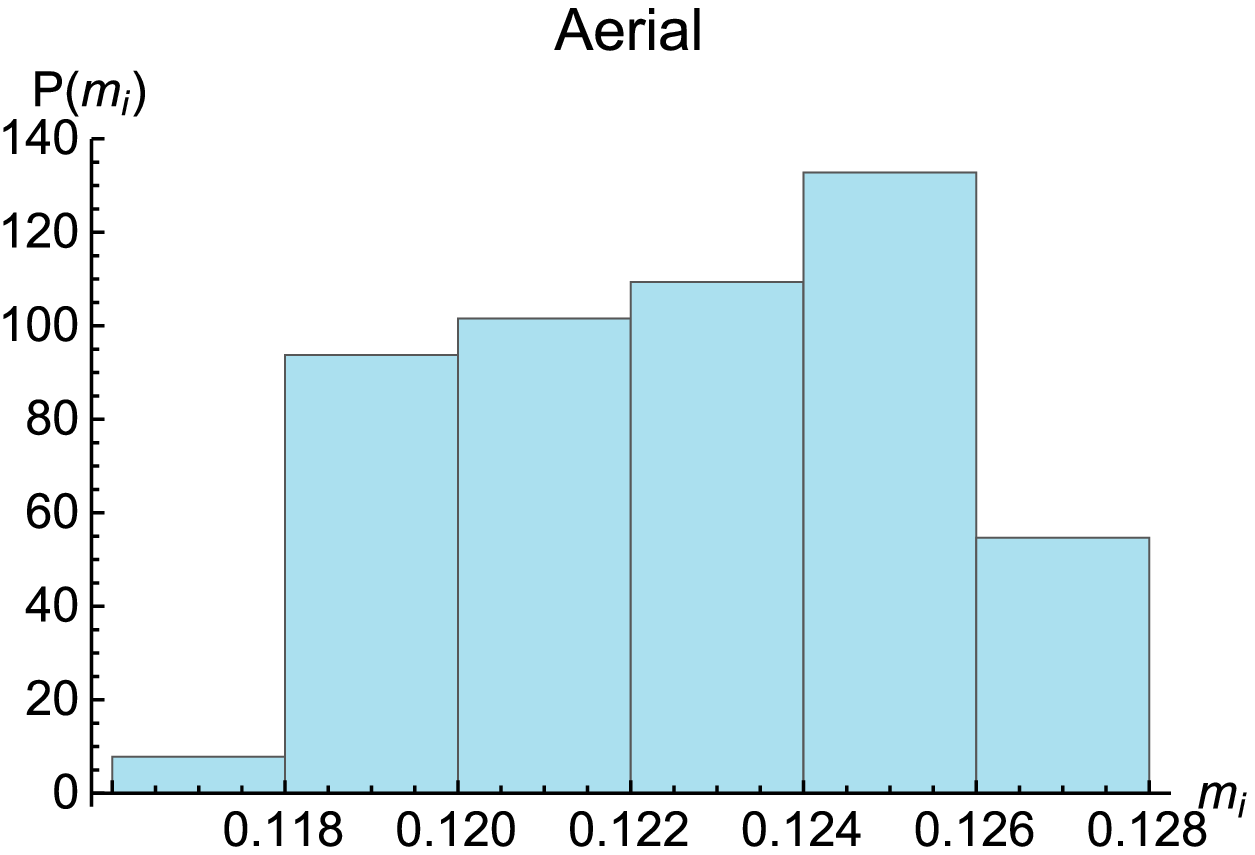}
  \includegraphics[width=0.32\columnwidth]{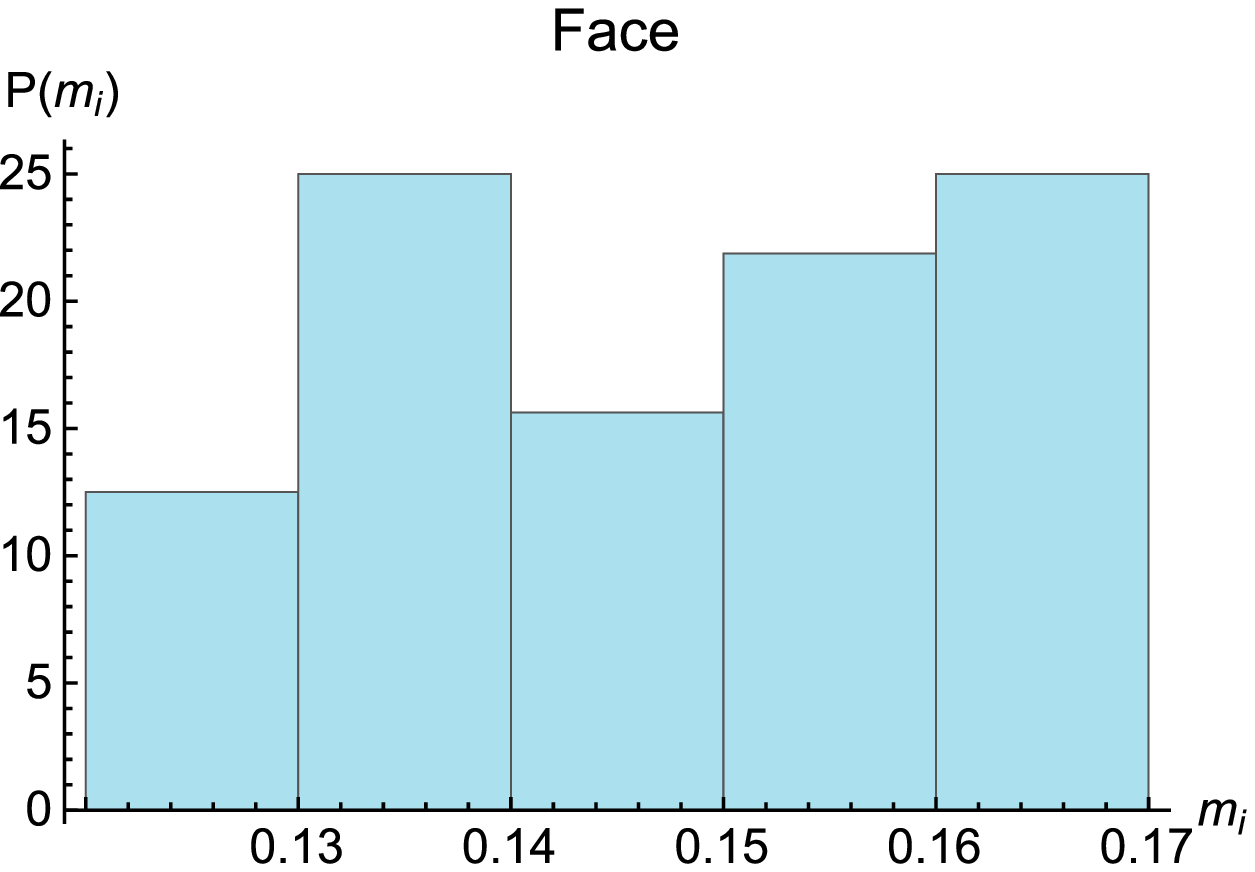}
  \includegraphics[width=0.32\columnwidth]{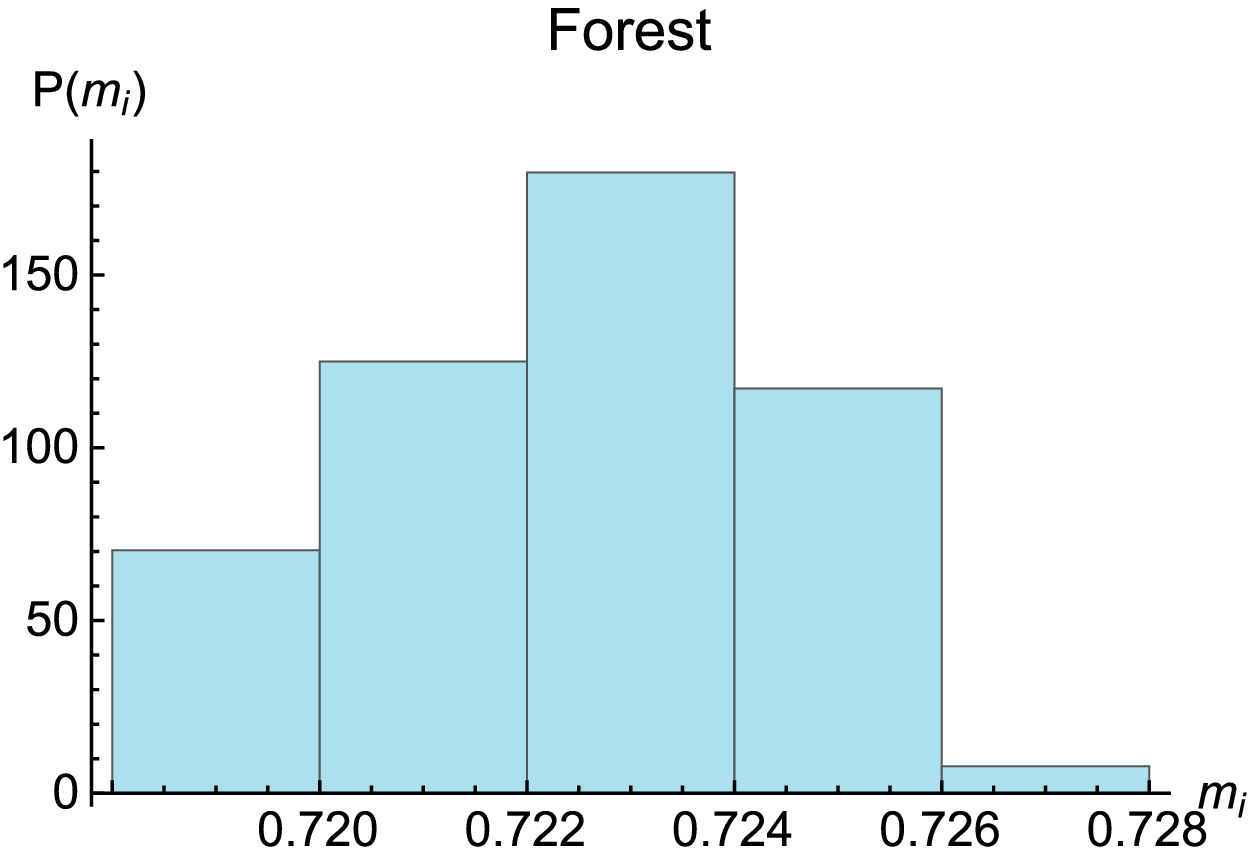}
 \caption{(Color online) Normalized histograms of the local magnetizations of aerial (left),
 face (center), and forest pictures (right) for patch size $L=8$.
 They are normalized as probability distribution, namely, the volume of the histogram is unity.
 In all cases, the magnetizations are positive, meaning that black pixels appear more frequently.}
\Lfig{L=8-magnetizations}
\end{center}
\end{figure}
All the histograms have a simple peak structure in the positive $m_i>0$ region.
This implies black pixels appear more frequently in all the sets of pictures.

In \Rfig{L=8-fields-NMF}, we show the histograms of the local fields inferred by the NMF for aerial (left), face (center), and forest pictures (right). The same histograms for the MC simulation are also given in \Rfig{L=8-fields-MC}.
\begin{figure}[htbp]
\begin{center}
  \includegraphics[width=0.32\columnwidth]{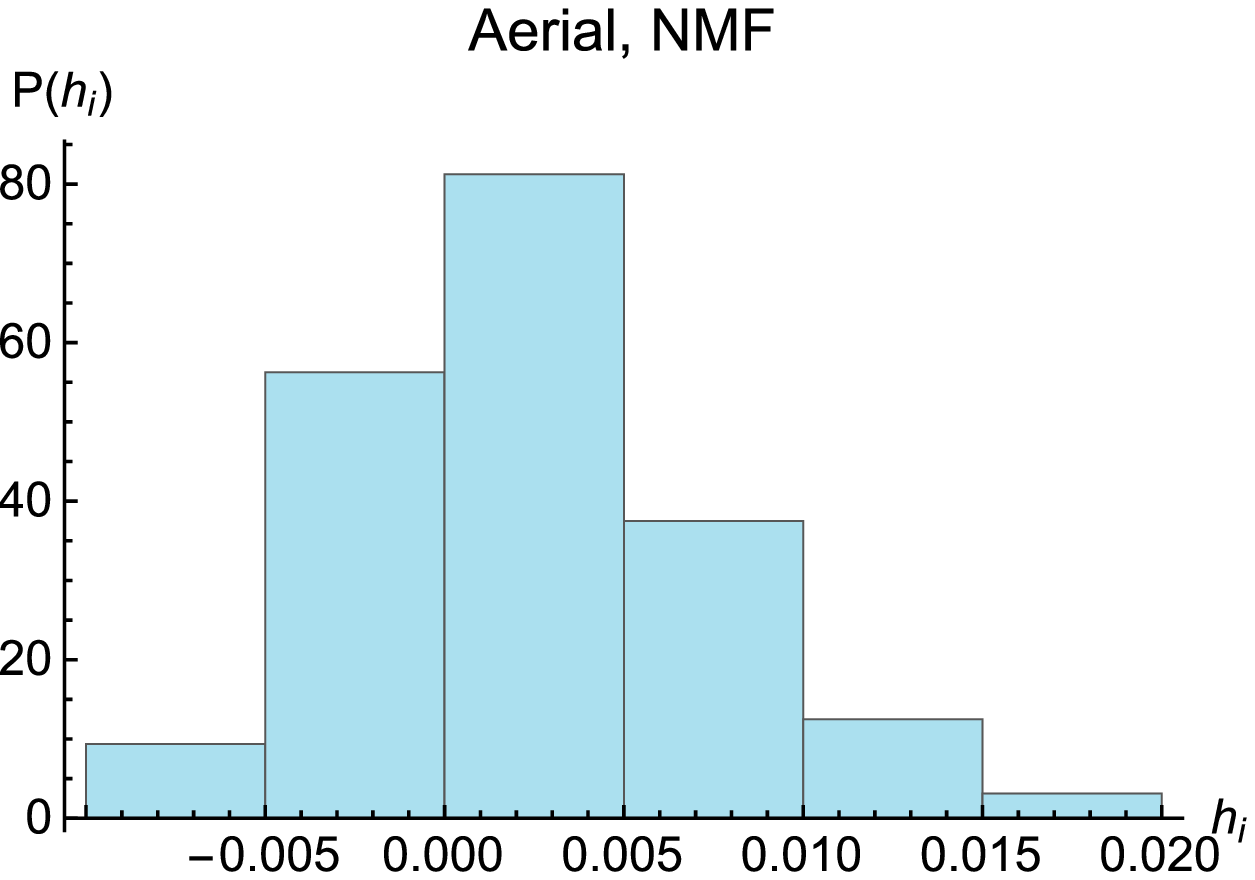}
  \includegraphics[width=0.32\columnwidth]{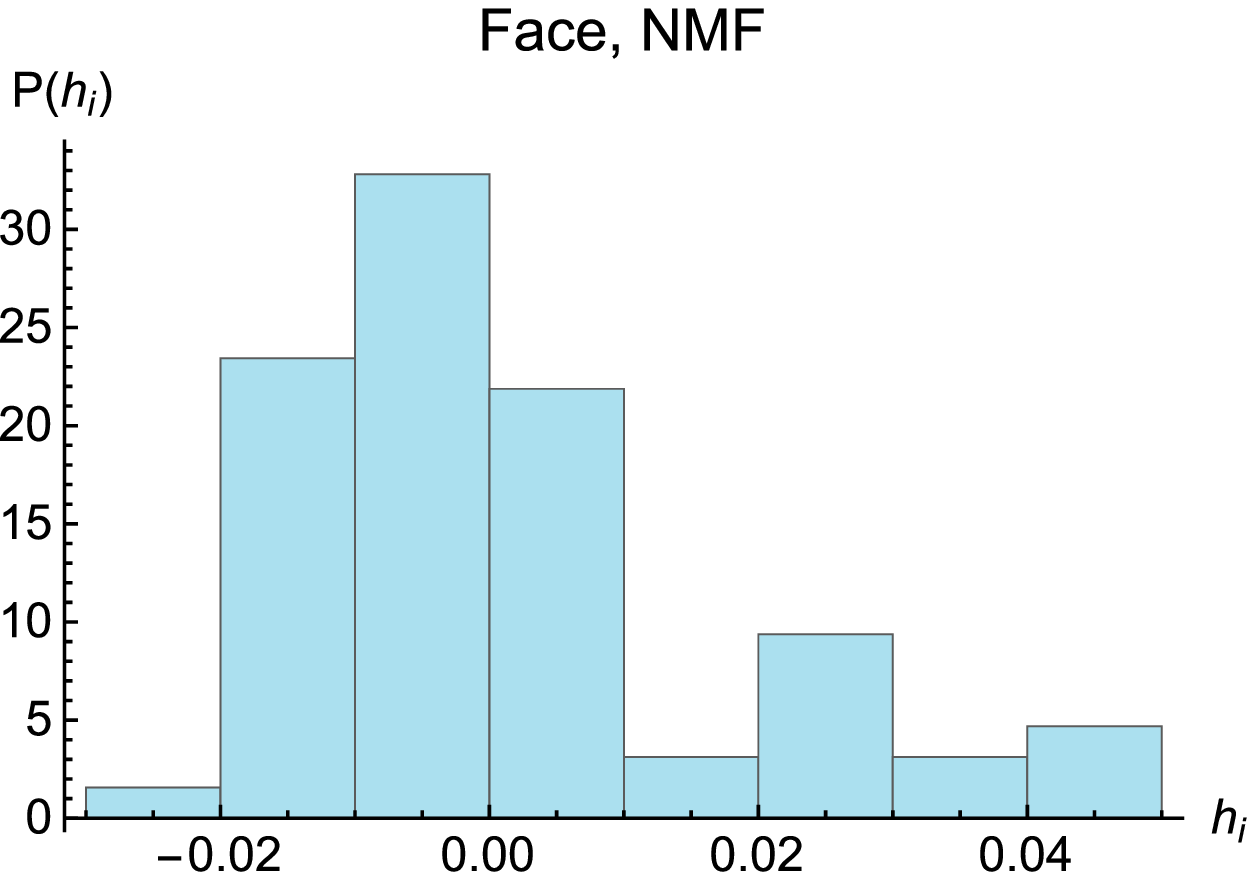}
  \includegraphics[width=0.32\columnwidth]{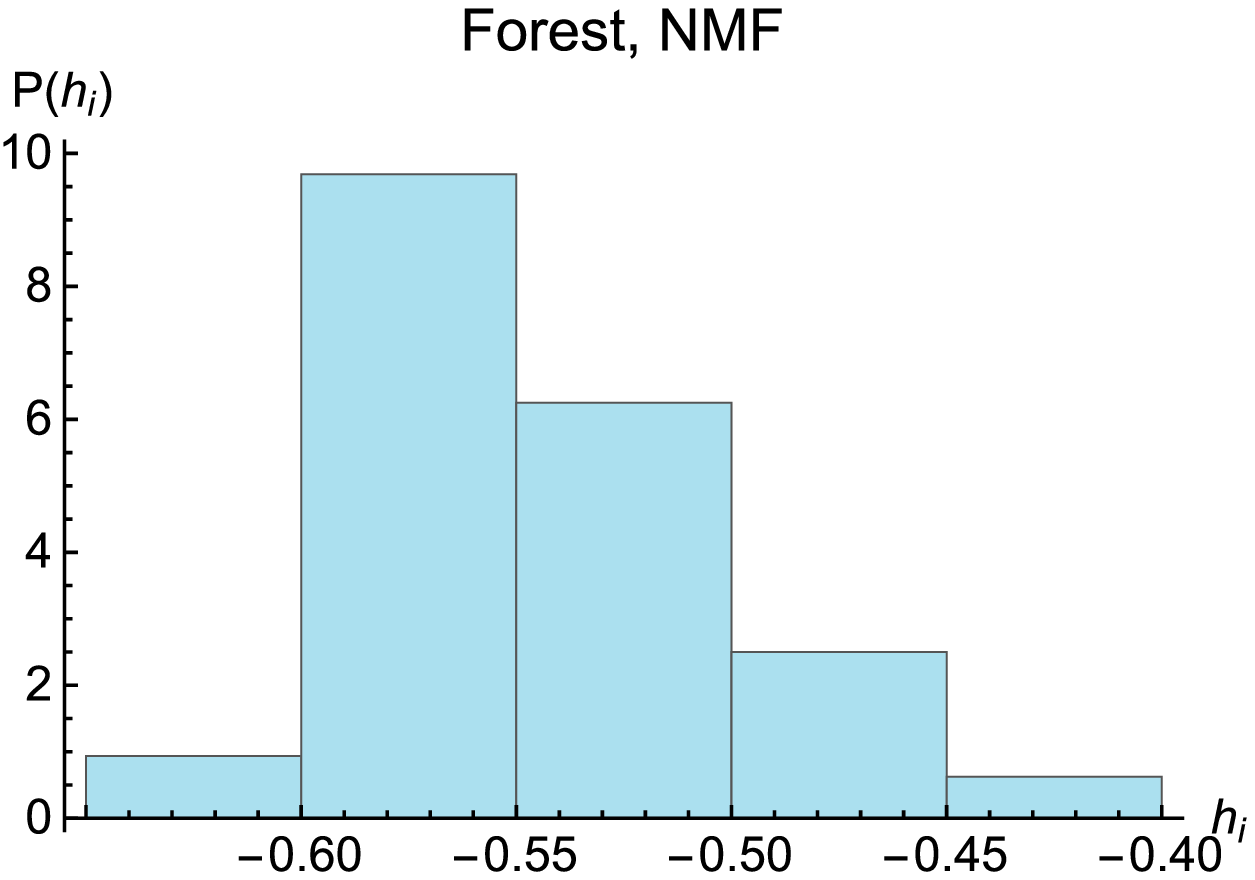}
 \caption{(Color online) Normalized histograms of local fields of aerial (left), face (center), and forest pictures (right), inferred by the NMF.  Although the magnetizations are positive, the local fields inferred by the NMF tend to be negative. }
\Lfig{L=8-fields-NMF}
\end{center}
\end{figure}
\begin{figure}[htbp]
\begin{center}
  \includegraphics[width=0.32\columnwidth]{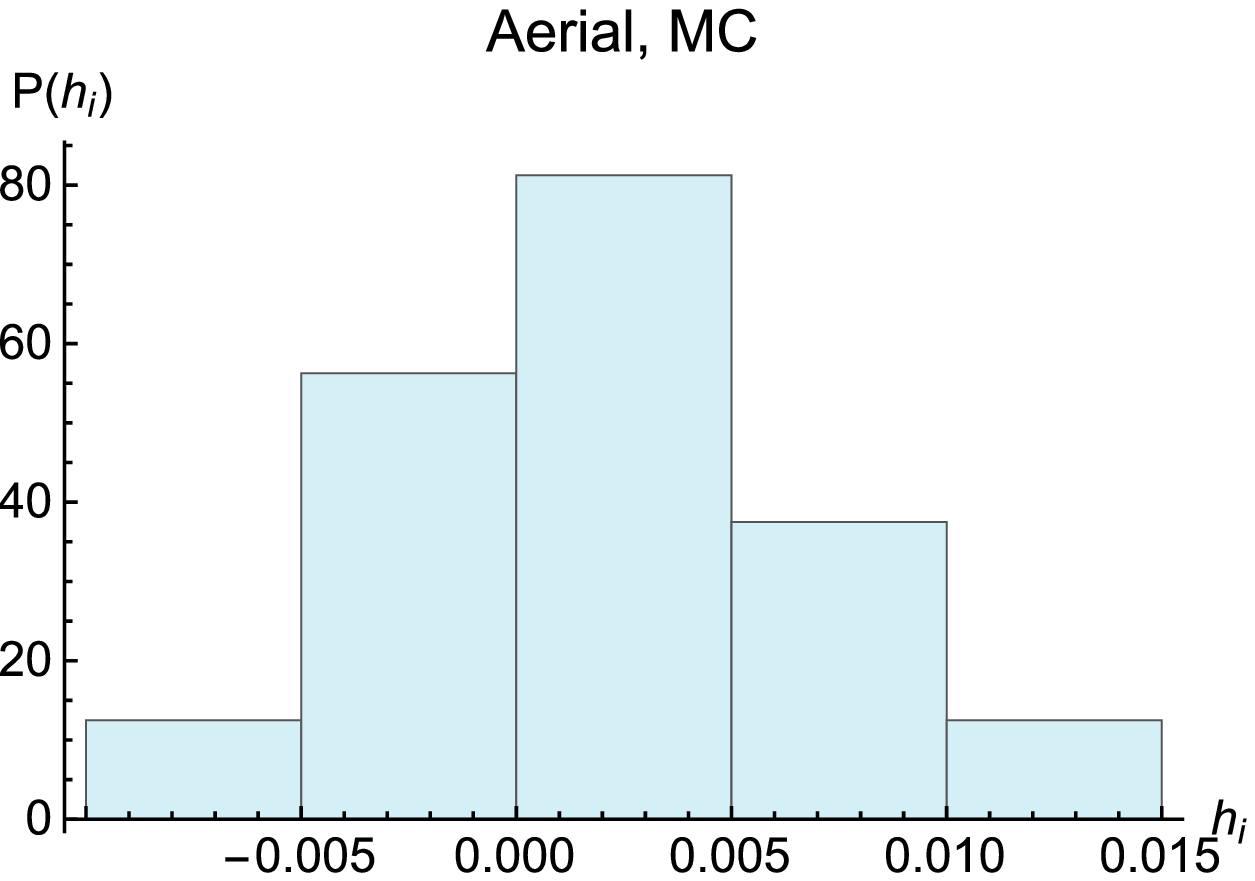}
  \includegraphics[width=0.32\columnwidth]{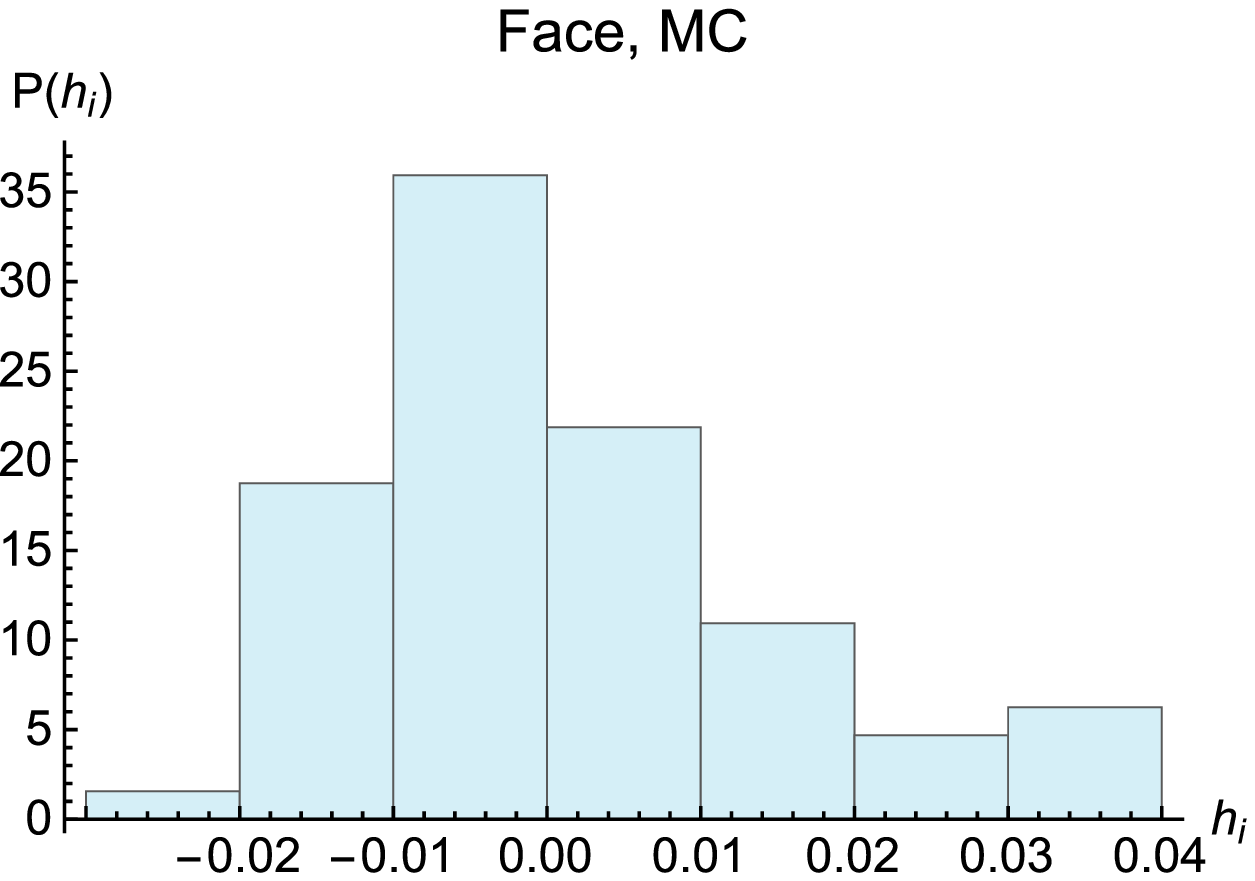}
  \includegraphics[width=0.32\columnwidth]{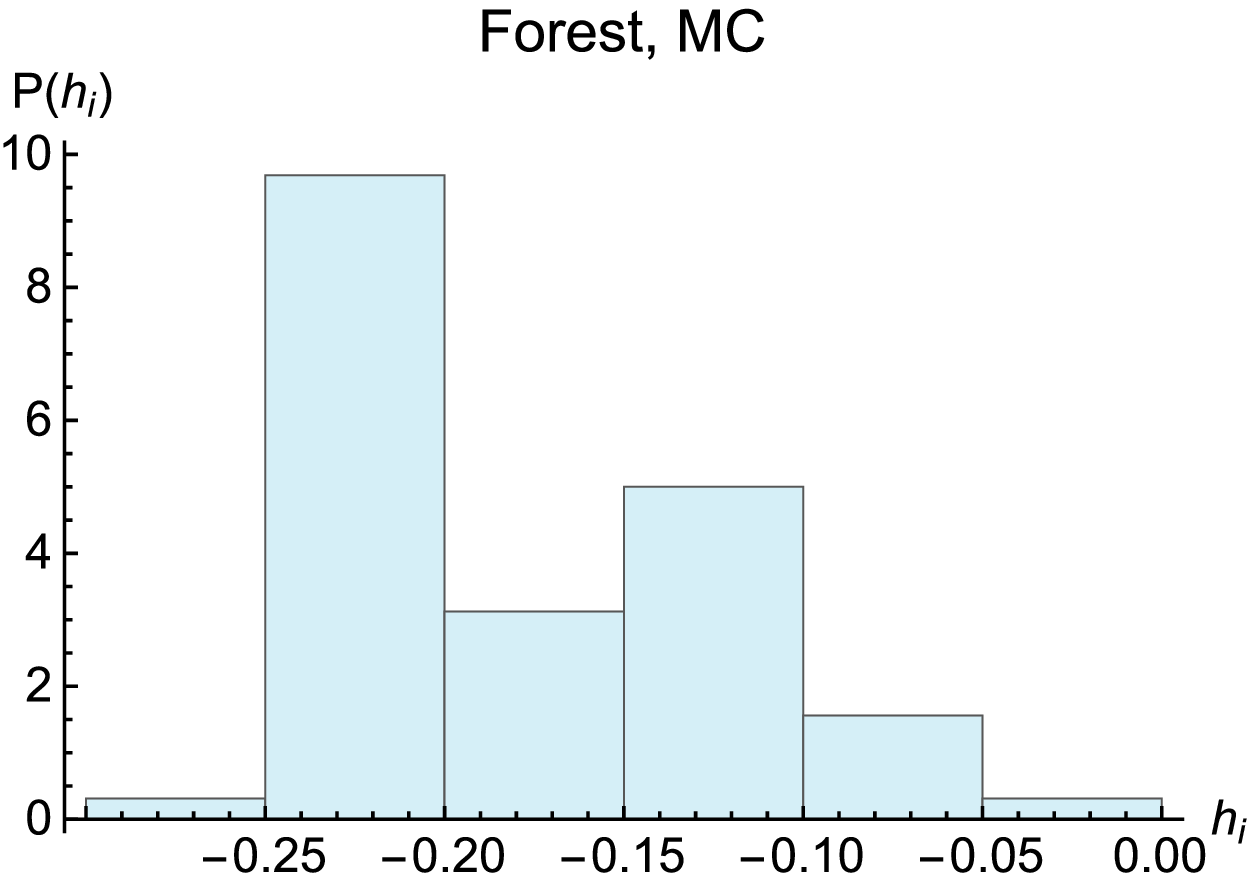}
 \caption{(Color online) Normalized histograms of local fields of aerial(left), face (center), and forest pictures (right), inferred by the MC simulation. The behavior is similar to the NMF one.}
\Lfig{L=8-fields-MC}
\end{center}
\end{figure}
These figures exhibit the inferred fields tend to be negative for some cases, while the local magnetizations are positive. We expect that this seemingly unnatural behavior is a consequence of the higher order statistics of natural images than the second. This can be understood by seeing an example. Suppose pictures are generated from the following simple Boltzmann distribution with $r$-body interactions:
\be
p_{r}=\frac{1}{Z_r}e^{-H_{r}(\V{S})},~\mathcal{H}_r(\V{S})=-NK\lb \frac{1}{N}\sum_{i}S_{i}\rb^r.
\Leq{p_r}
\ee 
In this case, it is possible to show the mean-field result is exact in the limit $N\to \infty$, and also possible to derive an analytical formula of the effective pairwise interaction and the local field of \Reqs{NMF1}{NMF2} in the limit $B\to \infty$. The computation is straightforward and we only refer to the result:
\be
w_{ij}=\frac{K}{N}r(r-1)m^{r-2},~h_i=-Kr(r-2)m^{r-1},
\Leq{solution-p_r}
\ee
where $m$ is the spontaneous magnetization of \Req{p_r} and is the solution of the following self-consistent equation 
\be
m=\tanh \lb  Krm^{r-1} \rb.
\ee
\BReq{solution-p_r} tells us that the sign of the local field changes at $r=2$ and negative for $r>2$ with positive $m$. Hence, it is possible to regard the negativity of inferred fields in \Rfigs{L=8-fields-NMF}{L=8-fields-MC} as the indirect evidence of the importance of high order statistics in natural images.  

\section{Discussion}\Lsec{Discussion}

\subsection{Criticality}\Lsec{Criticality}
In this section, we address the possible criticality of the
Boltzmann machine after learning. The set up of the investigation
is as follows. We use the interactions and fields derived by the
BA, $\V{w}=\V{w}_{\rm BA}$ and $\V{h}=\V{h}_{\rm BA}$, for each
set of pictures.  We define a new Ising model with temperature,
the probability distribution of which is \be p(\V{S}|\V{w}_{\rm
BA},\V{h}_{\rm BA},T)=\frac{1}{Z(\V{w}_{\rm BA},\V{h}_{\rm
BA},T)}e^{-\frac{\mathcal{H}(\V{S}|\V{w}_{\rm BA},\V{h}_{\rm
BA})}{T}}, \ee and hence, the original Boltzmann machine
corresponds to $T=1$. We employ a standard Monte-Carlo
technique to simulate this Ising model, which enables us to
calculate physical quantities while changing the temperature. If
the original Boltzmann machine is critical, characteristic
features in certain physical quantities appear around $T=1$.

To make the point clearer, we calculate the specific heat
($N=L^2$ is the total number of spins) \be C=\frac{1}{NT^2}\lb
\Ave{ \mathcal{H}^2(\V{S}|\V{w}_{\rm BA},\V{h}_{\rm BA})
}_{\V{w}_{\rm BA},\V{h}_{\rm BA},T}-\Ave{
\mathcal{H}(\V{S}|\V{w}_{\rm BA},\V{h}_{\rm BA}) }^2_{\V{w}_{\rm
BA},\V{h}_{\rm BA},T}\rb \ee and identify the peak location of
the specific heat as the ``critical'' point. Here, the brackets
$\Ave{\cdots}_{\V{w}_{\rm BA},\V{h}_{\rm BA},T}$ denote the
average over the Ising model. This is a natural choice, since the
specific heat is connected to the variation ratio of the entropy
as the temperature changes and is known in fact to show a
characteristic divergence at the critical point in many systems.
The data of the specific heat of the Ising models are shown in
\Rfig{C-BA}.
\begin{figure}[htbp]
\begin{center}
  \includegraphics[width=0.32\columnwidth]{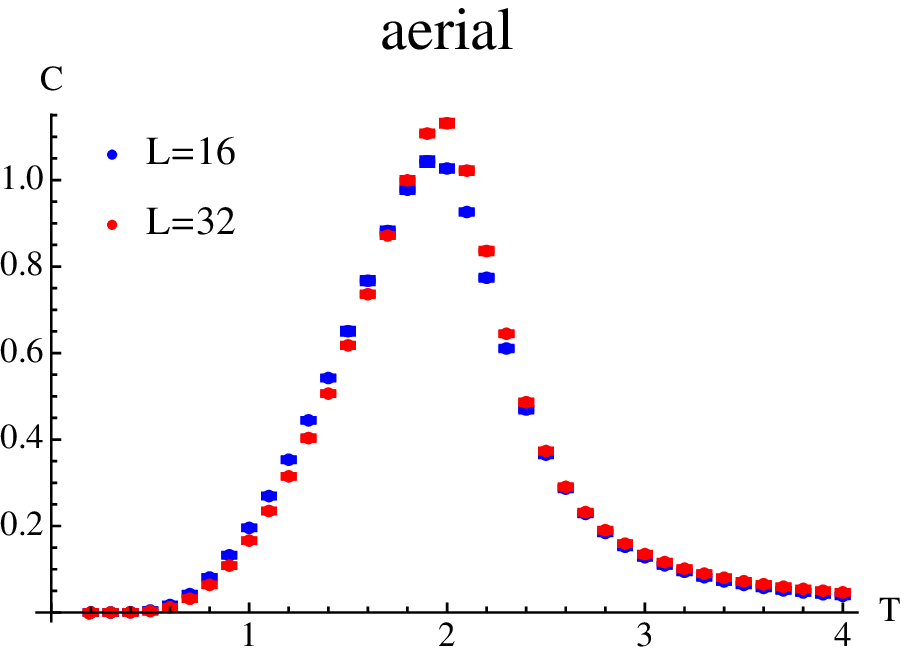}
  \includegraphics[width=0.32\columnwidth]{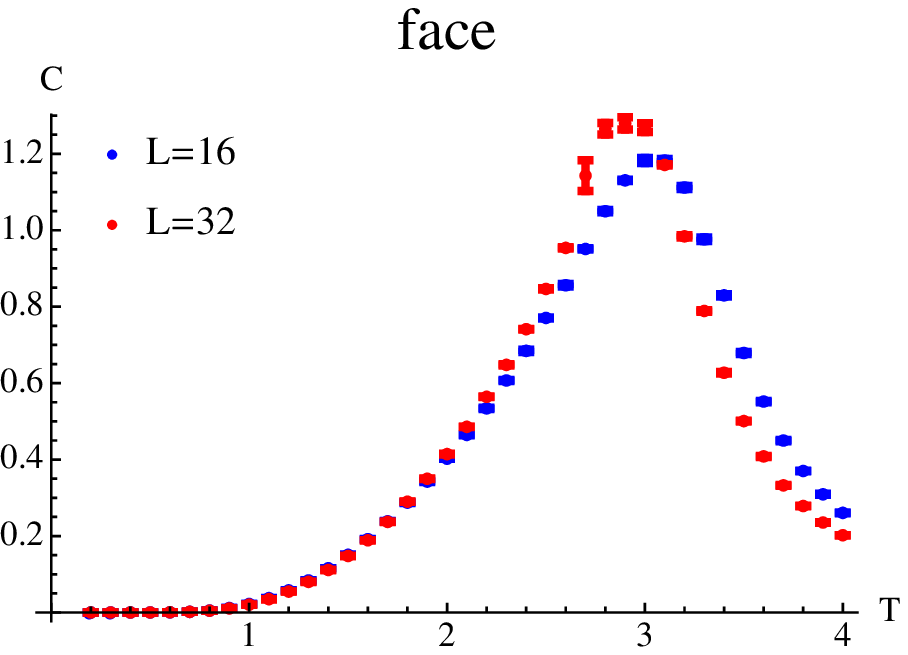}
  \includegraphics[width=0.32\columnwidth]{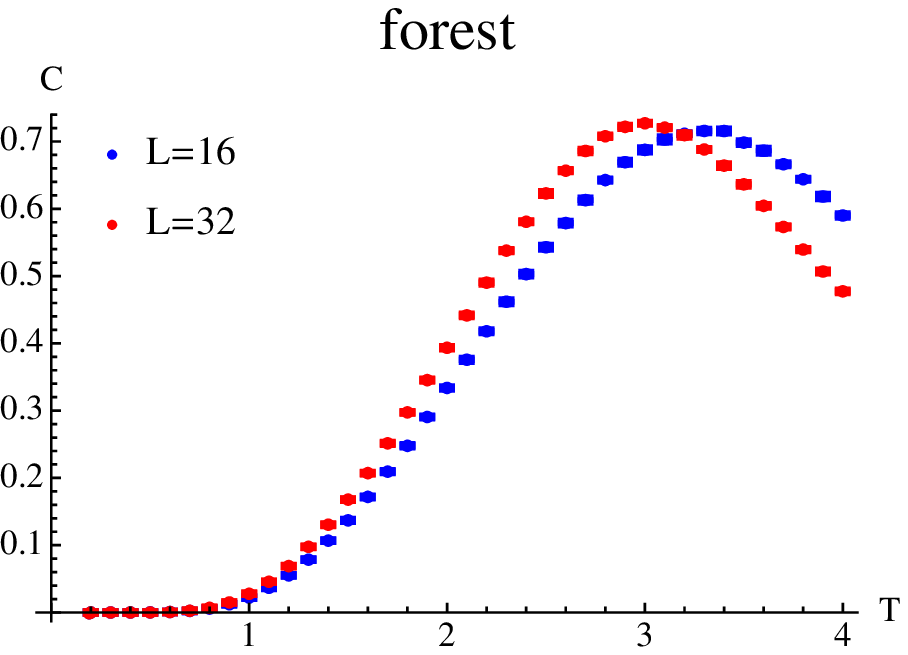}
  \includegraphics[width=0.32\columnwidth]{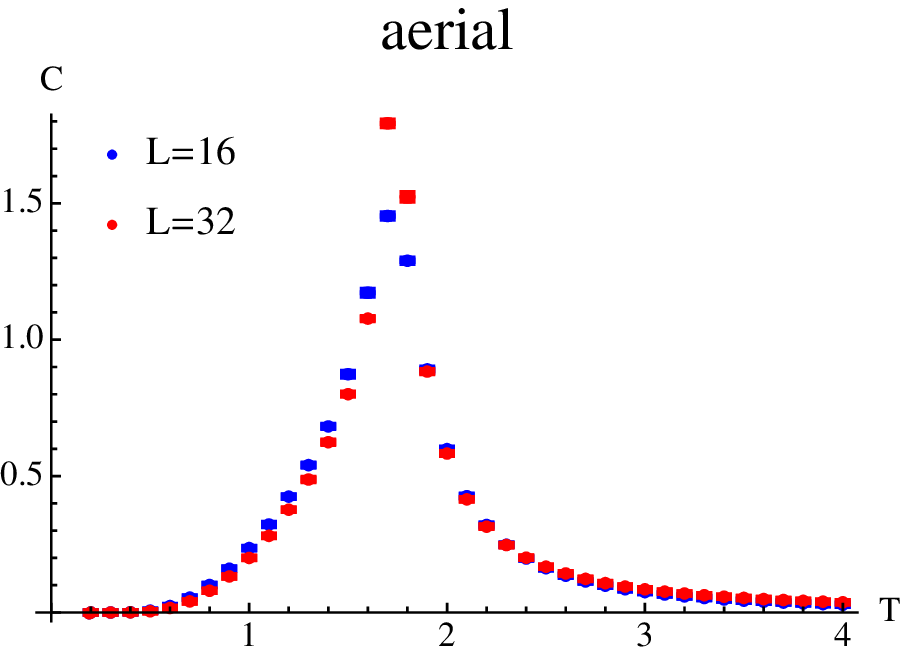}
  \includegraphics[width=0.32\columnwidth]{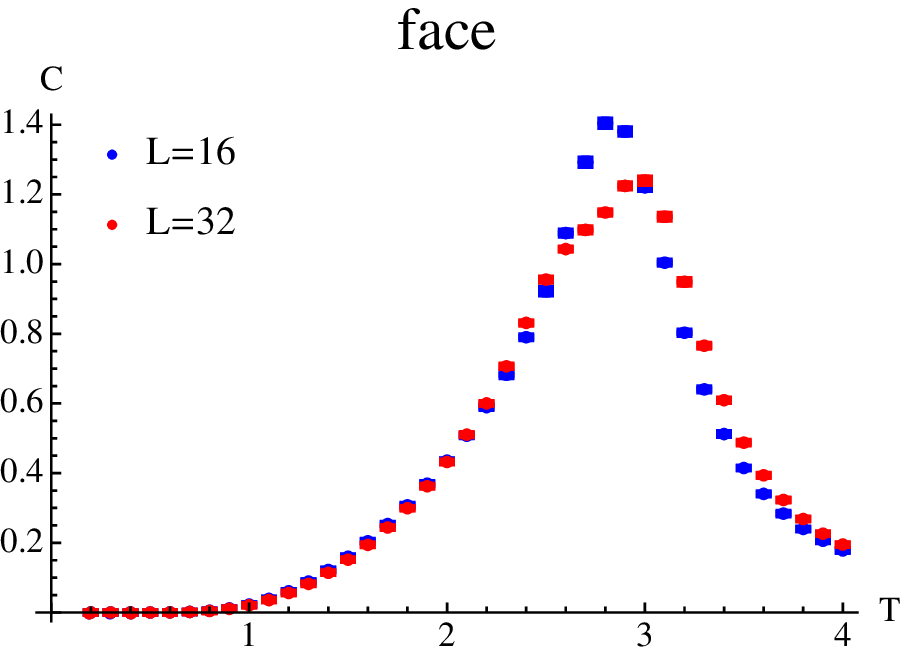}
  \includegraphics[width=0.32\columnwidth]{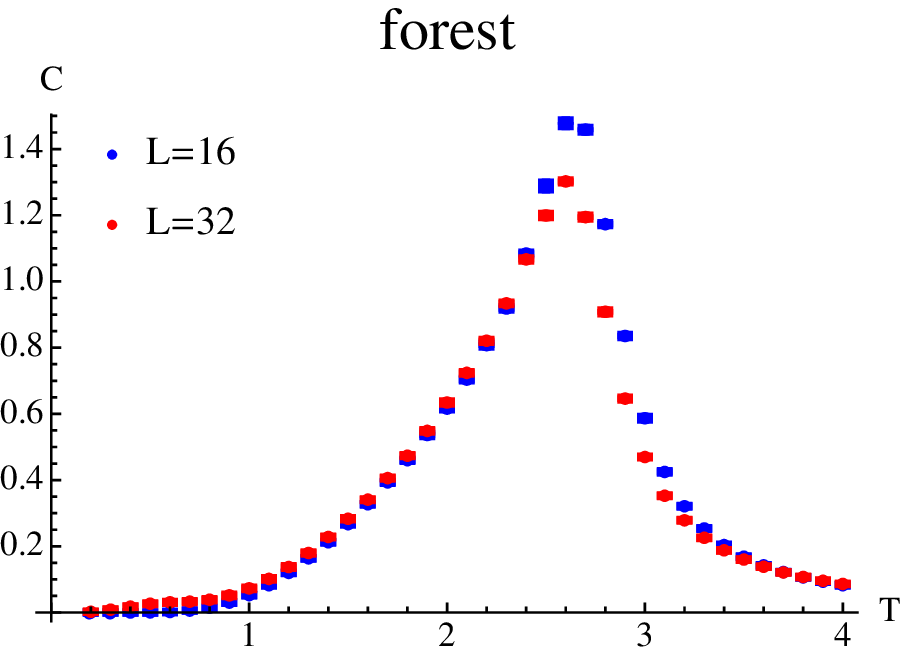}
  \caption{(Color online) Specific heat of the Ising models defined from the Boltzmann machines
  derived by the BA analysis of natural images, plotted against the temperature $T$.
  The data on the upper row are for $\V{h}=\V{h}_{\rm BA}$, but on the lower row correspond to $\V{h}=\V{0}$.
  The original Boltzmann machines correspond to $T=1$. Peak locations of the specific heat are far above $T=1$,
   irrespective of the presence of the on-site fields.  }
\Lfig{C-BA}
\end{center}
\end{figure}
For comparison, on the upper row we display the data with the
fields set to be the values inferred by the BA, $\V{h}=\V{h}_{\rm
BA}$, while on the lower row the data without the fields,
$\V{h}=\V{0}$, are shown. As seen clearly, the peaks of specific
heat locate far above $T=1$ in all the figures, implying that the
Boltzmann machines after learning are rather in the lower
temperature regions than at the critical point. The data without
the fields show sharp peaks and thus these Ising models do in fact
enjoy phase transitions. In addition, the Ising model for face
pictures shows another complicated structure: we see not only a
peak at $T\approx 3.0$ but also a shoulder-like structure at
$T\approx 2.6$. The shoulder may be an indication of another
transition, which may be similar to the multiple transitions in
the Ising model on the Union Jack lattice~\cite{EXAC}. This
seems to be reasonable, since the inferred interactions of the
face pictures are similar to those of the Union Jack lattice (see
\Rfig{L=16-faces-NMF-NNandNNN}).

Why does  the learned Boltzmann machine not exhibit criticality?
There are two possibilities: one is that our binarized images are
not at criticality; the other is that the Boltzmann machine cannot
extract the criticality of natural images. To examine the first
possibility, we plot the Fourier amplitude against the Fourier
spatial frequency in \Rfig{fourieramplitude-aerial} of the aerial
pictures as a representative.
\begin{figure}[htbp]
\begin{center}
  \includegraphics[width=0.5\columnwidth]{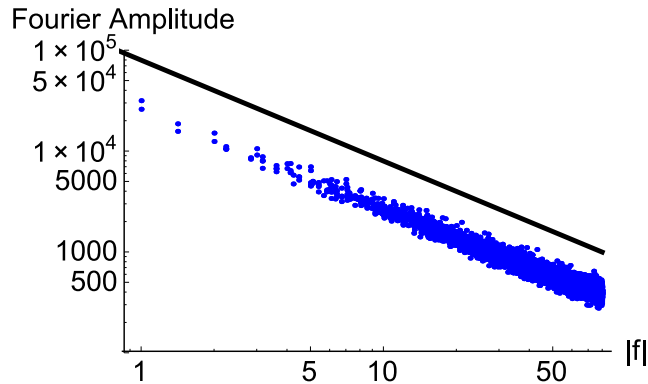}
  \caption{(Color online) Fourier amplitude of aerial pictures plotted against the absolute value
  of Fourier spatial frequency. The data show a clear power law comparable with $c/|f|$ (black straight line).}
\Lfig{fourieramplitude-aerial}
\end{center}
\end{figure}
In the figure, we see a clear power law, even after the pictures
are preprocessed by quantization and dithering, indicating that
the criticality still holds in our binarized pictures. Similar
behavior is seen in the face and forest pictures. Thus, the first
possibility, that our binarized pictures are not at criticality,
is not the case, and the second, that the Boltzmann machine cannot
extract the criticality of natural images, should occur. One
possible origin of this defect of the Boltzmann machine is that it
does not take into account higher order statistics than the
second. It is known that high order statistics exist in natural
images and significantly influence the properties of
images~\cite{Simoncelli:01,Hyvrinen:09}. It is thus likely that
the source of the criticality of natural images is these high
order statistics, which explains why our learned Boltzmann machine
does not show criticality, and reinforces the importance of high
order statistics.

\subsection{Relation to Simple-Cell Receptive Fields}\Lsec{Relation}
It is known that simple-cell receptive fields operate as certain
oriented bandpass filters~\cite{Jones:87}. Olshausen and Field
naturally derived these filters by processing natural images based
on {\it sparse coding}~\cite{Olshausen:96}. In this section, we
connect our findings with these known results.

First, we confirm that the oriented bandpass filters are also derived from our binarized pictures after quantization and dithering. We assume that picture $\V{S}$ is represented by a linear superposition of basis functions $\{\V{\phi}_k \}_k$ as 
\be
\V{S} \approx \sum_{k}a_k \V{\phi}_k. 
\ee 
and the coefficients $\{a_k\}_k$ are sparse (a few non-zero components are needed to express an image) if we choose an appropriate set of basis functions $\{\V{\phi}_k \}_k$. Olshausen and Field constructed such an appropriate set by solving the following optimization problem
\be
\min_{\V{\phi}} \min_{\{a_k\}_k} \lb ||\V{S}-\sum_{k}a_k \V{\phi}_k||_2^2+R(\{a_k\}_k) \rb,
\ee
where $R(\{a_k\}_k)$ is an appropriate regularization term to induce the sparsity of the coefficients $\{a_k\}_k$. Some typical choices are $R(\{a_k\}_k)=\lambda \sum_k |a_k|$ or $R(\{a_k\}_k)=\sum_k \log (1+a_k^2)$. To follow their way, we use their numerical package called {\it sparsenet}~\cite{sparsenet}. According to their method, we whiten our binarized images $\{ \V{S}^{(\mu)} \}_{\mu}$ and cut them into patches of an appropriate size (here we choose $8 \times 8$ patches). After these preprocessing, we construct the basis function by using their algorithm. 

The result of the construction of the basis function is depicted in \Rfig{filter}.
\begin{figure}[htbp]
\begin{center}
  \includegraphics[width=0.48\columnwidth]{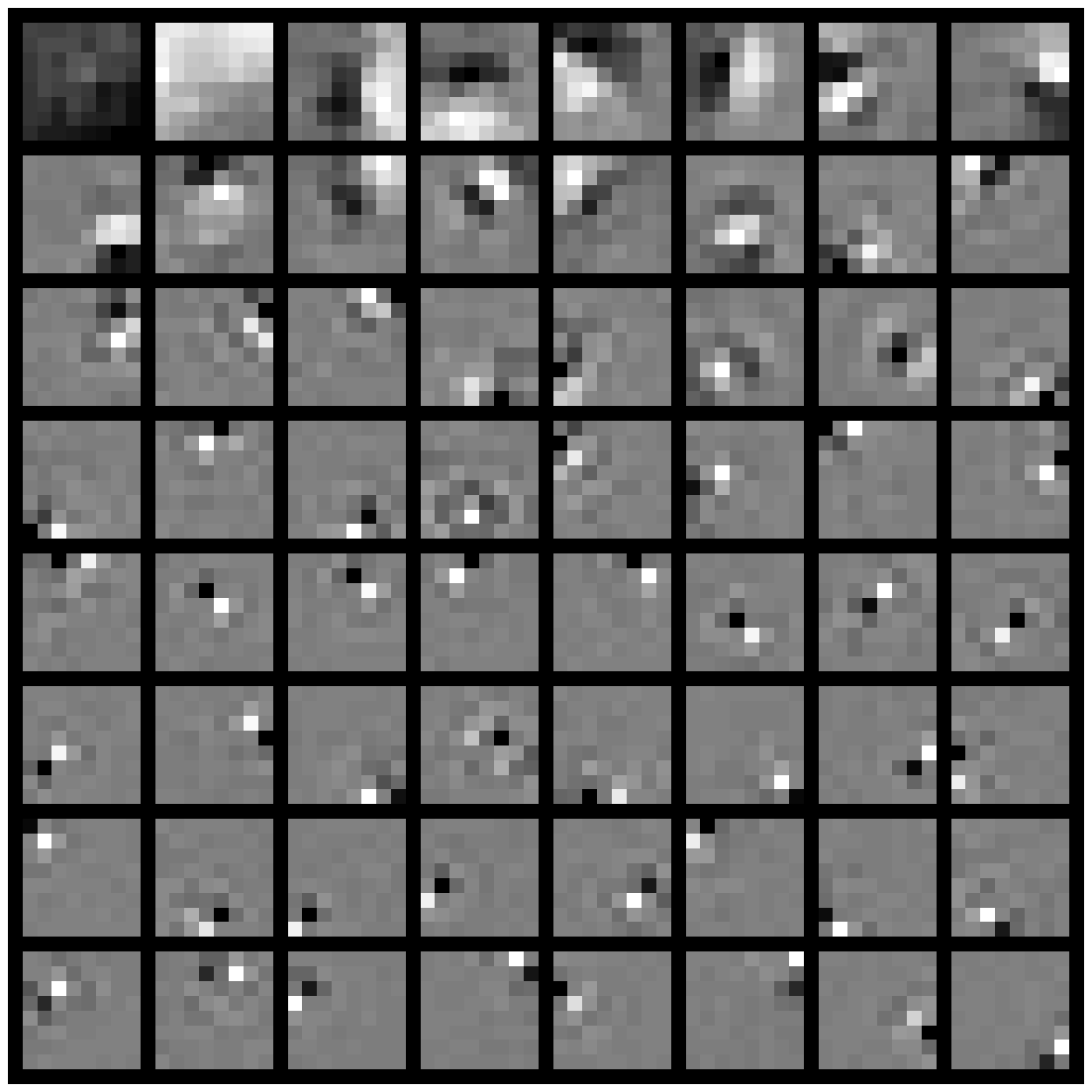}
  \includegraphics[width=0.48\columnwidth]{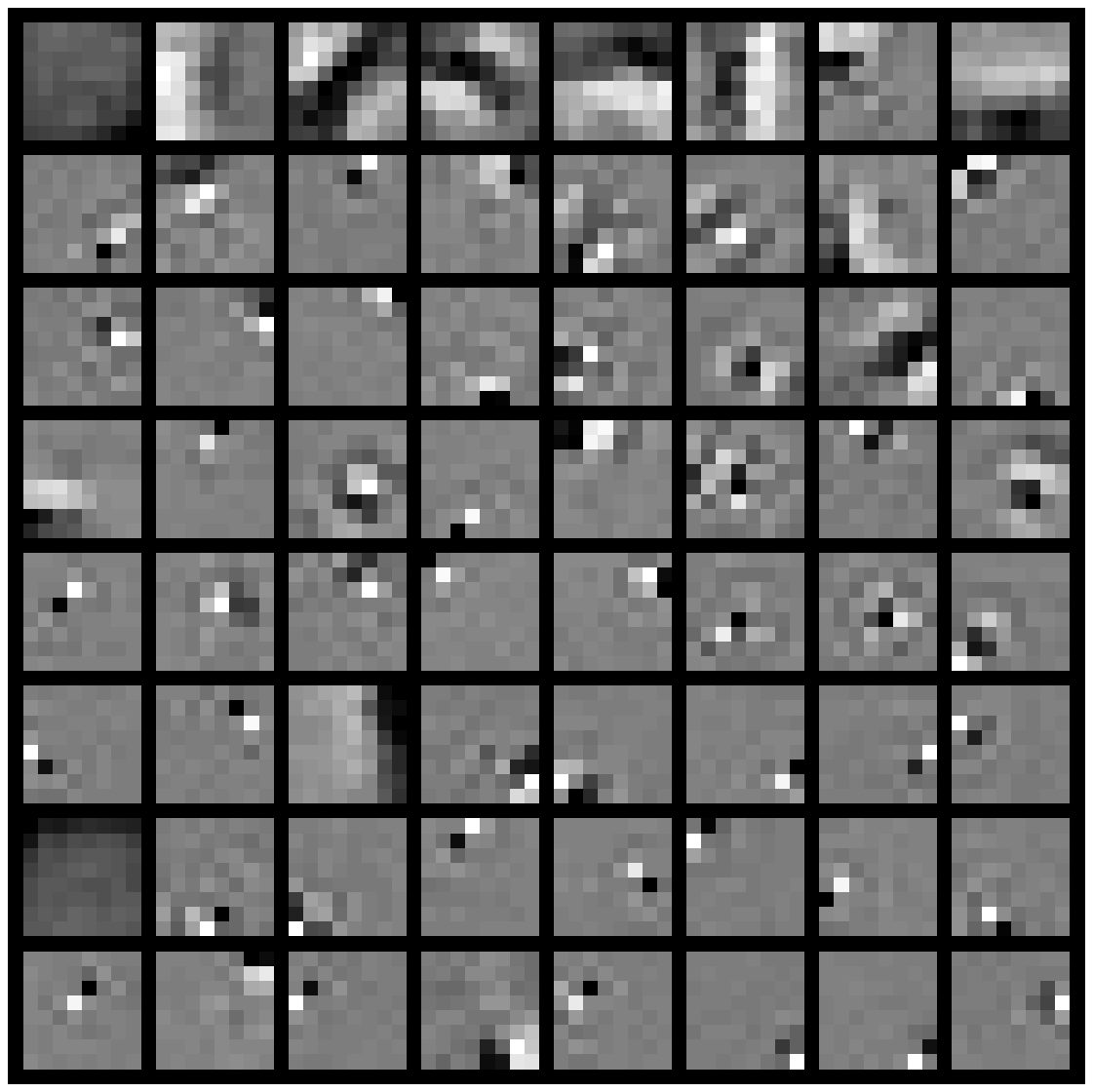}
  \caption{Results of sparse-coding construction of $64$ basis functions estimated from aerial (left) and face pictures (right), both of which are cut into $8\times 8$-size patches after binarization employing quantization and dithering. Both panels consist of 64 square images each of which represents a basis function. Each basis function has $8 \times 8$ pixels. These two panels are quite similar. Several basis functions exhibit orientations seen as a pair of neighboring white and black pixels.  }
\Lfig{filter}
\end{center}
\end{figure}
Here $64$ different basis functions are shown. Each basis function is a square image of $8\times 8$ pixels in a panel. The results of the two sets of pictures, aerial (left panel) and face pictures (right panel), are very similar. They are also similar to those derived in~\cite{Olshausen:96}, although our patterns are more
localized, which is presumably due to binarization. These results imply that the basis functions of natural images derived by sparse coding are fairly universal~\cite{Simoncelli:01}. Hence, we may assume the presence of an universal set of basis functions that can describe any natural image.

The probability distribution of images $P(\V{S})$, which is the object of the analysis and approximated by the Boltzmann machine in this study, can be connected to the sparse representation above by the Bayes rule 
\be 
P\lb \V{S} \rb=\int \prod_{k}da_k~P \lb
\V{S}| \{a_k\}_k, \{ \V{\phi_k} \}_k \rb P\lb \{a_k \}_k \rb,
\Leq{Bayes}
\ee 
where 
$P \lb  \V{S}| \{a_k\}_k, \{ \V{\phi_k}\}_k \rb=\delta(\V{S}-\sum_{k}a_k \V{\phi}_k)$ and $P\lb \{a_k\}_k \rb$ is the prior distribution of the coefficients. The results of the Boltzmann machine learning indicate that $P\lb \V{S} \rb$ is different among the different sets of images, implying that $P\lb \{a_k \}_k \rb$ is different among these sets of pictures, since the basis functions $\{ \V{\phi}_k \}_k$ are universal. These considerations show that our observation based on the Boltzmann machine captures some characteristics of the prior distribution $P\lb \{a_k \}_k \rb$.

The prior distribution $P\lb \{a_k \}_k \rb$ should reflect the
sparseness of $\{a_k\}_k$ and be far from Gaussian. In fact, the
high order statistics of natural images originate in the
non-Gaussianity of $P\lb \{a_k \}_k \rb$. If $P\lb \{a_k \}_k \rb$
is a multivariate Gaussian, the corresponding $P(\V{S})$ becomes
quadratic with respect to $\V{S}$, meaning that the Boltzmann
machine is sufficient to learn all the characteristics of images;
however, this would not be the case as discussed thus far. The
appropriate functional form of $P\lb \{a_k \}_k \rb$  contains
certain hyper parameters reflecting the sparseness of $\{a_k\}_k$.
Our results based on the Boltzmann machine suggest that such hyper
parameters can depend on the choice of images, thus we should change the hyper parameters when processing image data. That is the whole message in the present paper. 

Unfortunately, it is not easy to determine the functional form and the hyper parameters of $P\lb \{a_k \}_k \rb$ from our results. This would constitute an interesting future work.

\subsection{Summary and Model Proposition}\Lsec{Summary}
We summarize our observations thus far:
\begin{itemize}
\item{The range of interactions is about $\xi \approx 4$, and in the region $r\geq 2$ the interaction is positive and rapidly decays as $r$ grows. } 
\item{The sublattice structure shown in \Rfig{sublattice} is widely present.} 
\item{Boundary effects exist and tend to give larger values of interactions than
 bulk ones.} 
 \item{Frustration is absent, or quite weak even when it exists.} 
\item{Local fields can be negative even if the magnetizations are biased to be positive. This may be an indirect signal of the importance of high order statistics.} 
\item{The criticality of natural images is not captured by the Boltzmann
machine, may be because of the lack of higher order statistics than the second.}
\end{itemize}
According to these findings, we propose a model of the prior
distribution of natural images with six parameters: \be
w_{ij}=w(\V{r}_i,\V{r}_j)= \left\{
\begin{array}{cc}
w_{NN}^{1},  &  ( {\rm The\,\,NN\,\,interaction\,\,of\,\,blue\,\,link})    \\
w_{NN}^{2},  &  ( {\rm The\,\,NN\,\,interaction\,\,of\,\,magenta\,\,link})    \\
w_{NNN}^{1},  &  ( {\rm The\,\,NNN\,\,interaction\,\,of\,\,blue\,\,link})    \\
w_{NNN}^{2},  &  ( {\rm The\,\,NNN\,\,interaction\,\,of\,\,magenta\,\,link})    \\
a e^{-|r_{ij}-2|/b},  &  ({\rm otherwise})
\end{array}
\right.. \ee These six parameters can be adaptively determined
according to specific problems, but we admit this is not always
easy. For convenience in such situations, we display example
values of these parameters estimated through our Boltzmann machine
learning for patch size $L=16$ in Table \ref{tab:model
parameters}.
\begin{table}[htbp]
\begin{center}
\begin{tabular}{|c|c|c|c|}
\hline
\multicolumn{1}{|c|}{} & \multicolumn{1}{|c|}{Aerial} &\multicolumn{1}{|c|}{Face}  & \multicolumn{1}{|c|}{Forest} \\
\cline{2-4}
\hline
\hline
$w_{NN}^{1}$    & 0.07  & -0.85 & -0.03 \\
\hline
$w_{NN}^{2}$    & 0.32  & 0.2    & 0.43 \\
\hline
$w_{NNN}^{1}$ & 0.24  & -0.14 & 0.3\\
\hline
$w_{NNN}^{2}$ & 0.22  & 0.4    & 0.37\\
\hline
a                        & 0.1    & 0.3    & 0.16\\
\hline
b                        & 0.7    & 1.1    & 1.3\\
\hline
\end{tabular}
\caption{Model parameters estimated by our Boltzmann machine
learning of natural images of size $L=16$ by the NMF}
\label{tab:model parameters}
\end{center}
\end{table}
In particular, the NN and NNN interactions, $w_{NN}$ and
$w_{NNN}$, are the average values of $w_{ij}$ over the
corresponding histograms shown in Figs. \NRfig{sublattice},
\NRfig{L=16-faces-NMF-Hist}, and
\NRfig{L=16-forestneedles-NMF-Hist}. The parameters $a$ and $b$
are estimated through fitting based on \Req{exp}. Two different
values of $a$, as well as of $b$, corresponding to two sublattices
are averaged to give the values in Table \ref{tab:model
parameters}. As observed, $w_{NNN}^{1}$ and $w_{NNN}^{2}$ are
similar for aerial and forest pictures, in which it is difficult
to see clear periodicity in the NNN interactions. This prior with
estimation of the parameters is the main result of the present
study.

\section{Conclusion}\Lsec{Conclusion}
In this paper, we investigated the prior distributions of natural
images by employing the Boltzmann machine. We prepared three sets
of different pictures, aerial, face, and forest. To reduce the
huge computational time of the learning process, we used the NMF,
which enabled us to handle relatively large patch sizes up to
$L=32$. The results are stable against a change in system size if
the linear size $L$ is larger than or equal to $L=16$. The refined
mean-field method, the BA, was also employed to check the validity
of the NMF, and we found that the NMF results are reliable for the
interactions among the sets of pictures we studied. This conclusion was reinforced by using the Monte Carlo method for small sizes and by examining other dithering methods.  

As individual characteristics of each set of pictures, we found
that the NN and NNN interactions strongly depend on the set of
pictures. Both negative and positive values can appear for these
NN and NNN interactions. Meanwhile, as universal aspects, we observed that the inferred interactions are essentially short-range. For a distance longer than $r \geq 2$, the
interactions basically are positive and decay rapidly among all
the sets of pictures. The characteristic length scale is commonly
about $\xi \approx 4$. Simple periodic behaviors are also observed
in all the cases. Summarizing these properties, we proposed a
model prior distribution with six parameters at most. It will be
interesting future work to examine the performance of this model
distribution in image processing tasks, such as image restoration.

As an additional topic, we also examined the concepts of
frustration and criticality. Frustration is present for the
interactions inferred from the face pictures, but absent for the
other two sets of pictures. In all the cases, the criticality is
not observed in the Boltzmann machine after learning, although our
binarized images show a clear power law in the Fourier amplitude
plotted against the absolute value of the Fourier frequency. We
speculate that this is a weak point of the Boltzmann machine,
which employs only up to the second order statistics; the
criticality in natural images would be sustained by higher order statistics than the second.

The relation to simple-cell receptive fields was also examined.
Our results imply that it can be better to tune the distribution
of the sparse coefficients according to the target  images. This
distribution should have a nontrivial functional form different
from Gaussian. Constructing the functional form and determining
the hyper parameters also constitute interesting future work.

\section*{Acknowledgments}
T. O. is grateful to Y. Kabashima, M. Kikuchi, and K. Tokita for fruitful discussions. This work was supported by Grant-in-Aid for JSPS Fellows (No. 2011) (TO),  KAKENHI No. 26870185, 25120013 (TO) and No. 15K00330 (MY).

\end{document}